%% file: emnlp2023.tex
\pdfoutput=1

\documentclass[11pt]{article}

\usepackage{acl2023}

\usepackage{times}
\usepackage{latexsym}
\usepackage{multirow}
\usepackage{multicol}
\usepackage{amsmath}
\usepackage{booktabs}
\usepackage{xcolor}
\usepackage{float}
\usepackage{algpseudocode}
\usepackage{listings}
\usepackage{color}
\usepackage{footnote}
\usepackage{tabularray}
\usepackage{longtable}
\DeclareCaptionLabelFormat{andtable}{#1~#2  \&  \tablename~\thetable}
\usepackage{graphicx}
\usepackage[flushleft]{threeparttable}
\definecolor{backcolour}{rgb}{0.95, 0.95, 0.96}
\lstdefinestyle{mystyle}{
    backgroundcolor=\color{backcolour},   
    showtabs=false,                  
    tabsize=2
}

\usepackage{enumitem}

\usepackage[normalem]{ulem}
\newcommand{\ensuretext}[1]{#1}
\newcommand{\nertcomment}[4]{\ensuretext{\textcolor{#3}{[\ensuretext{\textcolor{#3}{\ensuremath{^{\textsc{#1}}_{\textsc{#2}}}}} #4]}}}
\newcommand{\an}[1]{\nertcomment{A}{A}{magenta}{#1}}

\DeclareMathOperator*{\score}{score}
\DeclareMathOperator*{\ts}{ts}
\usepackage{scalerel}
\newcommand\smallplus{\raisebox{1pt}{\scaleobj{0.8}{+}}}
\usepackage{array}
\newcolumntype{P}[1]{>{\centering\arraybackslash}p{#1}}
\newcolumntype{M}[1]{>{\centering\arraybackslash}m{#1}}

\newcommand{\tikzxmark}{%
\tikz[scale=0.23] {
    \draw[line width=0.7,line cap=round] (0,0) to [bend left=6] (1,1);
    \draw[line width=0.7,line cap=round] (0.2,0.95) to [bend right=3] (0.8,0.05);
}}

\usepackage{tikz}

\usepackage[T1]{fontenc}

\usepackage[utf8]{inputenc}

\usepackage{microtype}

\usepackage{inconsolata}

\title{An Efficient Approach for Studying Cross-Lingual Transfer in Multilingual Language Models}

\author{Fahim Faisal, Antonios Anastasopoulos\\
Department of Computer Science, George Mason University\\
\texttt{\{ffaisal,antonis\}@gmu.edu}}

\begin{document}
\maketitle
\begin{abstract}
The capacity and effectiveness of pre-trained multilingual models (MLMs) for zero-shot cross-lingual transfer is well established. However, phenomena of positive or negative transfer, and the effect of language choice still need to be fully understood, especially in the complex setting of massively multilingual LMs.
We propose an \textit{efficient} method to study transfer language influence in zero-shot performance on another target language. Unlike previous work, our approach \textit{disentangles downstream tasks from language}, using dedicated adapter units. Our findings suggest that some languages do not largely affect others, while some languages, especially ones unseen during pre-training, can be extremely beneficial or detrimental for different target languages. We find that no transfer language is beneficial for all target languages. We do, curiously, observe languages previously unseen by MLMs consistently benefit from transfer from \textit{almost any} language. We additionally use our modular approach to quantify negative interference efficiently and catagorize languages accordingly. Furthermore, we provide a list of promising transfer-target language configurations that consistently lead to target language performance improvements.\footnote{Code and data are publicly available: \url{https://github.com/ffaisal93/neg_inf}}

\end{abstract}

\section{Introduction}


Pretrained Multilingual Models (MLMs) perform surprisingly well in terms of 
zero-shot cross-lingual transfer even though no explicit cross-lingual signal was present during pretraining. Subword fertility~\cite{deshpande-etal-2022-bert}, token sharing~\cite{dufter-schutze-2020-identifying}, script~\cite{muller-etal-2021-unseen}, as well as balanced language representation~\cite{rust-etal-2021-good} contribute to this effectiveness. But, by and large, the most important component seems to be the combination of languages the model is trained and evaluated on. It is important, hence, to understand why and when cross-lingual transfer is successful at the language level. 

\begin{figure}
    \centering
    \includegraphics[width=.5\textwidth]{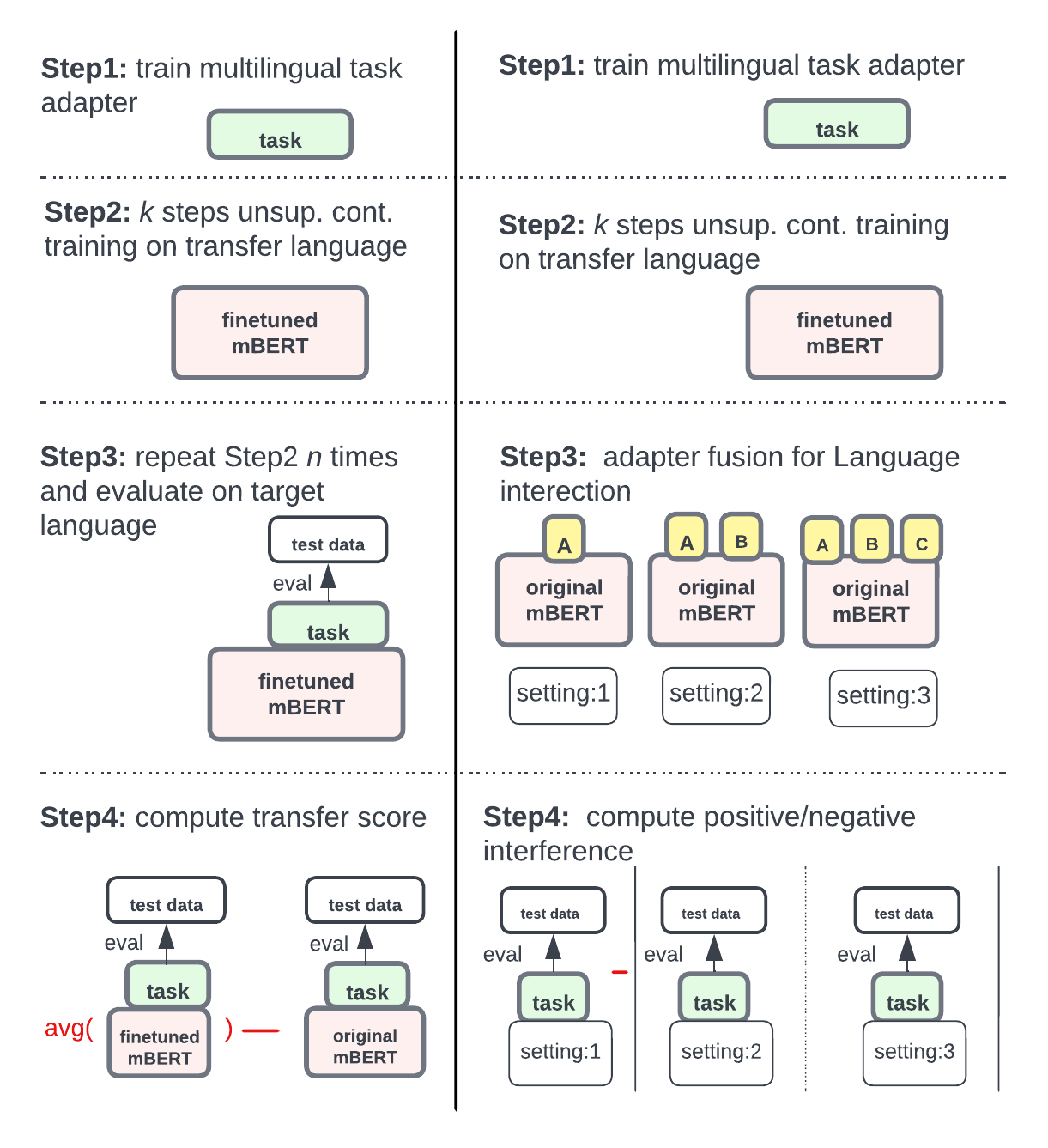}
    \caption{Our approach uses efficient few-step continued tuning (left) and adapter modules (right) to disentangle the effect of \textit{task} and \textit{language} to quantify the effect of a \textit{transfer} language for a given task and model. The left panel depicts the framework for our cross-lingual transfer, while the right panel represents the scenario of multiple language interactions followed by quantifying negative interference.}
    \label{fig:main_ex}
    \vspace{-1em}
\end{figure}

Previous attempts at studying cross-lingual transfer fall into two categories. First, the most popular approaches are those which, given a task and a MLM, task-tune the MLM on annotated data from a \textit{transfer} language and then evaluate on a \textit{target} language~\cite[e.g.][]{lin-etal-2019-choosing}. The problem with such approaches is that (a) they do not disentangle the effect of task and language, since they train directly on the task using \textit{annotated} data in the transfer language, and (b) it is expensive to task-tune the whole model for all possible transfer languages. 


Second, other approaches tackle the inefficiency problem by relying on bilingual approximations: \citet{malkin-etal-2022-balanced} for instance train bi-lingual BERT~\cite{devlin-etal-2019-bert} models, task-tune them on the transfer language and then evaluate on the \textit{target} one, and contrast this performance to a monolingual target-language BERT. While this approach ignores the fact that language interactions can be different in multilingual and bilingual models~\cite{wang-etal-2020-negative,papadimitriou2022multilingual}, it does correlate decently with transfer performance on multilingual models. However, it still does not disentangle task from language and is quite expensive, as studying $n$ languages requires training $n^2+n$ BERT models.

In this work, we propose an \textit{efficient} approach to study cross-lingual transfer, outlined in Figure~\ref{fig:main_ex}, that also disentangles the effect of task-tuning and the effect of language, while operating within the framework of the same MLM. Our approach relies on learning a separate \textit{task} adapter module to perform the downstream task, which needs to only be trained once (hence it is efficient). We then perform unsupervised finetuning on unannotated transfer language data for a minimal number of steps. Comparing the performance of the model on the target language with and without the previous step results in a direct assessment of the effect of the transfer language without changing the conditions under which the downstream task was learned. In addition, we extend this framework to quantify the negative interference resulted from the interaction of multiple languages (Figure~\ref{fig:main_ex}(right)). With the aid of adapter-fusion tuning~\cite{pfeiffer-etal-2021-adapterfusion}, we compare different combinations of language adapters and compute the interference occurring due to increased interactions. 

We perform extensive analysis using this efficient approach on five downstream tasks using dozens of transfer and target languages (\textit{184 in total}) and devise a metric (which we dub \textit{transfer score}) to quantify which languages have/receive positive or adverse effects on/from others. Last, we focus our analysis on cross-lingual transfer for languages unseen during the pre-training of the MLM.

\section{Methodology}
\label{sec:method}
Adapters~\cite{pfeiffer-etal-2022-lifting} are light-weight parameter-efficient modules that can be injected between the layers of pretrained models. In their typical usecase, the rest of the model is frozen and only the adapter modules are trained, to adapt a model to a new language, domain, or task. Importantly, for our goals, these adapters are also composable: one can stack an independently trained language adapter and task adapter to achieve decent performance for that language on that task. First we use an adapter-based setting to perform our analysis on cross-lingual transfer. Furthermore, we extend our study to negative interference and language interaction through another adapter-fusion-based setting.

\paragraph{Cross-Lingual Transfer} The composable property of adapters allows us to disentangle learning a task from the language representations (the process is also outlined in Figure~\ref{fig:main_ex}).
In step 1, we first train a task-specific adapter \texttt{\texttt{[T]}} (e.g. named entity recognition), on data from as many languages as possible. This module will be responsible for performing the downstream task independently of input language. We then (step 2) finetune the \texttt{[base]} model (e.g. \texttt{mBERT}) on a transfer language $\alpha$ with only a few steps (1, 10, or 100) using masked language modeling, obtaining \texttt{[base$^{\alpha}$]}. Now the language representations of this finetuned model will be (slightly) biased towards the transfer language.

Last, in step 3 we reinsert the task adapter in both the finetuned and the original pretrained model, and use both models to test and evaluate on target language data $\beta$. The difference in performance between these two models $\score(\beta\ ; \mathtt{[base}\smallplus\mathtt{t]}^{\alpha}) - \score(\beta\ ;\mathtt{[base\smallplus t]})$ will reveal whether transfer language $\alpha$ benefits (if positive) or hurts (if negative) target language $\beta$.

An obvious caveat of our approach so far is that a single update (or 10 or 100) with a randomly sampled batch in any language does not allow for any robust conclusions. To avoid this issue, we repeat the above process $n$=$10$ times for each transfer language with different data and aggregate these scores. 

Our final transfer score $\ts(\alpha\rightarrow\beta\ ; \mathtt{base}, \mathtt{t})$ for a given model $\mathtt{base}$ and task $\mathtt{t}$ turns the difference of the finetuned and original model into a percentage of the original baseline performance, for fairer comparisons at different levels of performance:
\begin{align*}
\ts(&\alpha\rightarrow\beta\ ; \mathtt{base}, \mathtt{t}) = \\ 
&\frac{\frac{\sum_{1}^{10}\score(\beta\ ; \mathtt{[base}\smallplus\mathtt{t]}^{\alpha})}{n} - \score(\beta\ ;\mathtt{[base\smallplus t]})}{\score(\beta\ ;\mathtt{[base\smallplus t])}}
\end{align*}

\paragraph{Negative Interference}

The typical definition of negative interference describes it as the phenomenon when batches in different languages produce opposite gradients during training. We instead focus on downstream performance, in line with most studies focusing on cross-lingual transfer, assuming that a negative effect on performance implies negative interference. Another reason is that, in $n$ dimensional spaces, there extremely high probability of two random vectors being orthogonal; hence any two gradient vectors could certainly be orthogonal without necessarily impacting downstream performance.

To quantify negative interference, we follow a modular-based approach depicted in Figure~\ref{fig:main_ex}(right). Like before, we separate the task and language, followed by performing interaction among multiple languages. However, we use language adapters at this time instead of continuously finetuning the base  model. This strategy allows us to efficiently train multiple language sub-parts only once (Step2) followed by mixing those modules through adapter fusion~\cite{pfeiffer-etal-2021-adapterfusion}. In our experiments, we train a set of language adapters and make either monolingual settings or a combination of bilingual/trilingual interactions (Step3). Then we stack previously trained task adapter while only changing the underlying language combination. Finally, we extract the interference score from the difference between already computed multilingual and monolingual counterparts (Step4). 

Having these interference scores at hand, we can tell whether a language actually gets benefits or not while influencing the associated languages in a positive/negative manner. For example, consider language $\mathtt{A}$ interacting with language $\mathtt{B}$. We can easily quantify the interference of language $\mathtt{A}$ by calculating the loss/gain of this bilingual interaction \texttt{[AB]}: a score increase for $A$ compared to its monolingual counterpart (i.e. $\mathtt{+A = +_{\texttt{[AB]}-[A]}}$) means positive interference for $\mathtt{A}$ in this particular setting. We can further extend this to a trilingual setting as well (i.e. $\mathtt{+A = +_{\texttt{[ABC]}-[A]}}$). Using these scores, we can get different combinations of interference scenarios by counting the co-occurred positive/negative interference. We use $\mathtt{|+A,+B|}$ to denote the number of cases where $\mathtt{A}$ benefits both itself and $\mathtt{B}$, presenting all possible rules in Table~\ref{tab:inf_calc}. Utilizing these rules, we can identify how much language A actually gains or loses during its bilingual/trilingual interactions while providing substantial interference to other languages.




\begin{table}[t]
    \centering
    \small
    \begin{tabular}{p{7cm}}
    \toprule
    Notations ($+$: win, $-$: loss)\\
    \midrule
    \begin{enumerate}[noitemsep,nolistsep,leftmargin=*]
    \item $\mathtt{|+A|}$ = \textit{count}( A gains in interaction \texttt{\texttt{[AB]}} or \texttt{[ABC]}) 
    \item $\mathtt{|-A|}$ = \textit{count}(A losses in interaction \texttt{[AB]} or \texttt{[ABC]})
    \item $\mathtt{|+A, +B|}$ = \textit{count}(Both language gets benefit). In other words, A gains. At the same time, B receives benefits while interacting with A.
    \end{enumerate}\\
    \midrule
    \end{tabular}
    \begin{tabular}{c|cc}
     Bilingual Interactions    &  \multicolumn{2}{c}{Trilingual Interactions}\\
    \midrule
     $\mathtt{|-A,-B|}$  & $\mathtt{|-A, -B, -C|}$ & $\mathtt{|-A, -B, +C|}$\\
    $\mathtt{|-A, +B|}$  & $\mathtt{|-A, +B, -C|}$ & $\mathtt{|-A, +B, +C|}$\\ 
    $\mathtt{|+A, -B|}$ 
         & $\mathtt{|+A, -B, -C|}$
         & $\mathtt{|+A, -B, +C|}$\\
    $\mathtt{|+A, +B|}$  
         & $\mathtt{|+A, +B, -C|}$
         & $\mathtt{|+A, +B, +C|}$\\
    \bottomrule
    \end{tabular}
    \caption{Interference calculation for language A. $\mathtt{|+A|}$ means the number of cases where A itself gets benefits. If the setting is bilingual, then $\mathtt{|+A|=} \text{count}(+_{\mathtt{[AB]}-[A])}$ (i.e. if the evaluation score on task language A: $\mathtt{[AB]-[A]>0}$ for the combination \texttt{[AB]}, we get a $\mathtt{+A}$.)}
    \label{tab:inf_calc}

\vspace{-1em}

\end{table}

Moreover, we can use these interference combination counts to project languages in an interference representation space. For example, consider a 2-D space of bilingual interaction where the X-axis represents the negative/positive interference a language receives from one such interaction and the Y-axis is for the interference it provides to other languages. We can project a language using the dot product of counts (eg. $\mathtt{|+A, -B|}$) with its corresponding quadrant identifier $[1,-1]$. As a result, the projection coordinates $\mathtt(x_A, y_A)$ for language $A$ in a bilingual interaction could be obtained as follows:
\begin{align*}
\mathtt{C} =& \mathtt{|-A, -B|\smallplus|-A, \smallplus B|\smallplus|\smallplus A, -B|}\\&
\smallplus\mathtt{|\smallplus A, \smallplus B|}  \\
\mathtt{(x_A, y_A)} =&\mathtt{\frac{1}{C}\times (|-A, -B|\cdot[-1,-1]}\\&
\mathtt{\smallplus |-A, \smallplus B|\cdot[-1,1]}
\mathtt{\smallplus |\smallplus A, -B|\cdot[1,-1]}\\&
\mathtt{\smallplus |\smallplus A, \smallplus B|\cdot[1,1])}
\end{align*}
Using the above-mentioned projections, we visualize a language in a way that represents how much interference it provides as well as receives (see example with each step of the calculation in Appendix~\S\ref{sec:int_example}). We can further extend this strategy to the trilingual setting, but now we have to deal with eight axes instead of four. In Figure~\ref{fig:int_result} of the result section, we present the language interaction visualizations for bilingual and trilingual scenarios.

\section{Experimental Setup}
We conduct our experiments in two different settings targeted to perform two different analyses: first understanding the language effect on cross-lingual transfer and then, extending this to quantify language-language interaction.

Primarily, we use multilingual BERT as our base model and report XLM-R results for comparative model evaluation. We use a total of 38 transfer languages (11 unseen during pretraining) to finetune the MLM using masked language modeling with the process described above. Using these transfer languages, we do monolingual finetuning on \texttt{mBERT} for either 1, 10, 100, or 1000 steps and each experiment is repeated for 10 times. At the sametime, we trained multilingual task adapters followed by task evaluation on the following tasks:

\begin{itemize}[noitemsep,nolistsep,leftmargin=*]
    \item \textbf{Token-level:} Dependency Parsing (DEP), Part-of-Speech (POS) tagging and Named Entity Recognition (NER). Parsing and POS tagging are evaluated on a set of 114 languages from Universal Dependencies v2.11 \cite{10.1162/coli_a_00402}. For NER, we use 125 languages from the Wikiann~\cite{pan-etal-2017-cross} dataset.
    \item \textbf{Sentence-level:} Natural Language Inference (NLI) evaluated on XNLI~\cite{conneau-etal-2018-xnli} and AmericasNLI (ANLI)~\cite{ebrahimi-etal-2022-americasnli} datasets. 
    \item \textbf{Extractive Question Answering:} Evaluated on TyDiQA~\cite{clark-etal-2020-tydi} gold task.
\end{itemize}

Additionally, we train 38 language adapters to perform the experiment on language-to-language interaction and negative interference. Here, we stack the previously trained task adapter on top of either one or a combination of double or triple language adapters (Figure \ref{fig:main_ex}(b)) and then perform the evaluation on the transfer languages having task data available. All training and evaluation datasets, implementation and hyper-parameter details are provided in Appendices~\ref{sec:dataset}-\ref{sec:parameters} (Table \ref{tab:src_lang}-\ref{tab:eval_dataset}).

\section{Results and Discussion}
First, in \ref{sub:compare}, we present a comparative scenario in between continuous training and language interaction in terms of performance improvement over the baseline model. Then in \ref{sub:cont}, we discuss the findings of continuous training in the context of cross-lingual transfer. After that, in \ref{sub:int}, we present the representation of language interactions as well as interference following the strategy discussed in Section \ref{sec:method}.

\subsection{Continuous Training vs Language Interaction}
\label{sub:compare}

\begin{table}[t]
    \centering
    \small

    \begin{tabular}{l|r|l@{ }c|rrr}
    \toprule
    &&\multicolumn{2}{c|}{Continious Steps}&\multicolumn{3}{c}{Lang. Interaction}\\
          Lang. &  Base &     k=10 &  k=1000 &    \texttt{[1A]} &    \texttt{[2A]} &    \texttt{[3A]}  \\
    \toprule
    \multicolumn{7}{c}{\textbf{Parsing}}\\
    \midrule
    pcm & \textbf{81.1} & 79.1  & 77.9 & 79.3 & 79.5 & 79.5  \\ 
    wol & \textbf{69.5}  & 68.1  & 67.3 & 68.9 & 69.1 & 69.1 \\
    kmr & 31.9  & 31.7  & \textbf{45.3} & 32.6 & 32.1 & 32.0  \\
    bam & 29.9  & 30.9 &  \textbf{38.1} & 30.8 & 30.8 & 30.8  \\
    gub & 21.7  & 20.9 &  \textbf{34.5} & 23.8 & 23.73 & 23.5  \\
    \midrule
\multicolumn{7}{c}{\textbf{POS Tagging}}\\
    \midrule
    pcm & \textbf{92.9}  & 92.2 &  91.2 & 92.3 & 92.5 & 92.6 \\
    wol & \textbf{85.6}  & 84.2 &  82.1 & 84.1 & 84.7 & 84.8  \\
    kmr & 40.2  & 40.5 &  \textbf{55.8} & 41.1 & 40.8 & 40.7  \\
    bam & 30.3 & 30.8 &  \textbf{49.5} & 30.7 & 30.5 & 30.5  \\
    gub & 28.5  & 28.7 &  \textbf{36.7} & 28.8 & 28.8 & 28.9 \\
\midrule
\multicolumn{7}{c}{\textbf{NER}}\\
    \midrule
    ibo & \textbf{61.1}  & 57.2  & 55.4 & 57.5 & 57.8 & 57.7  \\
    pms & 88.2  & \textbf{88.9}  & 87.6 & 88.2 & 87.5 & 87.6 \\
    kin & \textbf{72.4}  & 71.8 & 68.5 & 70.5 & 71.1 & 71.9  \\
\bottomrule
    \end{tabular}
    \caption{Task results for transfer languages unseen by \texttt{mBERT}. \textbf{base:} zero-shot with task adapter \texttt{[T]}. \textbf{Continuous Steps:} do $k$ steps of finetuning on that language plus \texttt{[T]}. \textbf{Lang. Interaction:} introducing language adapters; \texttt{[1A]}: just 1 adapter (in language) and evaluate on it; \texttt{[2A]}: 2 language adapters, the target lang. and one test (the result is averaged for all transfer langs.); \texttt{[3A]}: 3 lang. adapters (results are average again). The highest obtained score for each language is bolded.}
    \label{tab:comp_score_unseen}


\end{table}
Here we present 8 sets of scores for each token-level task. The baseline is where we stack the task adapter on the base pretrained \texttt{mBERT} (i.e. zero-shot task on pretrained \texttt{mBERT}+ \texttt{[T]}). Then for all the evaluation languages, we perform 4 sets of cross-lingual transfers (i.e. 1, 10, 100, and 1000 steps of continuous training). For the language-language interaction experiment, we only perform the evaluation on transfer languages where either 1, 2 or 3 language adapters are fused together before stacking the task adapter (i.e. \texttt{[1A]}, \texttt{[2A]}, \texttt{[3A]}). 

\paragraph{Only Unseen Transfers} In Table \ref{tab:comp_score_unseen}, we present our token-level evaluation report for transfer languages unseen during the pretraining phase. For the \texttt{[2A]} and \texttt{[3A]} language interaction results, we compute and report the average score where the evaluation language is also present in the \texttt{[2A]} or \texttt{[3A]} adapter fusion. For tasks where word-to-word relation plays a critical role (parsing and pos tagging), we observe similar patterns of improvement over baseline in both Cont. steps and lang. interaction settings. Whereas, for a task like NER, we do not observe any improvement over baseline both in sustained cont. (k=1000) and interaction settings. Even though we are evaluating the same language after continuous masked language modeling (mlm) or adapter fusion with another high-resource language, there is no clear winning formula that can always serve the unseen low-resource languages. 

\begin{figure}
    \centering
    \includegraphics[width=.45\textwidth]{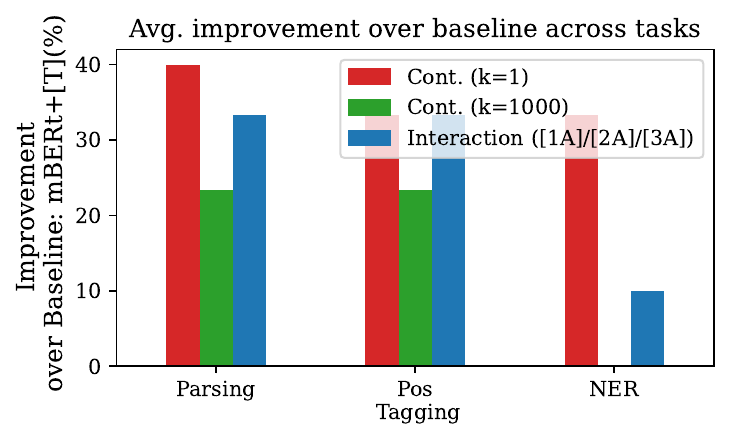}
    \caption{Average score improvement over baseline across tasks for the transfer languages (evaluated on itself). We observe a spike of over 33\% positive score at continuous training step 1. Among these, only 23.3\% cases result in sustained improvement after 1000 steps (0\% in NER). On the contrary, standard language adapter interaction stays at 25\% average improvement.}
    \label{fig:percent}
\end{figure}
\paragraph{Unseen+Seen Transfers} On the other hand, when we consider the case of both unseen and seen languages together in token-level tasks, we see a spike of 33\% average improvement over baseline with just 1 step of mlm training. However, this improvement percentage gets down to a sustained 23.3\% (except task NER) when we evaluate again after having 1000 steps of training. Whereas, in language interaction settings where we fuse standard well-trained language adapters, we generally observe improvement for those languages which also get benefited from continuous training. The improvement percentage averaged over all 38 transfer languages is presented in Figure \ref{fig:percent}. In addition, we present all the scores for all 38 transfer languages and token-level tasks in App. Tables~\ref{tab:comp_score_udp}, \ref{tab:comp_score_pos}, and~\ref{tab:comp_score_ner}. 


\subsection{Takeaways from Continuous Training}
\label{sub:cont}
\paragraph{No Universal Donor}
\input{donors}

First, we search for transfer languages that can be used for positive transfer for a large set of languages. However, we find no language out of 38 that can positively influence almost all languages using mBERT as base model. For this experiment, we rank the transfer languages based on their averaged transfer score (i.e. \texttt{aggregated-transfer}). In Table~\ref{tab:donors}, we list the top 5 ranked transfer languages with their transfer score (base model: mBERT) and the percentage of target languages that do benefit from them (more details in Appendix~\ref{sec:src_rank}).
We observe,  most languages benefit within the range of 30-45\% of target languages across tasks except NLI. However, we did not receive any positive transfer for both of the two different NLI task datasets (XNLI and ANLI). The maximum positive transfer percentage is from zho in both NER and TyDiQA. Interestingly, low-resourced unseen languages perform well in general as transfer languages: 31.7\% (token-level) and 28.3\% (sentence-level) of top 20 transfer languages are unseen languages.

\paragraph{Base Model and Task Matters} To further investigate the discrepancy observed in NLI task, we replace the base model mBERT with XLM-R (Table \ref{tab:xlmr-mbert-nli}). Unlike mBERT, XLM-R in NLI provides superior performance (XNLI: +44.4\% and ANLI: +41.6\%). This signifies how the choice of the base model in a setting with a disentangled language-task effect could drastically change the cross-lingual transfer performance of certain tasks.

Moreover, we observe the above-discussed rankings of transfer languages vary across tasks. To investigate the underlying similarity, we select a large subset of languages (the common 62 target languages across three token-level tasks) and rank the transfer languages as before. We then compute the Spearman rank correlation and statistical significance \texttt{(p<0.05)} of 
 their transfer scores tasks (see Appendix Table~\ref{tab:correaltion}).
  Only parsing and NER are positively correlated ($\rho$=$0.4$) whereas POS tagging is negatively correlated with the other two tasks. 
 This is somewhat surprising, because we use the same underlying dataset for the parsing and POS tagging tasks. We find only a few transfer languages could effectively provide positive transfer simultaneously across tasks. The 5 common languages in the top 20 across tasks are: \texttt{yor}, \texttt{\textbf{mos}}, \texttt{\textbf{kin}}, \texttt{\textbf{hau}}, and \texttt{tel}. In sort, languages unseen by \texttt{mBERT} (in \textbf{boldface}), exhibit similar ranking across tasks (see Table~\ref{tab:src_rank_t_mbert}-\ref{tab:src_rank_s_x}), whereas others vary. For example, \texttt{zho} is the lowest-ranked one in parsing while being top-ranked in NER! Appendix Figure~\ref{fig:common} shows the number of common languages across tasks.

\paragraph{Unseen Languages Transfer with High Variance} 
\input{variance}

\begin{figure*}[t]
\small
\centering
    \begin{tabular}{p{.5\textwidth}p{.35\textwidth}}
\multicolumn{1}{c}{\includegraphics[width=.5\textwidth]{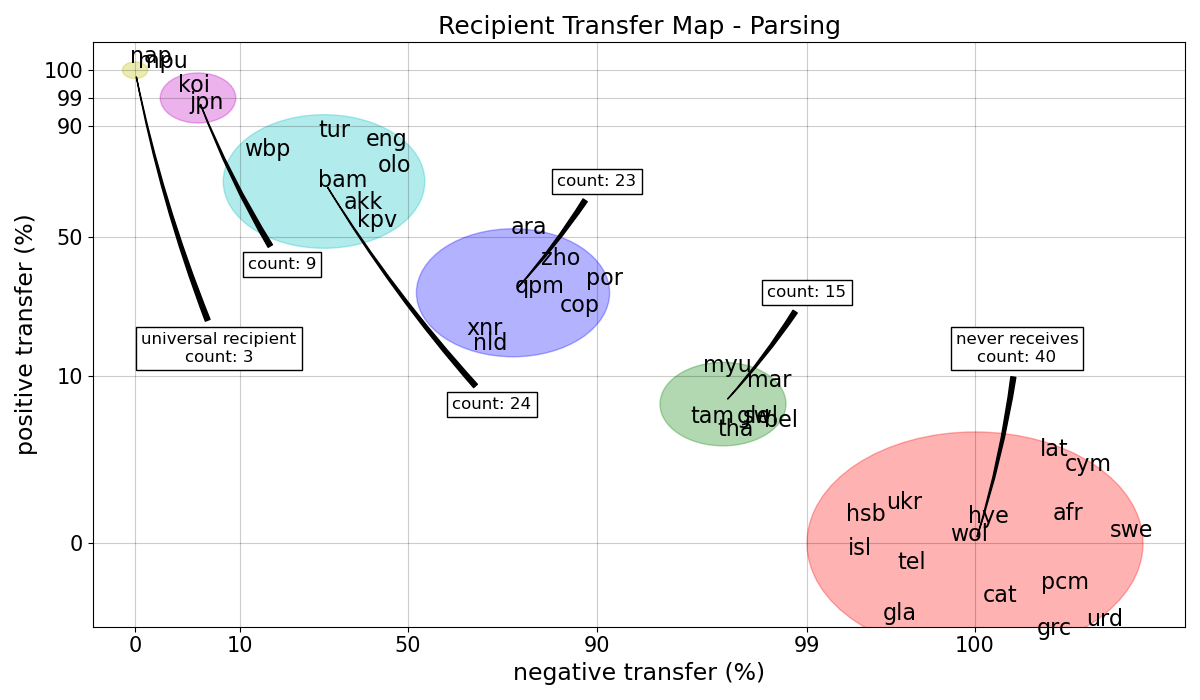}}&
\multicolumn{1}{c}{\includegraphics[width=.25\textwidth]{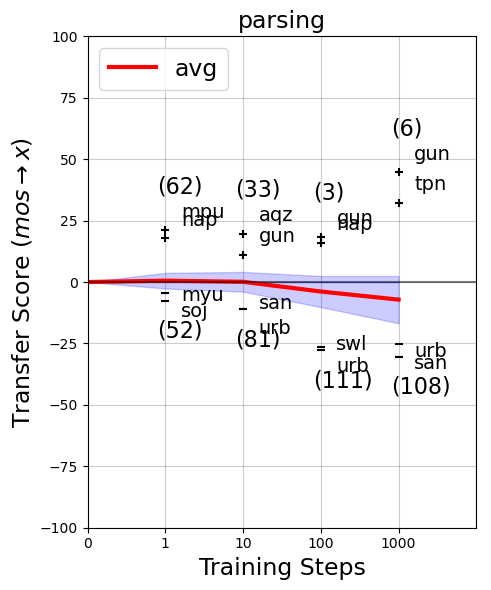}}\\
(a) Target languages mapped based on percentage of receiving positive/negative transfers. & (b) \texttt{Aggregated-transfer} score line with standard deviations through different training steps for Mossi (mos) as transfer language.
    \end{tabular}
    \caption{(a) Some languages exhibit universal recipient nature (yellow) while some never receive positive transfer (red). (b) Shown are the top and bottom two languages receiving maximum/minimum scores (eg. \texttt{gun}, \texttt{tpn} at 1000 steps) at each step, with total positive/negative transfers (in parenthesis) also shown. See Appendix \ref{sec:tr_graphs} for other transfer language score lines.}
    \label{fig:maps}
\vspace{-1em}
\end{figure*}


 We observe that transfer languages with high variance mainly fall into one of three categories: 
 \begin{enumerate}[nolistsep,noitemsep,leftmargin=*]
     \item (+ \textit{and} -): boost performance for some languages while hurt significantly some others; 
     \item (+): mostly (small) positive transfer, significantly hurts only a few languages;
     \item (-): mostly (small) negative transfer, significantly helps only a few languages.
 \end{enumerate}
See examples in Table~\ref{tab:variance} and Appendix \ref{sec:trans_var} for details. Though unseen languages perform well as transfer languages, they usually exhibit the traits of high-variance transfer. Around 90\% of unseen transfer languages are within top-20 languages sorted by variance (see Appendix~Figure~\ref{fig:trans_var}).

\paragraph{Target Language Differences} 
Unlike transfer languages, we find target languages that are almost universal recipients of positive cross-lingual transfer, many of which are unseen by \texttt{mBERT}. On the other hand, some languages do not receive any benefit from the diverse set of transfer languages. In Figure~\ref{fig:maps}(a), we plot the target languages based on the percentage of languages from which they receive positive or negative transfer (see additional maps in Appendix Figure~\ref{fig:all_receiver}). We find around one-third of target languages across three token-level tasks never receive any positive transfer (parsing: 35.1\%, POS: 28.1\%, NER: 32.8\%). Nevertheless, there are target languages (mostly unseen by \texttt{mBERT}) that benefit from all transfer languages (eg. \texttt{nap}, \texttt{mpu} in parsing). See Appendix~\ref{sec:all_receiver} and Table~\ref{tab:receive} for additional results.

\paragraph{Seen vs Unseen Languages}
 Transferring from either seen or unseen languages to unseen languages (i.e. \texttt{transfer(seen/unseen$\rightarrow$ unseen)}) generally helps. For this experiment, we use the large set of token-level task evaluation and 11 transfer languages unseen during \texttt{mBERT} pertaining from diverse families including Indo-European, Afro-Asiatic, Mande, Niger-Congo and Tupian. We observe, that transferring to a large and diverse set of seen languages from unseen languages (i.e. \texttt{transfer(unseen$\rightarrow$ seen)}) does not provide any substantial utility.  Among the three tasks, we get the average transfer as positive for unseen transfer languages just once (dependency parsing, \texttt{transfer(unseen$\rightarrow$ unseen)}). See Figure~\ref{fig:seen-unseen} for the difference of utility provided when the transfer/target languages are seen vs unseen.


\paragraph{Sustained Cross-Lingual Transfer}
Our approach limits step 2 (continued training on the transfer language) to a minimal number of steps. For this section, we extend this to 1000 steps. In the vast majority of transfer-target language combinations, this leads to (small) negative transfer under our setting. We suspect this is due to the underlying model undergoing the first steps of catastrophic forgetting~\cite{mccloskey1989catastrophic}.

There are some languages, though, mostly unseen ones (eg. \texttt{nap, gun, tpn, aqz}) that benefit more from this extended setting. See Appendix~\ref{sec:max_rec} Table~\ref{tab:max_unseen}, where we report the target language receiving the highest benefit from each transfer language for each setting (1,10,100,1000 steps). All the max-utility recipients aside from \texttt{bar} and \texttt{nds} are unseen languages. 
Figure~\ref{fig:maps}(b) presents the training step progression of \texttt{aggregated-transfer} scores for \texttt{Mossi}, one of the most donating transfer languages, and Appendix~\ref{sec:tr_graphs} (Figures~\ref{fig:first}-\ref{fig:last}) shows the transfer progression graphs for all transfer languages.
At the task level, POS tagging always ends up having comparatively higher target language performance variance with more training steps, while NER almost always ends up with negative results with longer training.

\subsection{Takeaway from Language Interactions}
\label{sub:int}
\input{adapter_interference}
We plot all the transfer languages in a 2d axis for both two-language interactions and three-language interactions as shown in Figure \ref{fig:int_result}. 

\paragraph{Bilingual Interactions} First of all, we observe most of the languages mainly fall into either one of the two categories: (1) A(+), B(+): getting benefits from interactions and helping others at the same time, (2) A(-), B(+): Helping other languages but do not get benefits from those languages. Secondly, there are resemblances in how certain languages from specific categories interfere across all 3 tasks. For example, consider the case of \texttt{zho, swe, spa} and \texttt{fra}. These languages fall to the lower right part of all three graphs. However, there are languages like \texttt{ara} that do not uniformly get benefits across three tasks while maintaining it's positive interfering status. Although, there are debates whether English (\texttt{eng}) is an appropriate "hub" language or not~\cite{anastasopoulos-neubig-2020-cross}, \texttt{eng} maintains its status in the upper right quarter making it a good transfer language in all Latin script majority settings. 

\paragraph{Trilingual Interactions} Now we increase the number of languages for a specific transfer language to influence. When we compare the bilingual settings with the trilingual ones (Figure \ref{fig:int_result} (2)), the left-right categorization remains the same. However, many languages receive an uplifting position meaning the strength of performing positive interference increases for those languages (eg. \texttt{are} in dependency parsing, \texttt{zho} in NER). Moreover, we observe an overall decrease in the lower-right corner for both dependency parsing and NER. However, there are languages like \texttt{wol} in POS tagging that goes from upper-left to lower-left. Nonetheless, very few different colored points (i.e. negative coordinate for 3rd language) signify the fact that a multilingual setting is beneficial towards a larger group of recipients. 



\section{Recommendations}
Based on our above findings, we make a number of recommendations in choosing the appropriate transfer language and training scheme for a low-resource setting. 

\begin{enumerate}[nolistsep,noitemsep,leftmargin=*]
    \item There is no universal donor but having multiple transfer languages in the training scheme helps in terms of language interference. 
    \item For universal recipient languages (eg. Typologically diverse unseen ones), including almost any language in the transfer scheme help. 
    \item Low resource unseen languages generally transfer with high variance. A good idea is to include them with other seen languages in the transfer scheme to stabilize the transfer output across a large number of target languages.
    \item Only some of the unseen low-resource ones show sustained transfer toward other low-resource languages through continuous thousand-step training. Usually, the deviation happens during an early stage of training. So just continuing pretraining for longer is not optimal for a scenario with mixed-category languages.
    \item The patterns of receiving positive transfer are similar when we use either one language small-step continuous training or 2/3 standard adapter fusion. So using a large set of trained language adapters fused together according to the need is a simpler way to deal with a large set of mixed-category target languages.
\end{enumerate}

\section{Conclusion and Future Work}
We devise an efficient approach to study cross-lingual transfer in multilingual models for various tasks that disentangles task and language effects. We believe this disentanglement coupled with few-step fine-tuning has the potential to uncover currently uncharted model behaviors (eg. NLI evaluation).  Our findings suggest languages unseen by MLMs clearly exhibit different behavioral pattern compared to other languages in general: they are universal as target, exhibit high variance as transfer language, and their behavior follows similar patterns across tasks. In addition, we do not find a universal donor (a language that benefits all others). Last, we find that some languages 
consistently benefit from settings that resemble "catastrophic forgetting" for other languages, an observation we believe merits a dedicated follow-up study. 

We hope that our approach will allow for further study of cross-lingual transfer for more languages and MLMs, and we plan to extend this in future work, as our findings suggest interesting differences in the behavior of languages used in pre-training and unused ones. Eventually, we hope that our study will also lead to guidelines for selecting appropriate transfer languages, as well as more informed methods for the adaptation of MLMs to new under-served languages.
While our proposed approach being highly efficient to expand the paradigm of cross-lingual transfer evaluation, the findings shed light onto the easy adaptation of MLMs for new languages in a low-resource setting.

\section*{Limitations}
In this work, we primarily experiment with encoder models like mBERT and XLM-R, token-level syntactic tasks and two sentence-level tasks. In future, we would expand this work to recent large language models and tasks involving natural language understanding. Moreover, our work only focus on low-resource setting with small-scale training data and parameter-efficient adapters. In future, instead of monolingual finetuning we will use this parameter efficient approach for multilingual finetuning thus unfolding effective multilingual pretraining configurations. As the base-language model choice, we only use mBERT. The evaluation of cross-lingual transfer needed to be expand to decoder based language models.

\section*{Acknowledgements}
This work has been generously supported by the National Science Foundation under grants IIS-2125466 and IIS-2327143.
We are thankful to the anonymous reviewers and area chairs for their constructive feedback.
This project was supported by resources provided by the Office of Research Computing at George Mason University (\url{https://orc.gmu.edu}) and funded in part by grants from the National Science Foundation (Awards Number 1625039 and 2018631).


\bibliography{anthology,custom}

\clearpage
\appendix

\section{Related Works}
\label{sec:related}
\paragraph{Cross-Lingual Transfer} Studying cross-lingual transfer to prepare a better pretraining configuration is a well-explored topic. \citet{malkin-etal-2022-balanced} propose a balanced-data approach to identify effective set of languages for model training through constructing bilingual language graph. They formulate the problem in terms of linguistic blood bank where language can either play the role of donor or receiver. This study comprises over a large set of languages while training  a large number of bilingual models. However, how a large multilingual model (eg. \texttt{mBERT}) having a shared representation space larger than bilingual models perform in similar setting is not evaluated yet. \citet{fujinuma-etal-2022-match} points out it is always better to have a diverse set of languages during pretraining for zero-shot adaptation. At the same-time, language relatedness in pretraining configuration always helps. 

\paragraph{Adaptation to Unseen Languages} The idea of performing effective zero-shot transfer is highly beneficial for model adaptation to new languages. According to \citet{muller-etal-2021-unseen}, transfer learning helps some new languages while some hard languages does not get the benefit mainly because of the difference in writing systems. Transliterating those languages to a more familiar form is a useful approach in this case.

\paragraph{Parameter Efficiency} Recently parameter-efficient language modeling approaches are becoming more and more popular and capable. Adapter units ~\cite{pfeiffer-etal-2022-lifting} are such modular units containing small trainable set of parameters. Using adapters resolve the problem of model-capacity and training bottleneck. In addition, most of the parameters remain unchanged thus preventing the problem of negative interference. The most important benefit of adapter untis are it's modular design. It is also possible to train the adapters using language-phylogeny information~\cite{faisal-anastasopoulos-2022-phylogeny} thus extending the base model capacity to unseen new language in an informed manner.

\section{Terminologies}
\paragraph{Transfer Language:} The languages we use to perform monolingual finetuning of the base language model (\texttt{mBERT}) using masked language modeling.
\paragraph{Target Language:} The languages we use to evaluate both the pretrained as well as finetuned \texttt{mBERT} on downstream tasks.
\paragraph{Negative Transfer:} The scenario where language model performance drops because of finetuning it on a transfer language.
\paragraph{Cross-lingual Transfer:} The established method of finetuning a language model on one transfer language and deploy it on another target language.
\paragraph{Unseen Languages} Any language that were not part of the original pretraining step.

\section{Dataset Details}
\label{sec:dataset}
\subsection{Transfer Languages}
We perform mono-lingual finetuning as well as language adapter training on 38 transfer languages. Each language dataset contains 10k lines of text. We use texts from several corpus including OSCAR \cite{2022arXiv220106642A} and African News Translation dataset \cite{adelani-etal-2022-thousand}. 11 out of these 38 languages are unseen by \texttt{mBERT} during pretraining steps. The list is provided in Table \ref{tab:src_lang}.

\subsection{Adapter Training Dataset}
\paragraph{Dependency Parsing} We train a task adapter for performing dependency parsing task. For this step, we use Universal Dependency training dataset v2.11 \cite{10.1162/coli_a_00402}. To keep the data distribution balanced, we use not more than a thousand examples per language. Combining all these data together, we train a multilingual dependency tagging task adapter. The complete list of data-source languages for training this adapter is presented in Table \ref{tab:ud_dataset}.

\paragraph{Parts-of-Speech Tagging} Here we also use the Universal Dependency training dataset v2.11 \cite{10.1162/coli_a_00402}. The languages are also the same ones used for dependency parsing previously.

\paragraph{Named Entity Recognition} We use Wikiann~\cite{pan-etal-2017-cross} dataset for training a NER task adapter. The complete language lists are provided in Table \ref{tab:ner_dataset}.

\paragraph{Natural Language Inference} We use XNLI~\cite{conneau-etal-2018-xnli} dataset for training a NLI task adapter. The complete language lists are provided in Table \ref{tab:nli_dataset}.

\paragraph{Extractive Question Answering} We use TyDiQA~\cite{clark-etal-2020-tydi} dataset for training an Extractive Question Answering task adapter. The complete language lists are provided in Table \ref{tab:qa_dataset}.

\subsection{Evaluation Dataset}
We use 125 languages for evaluating NER task from Wikiann. For udp and pos-tagging tasks we use 114 languages from Universal Dependency dataset. There are 62 languages which are common between these two sets of 125 and 114 languages. For NLI evaluation, we use 15 languages from XNLI~\cite{conneau-etal-2018-xnli} dataset and 10 low-resource South American indigenous languages from Americas NLI (ANLI)~\cite{ebrahimi-etal-2022-americasnli} dataset. For the question answering task, we take 9 languages from TydiQA~\cite{clark-etal-2020-tydi} to evaluate. The complete list of 184 evaluation languages are provided in Table \ref{tab:eval_dataset}.

\section{Implementation Details}
\label{sec:implementation}
For all of our experiments, we use as well as modify the scripts from huggingface~\cite{wolf-etal-2020-transformers} and adapterhub~\cite{pfeiffer-etal-2020-adapterhub}. For base language model, we use the model \texttt{bert-base-multilingua-uncased} from huggingface model repository.

\section{Hyper-parameters}
\label{sec:parameters}
\paragraph{Masked Language Modeling finetuning}
\begin{itemize}[noitemsep,nolistsep,leftmargin=*]
    \item Train batch size: 8
    \item Evaluation batch size: 8
    \item Training Steps: 1, 10, 100 and 1000
    \item Learning Rate: 5e-5
    \item Maximum Sequence Length: 512
\end{itemize}

\paragraph{Language Adapter Training: Language Interaction}
\begin{itemize}[noitemsep,nolistsep,leftmargin=*]
    \item Train batch size: 8
    \item Evaluation batch size: 8
    \item Training Epochs: 3
    \item Learning Rate: 5e-4
    \item Maximum Sequence Length: 256
    \item Adapter Parameter Reduction Factor: 16
\end{itemize}

\paragraph{Task Adapter Training: Dependency Parsing}
\begin{itemize}[noitemsep,nolistsep,leftmargin=*]
    \item Train batch size: 36
    \item Evaluation batch size: 8
    \item Training Epochs: 5
    \item Learning Rate: 5e-4
    \item Maximum Sequence Length: 256
    \item Adapter Parameter Reduction Factor: 16
\end{itemize}

\paragraph{Task Adapter Training: POS Tagging}
\begin{itemize}[noitemsep,nolistsep,leftmargin=*]
    \item Train batch size: 36
    \item Evaluation batch size: 8
    \item Training Epochs: 5
    \item Learning Rate: 5e-4
    \item Maximum Sequence Length 256
    \item Adapter Parameter Reduction Factor: 16
\end{itemize}

\paragraph{Task Adapter Training: NER}
\begin{itemize}[noitemsep,nolistsep,leftmargin=*]
    \item Train batch size: 36
    \item Evaluation batch size: 8
    \item Training Epochs: 5
    \item Learning Rate: 5e-4
    \item Maximum Sequence Length: 256
    \item Adapter Parameter Reduction Factor: 16
\end{itemize}

\paragraph{Task Adapter Training: NLI}
\begin{itemize}[noitemsep,nolistsep,leftmargin=*]
    \item Train batch size: 32
    \item Evaluation batch size: 8
    \item Training Epochs: 5
    \item Learning Rate: 5e-5
    \item Maximum Sequence Length: 128
    \item Adapter Parameter Reduction Factor: 16
\end{itemize}

\paragraph{Task Adapter Training: Extractive QA}
\begin{itemize}[noitemsep,nolistsep,leftmargin=*]
    \item Train batch size: 32
    \item Evaluation batch size: 8
    \item Training Epochs: 5
    \item Learning Rate: 3e-5
    \item Maximum Sequence Length: 384
    \item Document Stride: 128
    \item Adapter Parameter Reduction Factor: 16
\end{itemize}

\section{Language Interference Projection (an example)}
\label{sec:int_example}
\input{example_int}

\section{Comparison}
\label{sec:comp}
\input{comp_score}

\section{Transfer Language Ranking}
\label{sec:src_rank}
\input{src_rank}
We rank the transfer languages by aggregating all the transfer scores. For example, consider getting transfer scores $\{ts_1,..ts_i,..ts_n\}$ for a set of $n$ target languages $L_{tg}$ where $i \in L_{tg}$ and the transfer language is $tf$. Then the aggregated transfer score for $tf$ would be:
\begin{align*}
\mathtt{aggregated-transfer(tf)} = & \frac{\sum_{i=1}^{n}ts_i}{n}
\end{align*}
The ranking of all transfer languages across three tasks are presented in Table \ref{tab:src_rank}. In addition, we report the percentage of positive transfers for each transfer language. Both in parsing and POS tagging, we observe significant presence of unseen languages in high ranked positions (percentage of unseen languages in top 10: parsing: 40\%, POS tagging: 40\%, NER: 20\%). At the sametime, they provide positive scores similar to the cases of seen languages. On the contrary, in NER, we observe most of the unseen African languages are at the lower ranked positions.

\section{Recipient Transfer Maps}
\label{sec:all_receiver}
\begin{figure*}
    \centering
    \includegraphics[width=.75\textwidth]{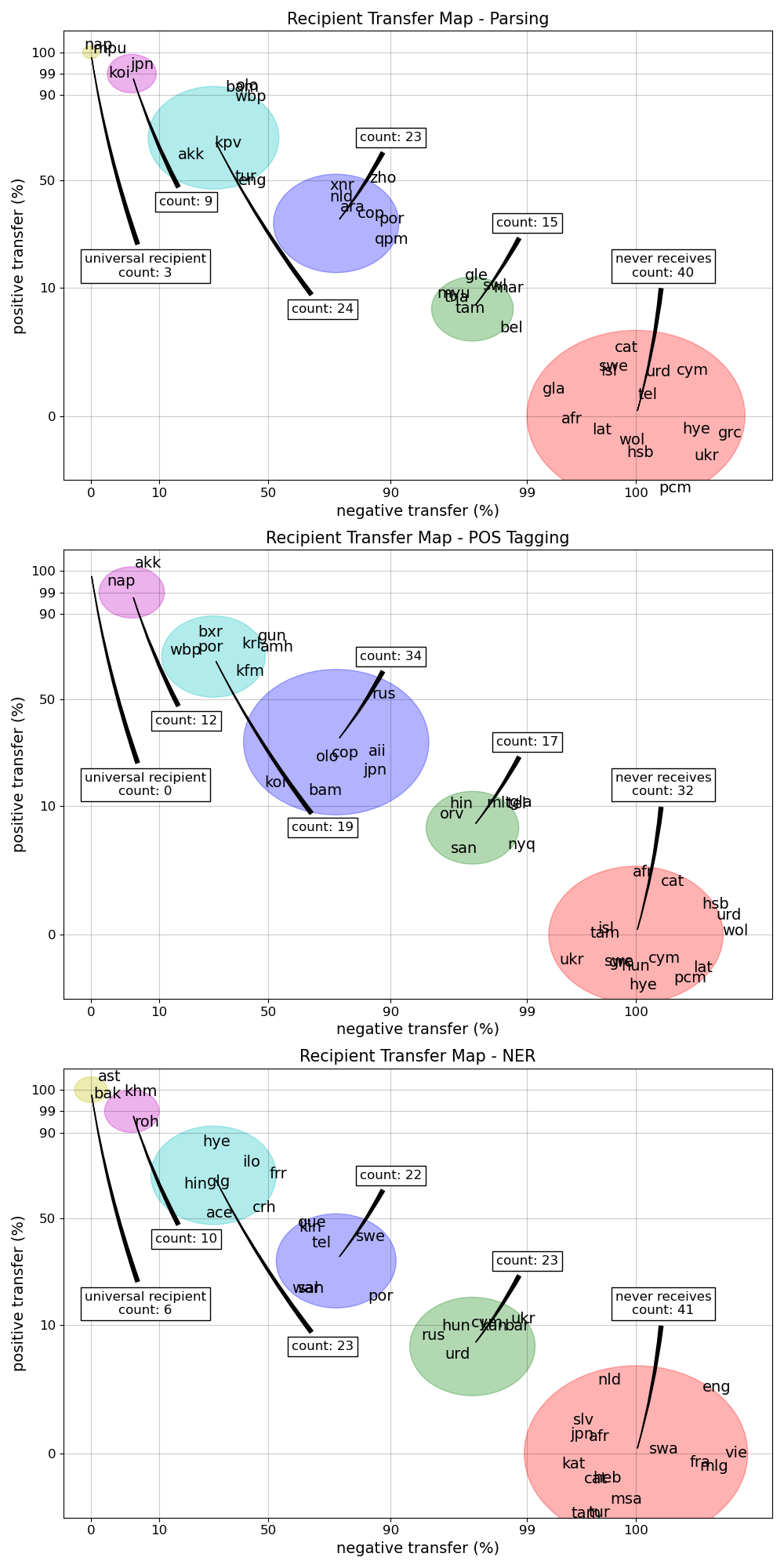}
    \caption{Recipient Transfer Map: we observe universal positive recipients as well as languages those never receive positive transfer across tasks. Circle size represents the percentage of languages fall to a transfer range.}
    \label{fig:all_receiver}
\end{figure*}
\input{receivers}

In a similar manner of calculating the \texttt{aggregated-transfer}, we calculate \texttt{aggregated-target}. For example, if a target language $tg$ receives scores $\{ts_1,..ts_i,..ts_m\}$ from a set of $m$ transfer languages $L_{tf}$ where $i \in L_{tf}$. Then the aggregated target score for $tf$ would be:
\begin{align*}
\mathtt{aggregated-target(tg)} = & \frac{\sum_{i=1}^{n}ts_i}{n}
\end{align*}

This way we identify how much a target language get benefited from all the transfer languages. In Figure \ref{fig:all_receiver}, we present the Recipient Transfer Maps across tasks. We plot the percentage of positive/negative \texttt{aggregated-target} scores and corresponding target languages. Now looking at these maps, we observe the presence of universal target languages (2-5 \%) which always receive positive transfer from all of the 38 source languages in two out of three tasks (exception: POS tagging). Wheres, around 28\% languages in parsing and tagging, 32.8\% in NER never receive any positive transfer. We observe out of 40 languages which receive positive transfer in more than 90\% times, 25 languages are unseen low resourced languages. The complete list of target languages which never receive and which almost always receive positive transfer is presented in Table \ref{tab:receive}.

\section{Maximum Score Recipients are low-resourced}
\label{sec:max_rec}
\input{all_max}
In Table \ref{tab:max_unseen}, we report all the recipients those receive maximum transfer scores at different steps of mlm fine-tuning. From the results, it is evident that, a multilingual model almost always benefits certain unseen, low-resource as well as endangered languages largely. We observe out of 19 max-recipients, 17 are \texttt{mBERT}-unseen languages. Moreover, the two other seen-languages: Bavarian German and Low German are also low-resourced languages.

\section{Task Matters}
\begin{figure}
    \centering
    \includegraphics[width=.4\textwidth]{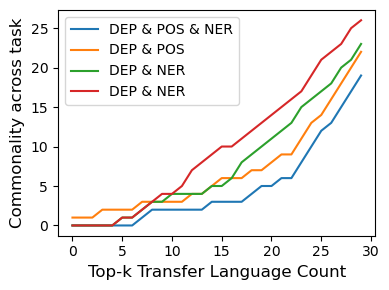}
    \caption{Extent of commonality of top-transfer languages across task. Unseen languages perform generally well while the other language rankings mostly vary across tasks.}
    \label{fig:common}
\vspace{-2em}
\end{figure}

\begin{table}[]
    \centering
    \small
    \begin{tabular}{l|rrr}
    \toprule
    {} &   DEP &   POS &   NER \\
    \midrule
    DEP &  - & \textbf{(-0.34, 0.04)} &  \textbf{(0.40, 0.01)} \\
    POS Tagging & \textbf{(-0.34, 0.04)} &  - & (-0.15, 0.37) \\
    NER &  \textbf{(0.40, 0.01)} & (-0.15, 0.37) &  -) \\
    \bottomrule
    \end{tabular}
    \caption{(Spearman Rank correlation, p value) for correlation of transfer language ranking across token-classification tasks. Statistically significant relations are in \textbf{bold} font.}
    \label{tab:correaltion}
\end{table}

In Figure \ref{fig:common}, we present the commonality graph of transfer language ranking across all three tasks. Spearman rank correlation with p value is presented in Table \ref{tab:correaltion}.

\section{Transfer Languages with High Variance}
\label{sec:trans_var}
\input{transfer_variance}
In Figure \ref{fig:trans_var}, we present the violin plots for all the transfer languages sorted by their \texttt{aggregated-transfer} score variance. We observe, the unseen languages (bold font) are the ones having large amount of variances across all three tasks. We find the languages with high variance can provide superior transfer for some languages but at the same time hurt significantly some other languages. For example, if we consider the case of depndency parsing, we find ibo (rank-1) and bam (rank-3) are two languages with high variance. They provide maximum amount of positive transfer some universal low-resourced target languages like akk, koi, apu, tpn from diverse families including afro-asiatic, uralic, tupian. At the same time, ibo also hurts a large number of languages (10) including fra, nyq, sme, san etc providing minimum amount of negative transfer. On the other hand, there can be languages with high variance providing either mostly positive \texttt{aggregated-transfer} scores like mos or mostly negative score like pcm. Interestingly, if we look at the \texttt{aggregated-transfer} score and variance of pcm in Table \ref{tab:trans_var}, we find the transfer is positive overall. Nevertheless it provides minimum negative scores to 11 languages thus making it a transfer language with high variance. On the other hand, low variance languages are the ones those do not significantly affect any transfer languages like arabic (rank 37). Though the overall transfer score is negative (-0.12) for arabic, it fails to provide maximum or minimum transfer score to any target language making it neutral. So, overall it is evident that, transfer languages with high variance are the ones with either (i) mostly positive while significantly hurting a few, (ii) mostly negative while significantly boosting performance for a few, or (iii) Performing both (i) and (ii) concurrently being highly influential as well as detrimental at the same time. Languages unseen by \texttt{mBERT} during pretraining exhibit all three kinds of characteristics with high intensity (see Table \ref{tab:select_var} for examples). In Table \ref{tab:trans_var}, we report the transfer score with variance as well as the count of maximum/minimum transfer score recipients for all transfer languages across tasks. 

\section{Seen vs Unseen Languages}
\label{sec:seen-unseen}
\begin{figure*}
    \centering
    \includegraphics[width=1\textwidth]{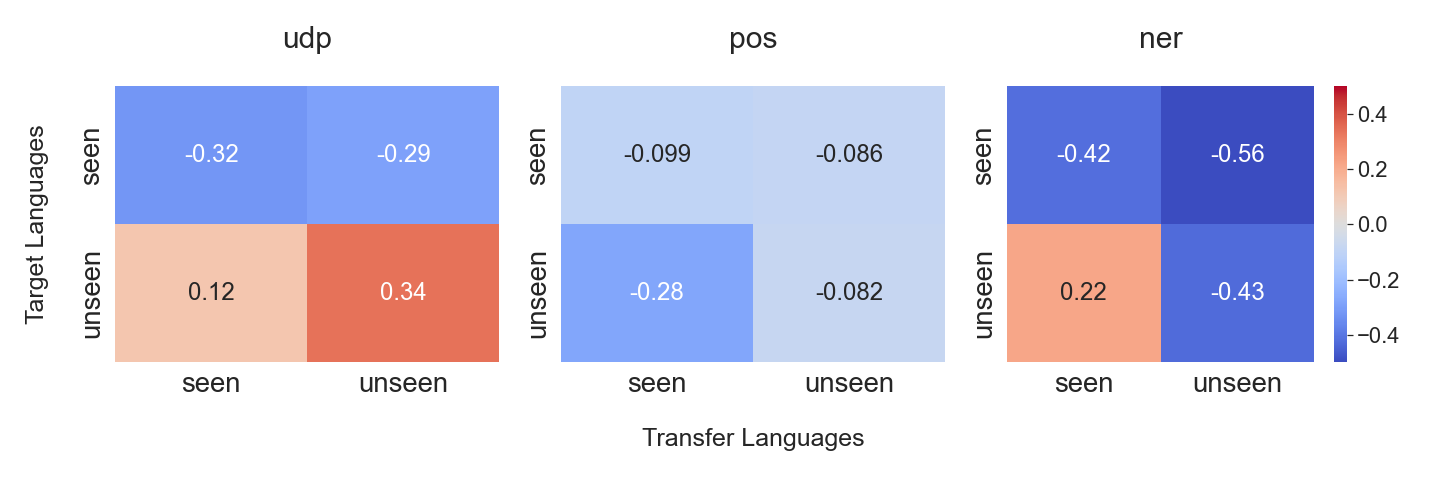}
    \caption{Transfer-Target Heatmap for mbert seen and unseen languages}
    \label{fig:seen-unseen}
\end{figure*}
In Figure \ref{fig:seen-unseen}, we report the aggregated and averaged transfer scores we get for \texttt{mBERT} seen vs unseen languages. 


\section{Transfer Progression Graphs}
\label{sec:tr_graphs}
From Figure \ref{fig:first} to \ref{fig:last}, we present the transfer progression graphs for all 38 transfer languages. We observe POS tagging always have comparatively larger deviation which increase with the progression of training steps. In addition, for different time steps in each graph, we provide percentage of positive/negative transfers and the top performing target languages. This way, we observe top target languages that can get continuous improvement for each transfer language even after thousands of steps. 

\input{src_lang}
\input{ud_train_data}
\input{ner_train_data}
\input{evald}
\input{transfer_graphs}

\section{FAQ}
\begin{enumerate}
    \item What are the main contributions of this study and the difference of our approach with other methods?
    \begin{itemize}
        \item First, note that our paper introduces a method for studying cross-lingual transfer, not necessarily a method for improving cross-lingual transfer. We deviate from this “standard” way of using adapters for two reasons:
        \begin{enumerate}
            \item Training a task adapter on many languages, as a preliminary step, allows this component to learning the \texttt{task}, regardless of language. This is necessary for disentangling the effect of task and language in our analysis. 
            \item We then finetune the whole model (and not introduce a new adapter) exactly because we now want to study the effect of the language. While introducing a new language adapter might have a similar effect, there’s additional hurdles to do so: the language adapter would need more data to be trained, as it would be randomly initialized; our approach instead can work even with a single batch/update, so it is applicable even for very, very low-resource scenarios.
        \end{enumerate}
        \item Secondly, we propose a strategy to visually represent the language-language interaction utilizing the adapter-based fusion method. In general, training fully bilingual or trilingual for a different combination of languages are very expensive. This is why, we opt to have trained language adapter modules and then fuse together according to the need in an efficient manner. 
    \end{itemize}

    \item What is the reason for selecting the 38 transfer languages, including the 11 unseen languages? Why why include the 11 unseen languages from pre-training?
    \begin{itemize}
        \item \textbf{Language selection}: No other particular reasons except selecting a broader range of transfer languages covering language families and typological diversity. These 38 languages in total cover 10 language families, 26 genus and 14 script variations.
    \end{itemize}
    
\end{enumerate}

\end{document}

%% file: donors.tex
\begin{table}[t]
    \centering
    \small
\begin{tabular}{@{}l@{ }|l@{ }c@{ }r|l@{ }c@{ }r|l@{ }c@{ }r@{}}
\toprule
{} &     Lang. &  \texttt{ts} & +(\%) &       Lang. &  \texttt{ts} &  +(\%) &    Lang. &  \texttt{ts} &  +(\%) \\
\midrule
& \multicolumn{3}{c}{\textbf{Parsing}} & \multicolumn{3}{c}{\textbf{POS Tagging}} & \multicolumn{3}{c}{\textbf{NER}} \\
1  &          mya &     0.33 &     40.4 &  \textbf{kin} &     0.41 &     35.1 &          zho &     0.16 &      49.6 \\
2  &          ell &     0.15 &     31.6 &  \textbf{kmr} &     0.36 &     36.9 &          tel &     0.08 &      32.8 \\
3  &  \textbf{kmr} &     0.14 &     35.9 &  \textbf{mos} &     0.27 &     34.2 &          hun &     0.08 &      40.8 \\
4  &          yor &     0.14 &     33.3 &          hye &     0.27 &     36.9 &          heb &     0.04 &      34.4 \\
5  &  \textbf{pcm} &     0.13 &     31.6 &          cym &     0.22 &     37.7 &          est &     0.03 &      36.8 \\
\toprule
& \multicolumn{3}{c}{\textbf{XNLI}} & \multicolumn{3}{c}{\textbf{ANLI}} & \multicolumn{3}{c}{\textbf{TyDiQA}} \\
1  &  \textbf{hau} &     -34.4 &        0.0 &  \textbf{bam} &     -15.0 &        0.0 &          zho &         0.7 &         77.8 \\
2  &  \textbf{bam} &     -34.9 &        0.0 &  \textbf{hau} &     -17.8 &        0.0 &          jpn &         0.1 &         44.4 \\
3  &  \textbf{gub} &     -36.4 &        0.0 &  \textbf{gub} &     -18.4 &        0.0 &          gle &        -0.1 &         44.4 \\
4  &  \textbf{ewe} &     -36.7 &        0.0 &          deu &     -19.8 &        0.0 &  \textbf{wol} &        -0.1 &         44.4 \\
5  &          hin &     -37.1 &        0.0 &          fin &     -19.9 &        0.0 &          cym &        -0.1 &         33.3 \\
\bottomrule
\end{tabular}
    \caption{Top 5 transfer languages per task ranked using the aggregated transfer score (\texttt{ts} columns; see App.~\ref{sec:src_rank} for computation). Unseen ones are \textbf{bolded}. \textit{+(\%)} is the percentage of languages receiving positive transfer. No transfer language helps all target languages. (Complete rank with transfer scores: Table~\ref{tab:src_rank_t_mbert}-\ref{tab:src_rank_s_x}).}
    \label{tab:donors}
    \vspace{-1em}
\end{table}

\begin{table}[t]
    \centering
    \small
\begin{tabular}{@{}l|l@{ }c@{ }r|l@{ }c@{ }r}
\toprule
{} & \textbf{Parsing} & \textbf{Pos} & \textbf{NER} & \textbf{XNLI} & \textbf{ANLI} & \textbf{TyDiQA} \\
{} & &\textbf{Tagging} &  &  &  &  \\
\midrule
mBERT  &              30.6 &   31.0 &   31.8 &  0 &    0 &   30.1 \\
xlmr  &               20.5 &  33.2 &   41.1 & 44.4  &    41.6 &  17.0\\

\bottomrule
\end{tabular}
    \caption{Average percentage of languages receiving positive transfer (avg. \textit{+(\%))} across models. Unlike mBERT, xlmr provides positive transferring in NLI.}
    \label{tab:xlmr-mbert-nli}
    \vspace{-1em}
\end{table}

%% file: variance.tex
\begin{table}[]
    \centering
    \small
\begin{tabular}{l|c|lll}
\toprule
{Rank} &   Lang.  & \texttt{ts} & Var. & Type \\
{} &   \# (max, min)  &   &  \\
\midrule
1 & \textbf{ibo} (10, 10) & 0.05 & 23.5 & (+ \textit{and} -)\\
3 & \textbf{bam} (11, 15) & 0.02 & 21.5 & (+ \textit{and} -)\\
6 & \textbf{mos} (13, 2) & 0.09 & 16.1 & (+)\\
8 & \textbf{pcm} (1, 11) & 0.13 & 13.4 & (-)\\
26 & eng (0, 0) & -0.22 & 6.4 & neutral\\
36 & ara (0, 0) & -0.12 & 5.1  & neutral\\
\bottomrule
\end{tabular}
    \caption{Example of transfer languages ranked with their \texttt{aggregated-transfer (ts)} score variance (task: parsing). Unseen languages (\textbf{bold} font) exhibit high variance. \# (max) represents the language count receiving maximum positive transfer. (see Appendix \ref{sec:trans_var}) }
    \label{tab:variance}
    \vspace{-1em}
\end{table}

%% file: adapter_interference.tex
\begin{figure*}[t]
\small
    \centering
    \begin{tabular}{ccc}
         & (1) Bilingual  interactions \texttt{[AB]}& \\
        \includegraphics[width=.31\textwidth]{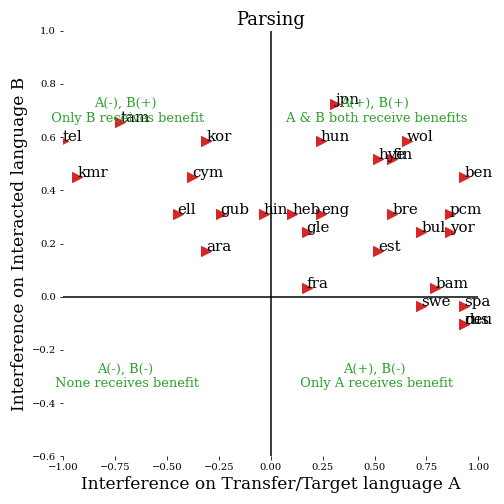}
        & \includegraphics[width=.31\textwidth]{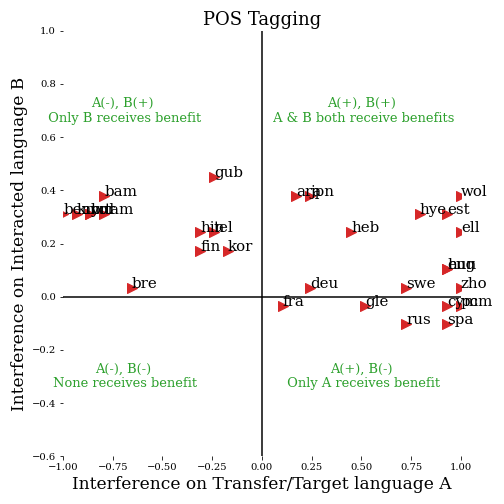}
        & \includegraphics[width=.31\textwidth]{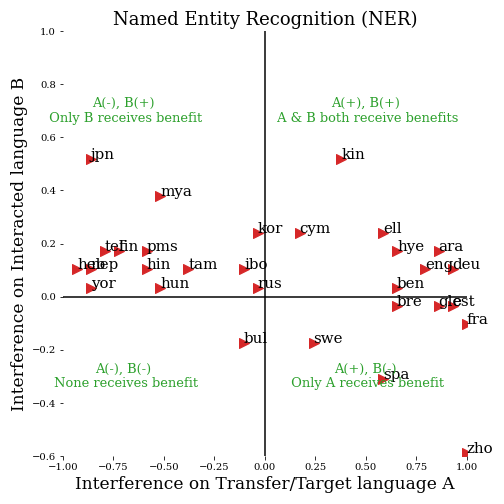}\\\\
        & (2) Trilingual  interactions \texttt{[ABC]} &\\
        \includegraphics[width=.31\textwidth]{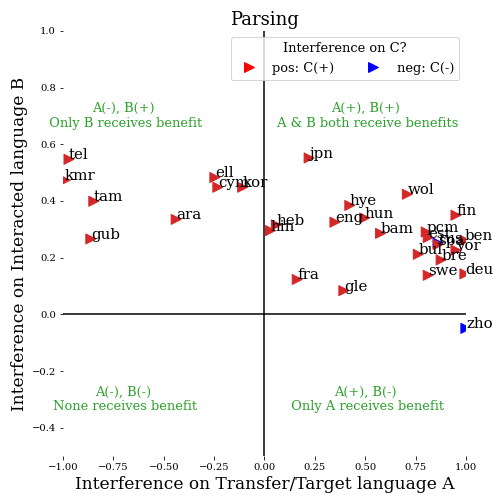}
        & \includegraphics[width=.31\textwidth]{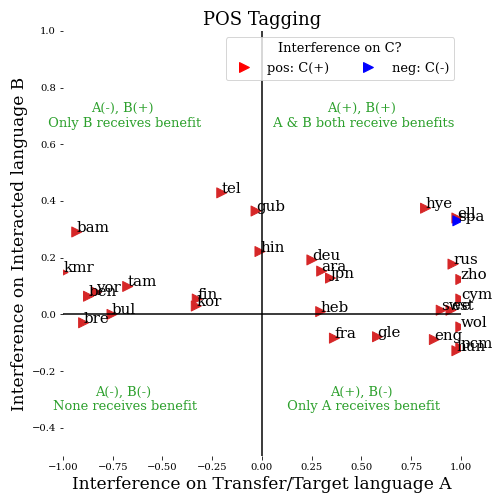}
        & \includegraphics[width=.31\textwidth]{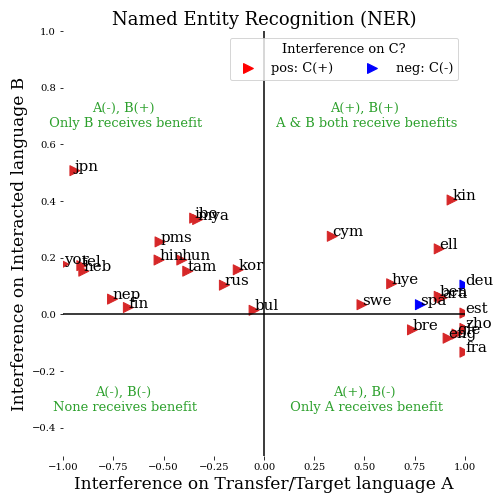}\\\
    \end{tabular}
    \vspace{-1em}
    \caption{Language interaction representation for bilingual and trilingual settings. To identify the language coordinates, we use two and three adapters (i.e. [2A], [3A]) jointly fused. In [3A] plots, we show the position only for one interacted language B along with transfer/target language A. For the 3rd language C, we use color variation (red/blue) to depict whether C receives positive transfer or not.}
    \label{fig:int_result}
\end{figure*}

%% file: example_int.tex
For example, consider the case of Arabic [A] that interacts with Bengali [B] in a bilingual setting [AB]. The count from pair combinations of positive and negative interference counts are as follows:

\begin{table}[]
    \centering
    \begin{tabular}{c|c}
    \toprule
       Combination  & Count \\
       \midrule
       $|-A,-B|$  & 1 \\
       $|-A,+B|$  & 1 \\
       $|+A,-B|$  & 3 \\
       $|+A,+B|$  & 2 \\
       \bottomrule
    \end{tabular}
    \caption{Bilingual interaction counts}
    \label{tab:my_label}
\end{table}

So for language A we get,
\begin{align*}
\mathtt{C} =& 1+1+3+2\\
           =& 7\\
\mathtt{(x_A, y_A)} =&\mathtt{\frac{1}{7}}\times (1\cdot[-1,-1]+1\cdot[-1,1]\\&+3\cdot[1,-1]+2\cdot[1,1])\\
=& (0.43, -0.14)
\end{align*}

Here, $|+A,+B|=2$ means, in total two cases, Arabic gets positive interference score while the other associated language (Bengali) also gets positive interference. Similarly, $|-A,-B|=1$ means, for one language, both Arabic and Bengali get negative interference scores. Now $(X_{A}, Y_{A})=(0.43,-0.14 )$. So Arabic will be in the lower-right quartile of the graph (+x, -y) means, Arabic generally gets positive interference but it does not equally beneficial to other languages (gets penalized for cases $|-A,-B|, |+A,-B|$). Here we consider only Bengali as a language to interact with. In practice, we use a set of other transfer languages to compute the total count of each combination for one specific language.

%% file: comp_score.tex
\begin{table*}[]
    \centering
\begin{tabular}{l|r|rrrr|rrr|lll}
    \toprule
    \multicolumn{12}{c}{Dependency Parsing}\\
    \midrule
    &mBERT&\multicolumn{4}{c|}{Continious Steps}&\multicolumn{3}{c|}{Lang. Interaction} & \multicolumn{3}{c}{Improvement}\\
          lang &  base &     1 &    10 &   100 &  1000 &    \texttt{[1A]} &    \texttt{[2A]} &    \texttt{[3A]} & \texttt{$Imp_{c:1}$} & \texttt{$Imp_{c:1000}$} & \texttt{$Imp_{i}$} \\
    \midrule
           ell & 92.82 & 92.73 & 92.39 & 92.09 & 91.46 & 91.97 & 91.91 & 91.98 &   no &        no &  no \\
           tel & 90.15 & 89.33 & 89.43 & 87.92 & 85.01 & 89.74 & 89.44 & 89.46 &   no &        no &  no \\
           spa & 90.02 & 89.87 & 89.42 & 89.00 & 88.16 & 89.09 & 89.20 & 89.24 &   no &        no &  no \\
           hin & 89.04 & 88.77 & 88.35 & 87.76 & 87.09 & 88.28 & 88.22 & 88.29 &   no &        no &  no \\
           hun & 87.20 & 87.14 & 86.49 & 85.65 & 85.06 & 86.25 & 86.29 & 86.36 &   no &        no &  no \\
\bottomrule

    \end{tabular}
    \caption{Dependency Parsing results. Improvement: \texttt{$Imp_{c:1}$} cont. step 1-base. \texttt{$Imp_{c:1000}$}: cont. steps 1000-base. \texttt{$Imp_{i}$}: improvement in language interactions (\texttt{[1A]/[2A]/[3A]}) versus baseline. Languages unseen by mBERT are in \textbf{bold} font.}
    \label{tab:comp_score_udp}
\end{table*}

\begin{table*}[]
    \centering
    \begin{tabular}{l|rrrrr|rrrrr}
    \toprule
    \multicolumn{11}{c}{Dependency Parsing (Model Comparison)}\\
    \midrule
    &\multicolumn{5}{c|}{mBERT}&\multicolumn{5}{c}{XLM-R} \\
    \midrule
    lang &  base &     1 &    10 &   100 &  1000 &  base\_x &   1\_x &  10\_x &  100\_x &  1000\_x \\
        \midrule
 ell & 92.82 & 92.73 & 92.39 & 92.09 & 91.46 &   93.33 & 93.27 & 93.01 &  92.59 &   92.86 \\
 tel & 90.15 & 89.33 & 89.43 & 87.92 & 85.01 &   88.49 & 87.88 & 86.34 &  85.99 &   86.41 \\
 spa & 90.02 & 89.87 & 89.42 & 89.00 & 88.16 &   89.66 & 89.70 & 89.38 &  88.92 &   89.26 \\
 deu & 89.37 & 89.40 & 88.67 & 88.39 & 87.43 &   89.71 & 89.63 & 89.09 &  88.74 &   89.30 \\
 bul & 89.32 & 89.32 & 89.01 & 89.01 & 89.09 &   90.19 & 90.31 & 90.22 &  89.73 &   90.19 \\
\bottomrule

    \end{tabular}
    \caption{Dependency Parsing results comparison using mBERT and XLM-R (for languages present in both transfer and target set.)}
    \label{tab:comp_score_udp_mbert_xlmr}
\end{table*}

\begin{table*}[]
    \centering
    \begin{tabular}{l|r|rrrr|rrr|lll}
    \toprule
    \multicolumn{12}{c}{POS Tagging}\\
    \midrule
    &mBERT&\multicolumn{4}{c|}{Continious Steps}&\multicolumn{3}{c|}{Lang. Interaction} & \multicolumn{3}{c}{Improvement}\\
          lang &  base &     1 &    10 &   100 &  1000 &    \texttt{[1A]} &    \texttt{[2A]} &    \texttt{[3A]} & \texttt{$Imp_{c:1}$} & \texttt{$Imp_{c:1000}$} & \texttt{$Imp_{i}$} \\
    \midrule
       spa & 97.84 & 97.80 & 97.66 & 97.60 & 97.41 & 97.67 & 97.71 & 97.72 &   no &        no &  no \\
       ell & 97.25 & 97.16 & 97.17 & 96.99 & 96.87 & 97.02 & 97.09 & 97.10 &   no &        no &  no \\
       heb & 95.22 & 95.07 & 94.91 & 94.72 & 94.52 & 94.86 & 94.89 & 94.89 &   no &        no &  no \\
       swe & 95.17 & 95.04 & 95.00 & 94.92 & 94.90 & 94.97 & 95.01 & 95.02 &   no &        no &  no \\
       hun & 94.28 & 94.17 & 94.03 & 93.94 & 93.82 & 93.91 & 94.01 & 94.03 &   no &        no &  no \\
\bottomrule

    \end{tabular}
    \caption{POS Tagging results. Improvement: \texttt{$Imp_{c:1}$} cont. step 1-base. \texttt{$Imp_{c:1000}$}: cont. steps 1000-base. \texttt{$Imp_{i}$}: improvement in language interactions (\texttt{[1A]/[2A]/[3A]}) versus baseline. Languages unseen by mBERT are in \textbf{bold} font.}
    \label{tab:comp_score_pos}
\end{table*}

\begin{table*}[]
    \centering
    \begin{tabular}{l|rrrrr|rrrrr}
    \toprule
    \multicolumn{11}{c}{POS Tagging (Model Comparison)}\\
    \midrule
    &\multicolumn{5}{c|}{mBERT}&\multicolumn{5}{c}{XLM-R} \\
    \midrule
    lang &  base &     1 &    10 &   100 &  1000 &  base\_x &   1\_x &  10\_x &  100\_x &  1000\_x \\
        \midrule
 spa & 97.84 & 97.80 & 97.66 & 97.60 & 97.41 &   97.80 & 97.78 & 97.74 &  97.65 &   97.61 \\
 ell & 97.25 & 97.16 & 97.17 & 96.99 & 96.87 &   97.51 & 97.50 & 97.47 &  97.43 &   97.45 \\
 bul & 96.12 & 96.13 & 96.07 & 96.08 & 96.12 &   96.79 & 96.75 & 96.68 &  96.58 &   96.58 \\
 heb & 95.22 & 95.07 & 94.91 & 94.72 & 94.52 &   96.41 & 96.30 & 96.16 &  96.16 &   96.24 \\
 swe & 95.17 & 95.04 & 95.00 & 94.92 & 94.90 &   96.22 & 96.20 & 96.29 &  96.26 &   96.17 \\
\bottomrule

    \end{tabular}
    \caption{POS Tagging results comparison using mBERT and XLM-R (for languages present in both transfer and target set.)}
    \label{tab:comp_score_pos_mbert_xlmr}
\end{table*}

\begin{table*}[]
    \centering
    \begin{tabular}{l|r|rrrr|rrr|lll}
    \toprule
    \multicolumn{12}{c}{NER}\\
    \midrule
    &mBERT&\multicolumn{4}{c|}{Continious Steps}&\multicolumn{3}{c|}{Lang. Interaction} & \multicolumn{3}{c}{Improvement}\\
          lang &  base &     1 &    10 &   100 &  1000 &    \texttt{[1A]} &    \texttt{[2A]} &    \texttt{[3A]} & \texttt{$Imp_{c:1}$} & \texttt{$Imp_{c:1000}$} & \texttt{$Imp_{i}$} \\
    \midrule
       spa & 90.66 & 90.56 & 90.18 & 89.41 & 87.74 & 90.20 & 90.26 & 90.26 &   no &        no &  no \\
       bul & 89.40 & 89.28 & 89.07 & 88.62 & 87.54 & 89.09 & 89.09 & 89.10 &   no &        no &  no \\
       fra & 88.14 & 87.95 & 87.56 & 86.53 & 85.19 & 87.63 & 87.77 & 87.80 &   no &        no &  no \\
       fin & 87.63 & 87.60 & 87.44 & 87.15 & 86.43 & 87.54 & 87.46 & 87.46 &   no &        no &  no \\
       est & 87.40 & 87.27 & 86.93 & 86.28 & 85.36 & 86.92 & 87.04 & 87.10 &   no &        no &  no \\
\bottomrule

    \end{tabular}
    \caption{NER results. Improvement: \texttt{$Imp_{c:1}$} cont. step 1-base. \texttt{$Imp_{c:1000}$}: cont. steps 1000-base. \texttt{$Imp_{i}$}: improvement in language interactions (\texttt{[1A]/[2A]/[3A]}) versus baseline. Languages unseen by mBERT are in \textbf{bold} font.}
    \label{tab:comp_score_ner}
\end{table*}

\begin{table*}[]
    \centering
    \begin{tabular}{l|rrrrr|rrrrr}
    \toprule
    \multicolumn{11}{c}{NER (Model Comparison)}\\
    \midrule
    &\multicolumn{5}{c|}{mBERT}&\multicolumn{5}{c}{XLM-R} \\
    \midrule
    lang &  base &     1 &    10 &   100 &  1000 &  base\_x &   1\_x &  10\_x &  100\_x &  1000\_x \\
        \midrule
 swe & 91.41 & 91.47 & 91.25 & 90.86 & 90.06 &   89.83 & 89.91 & 89.96 &  89.71 &   89.61 \\
 spa & 90.66 & 90.56 & 90.18 & 89.41 & 87.74 &   87.37 & 87.44 & 87.44 &  87.21 &   87.13 \\
 hye & 90.10 & 90.59 & 90.23 & 87.55 & 83.25 &   89.85 & 89.77 & 89.56 &  89.98 &   89.88 \\
 bul & 89.40 & 89.28 & 89.07 & 88.62 & 87.54 &   87.63 & 87.70 & 87.66 &  87.49 &   87.50 \\
 hun & 88.42 & 88.49 & 88.30 & 87.49 & 86.78 &   86.59 & 86.58 & 86.26 &  86.29 &   86.07 \\
\bottomrule

    \end{tabular}
    \caption{NER results comparison using mBERT and XLM-R (for languages present in both transfer and target set.)}
    \label{tab:comp_score_ner_mbert_xlmr}
\end{table*}

\begin{table*}[]
    \centering
    \begin{tabular}{l|r|rrrr}
    \toprule
    \multicolumn{6}{c}{XNLI}\\
    \midrule
    &&\multicolumn{4}{c}{Continious Steps}\\
          lang &  base &     1 &    10 &   100 &  1000  \\
    \midrule
    \multicolumn{6}{c}{mBERT}\\
\midrule
 eng & 81.02 & 45.42 & 45.01 & 42.86 & 43.18 \\
 spa & 77.33 & 46.59 & 47.52 & 44.87 & 40.96 \\
 deu & 76.27 & 46.79 & 47.71 & 46.63 & 39.57 \\
 zho & 75.43 & 46.45 & 44.74 & 43.23 & 40.07 \\
 bul & 75.13 & 46.57 & 46.70 & 44.85 & 36.58 \\
 ell & 73.89 & 44.51 & 44.38 & 44.21 & 36.28 \\
 rus & 73.59 & 45.78 & 46.28 & 44.24 & 38.53 \\
 ara & 71.36 & 42.82 & 41.44 & 41.37 & 39.34 \\
 hin & 67.68 & 41.74 & 42.61 & 40.19 & 35.02 \\

 \midrule
    \multicolumn{6}{c}{XLM-R}\\
\midrule
 eng & 84.03 & 83.91 & 83.87 & 83.38 & 83.54 \\
 spa & 80.60 & 80.67 & 80.92 & 80.26 & 80.38 \\
 bul & 80.24 & 80.24 & 80.16 & 79.72 & 80.45 \\
 deu & 79.38 & 79.33 & 79.32 & 78.79 & 79.20 \\
 rus & 78.10 & 78.20 & 78.43 & 78.05 & 78.05 \\
 ell & 77.82 & 77.77 & 77.61 & 77.22 & 77.59 \\
 zho & 77.41 & 77.44 & 77.33 & 77.49 & 77.37 \\
 ara & 75.63 & 75.47 & 75.15 & 74.55 & 74.91 \\
 hin & 74.81 & 74.67 & 74.35 & 74.02 & 74.55 \\
\bottomrule
     \end{tabular}
    \caption{XNLI results for (continuous training) languages present in both transfer and target set.}
    \label{tab:comp_score_xnli}
\end{table*}

\begin{table*}[]
    \centering
    \begin{tabular}{l|r|rrrr}
    \toprule
    \multicolumn{6}{c}{TyDiQA}\\
    \midrule
    &&\multicolumn{4}{c}{Continious Steps}\\
          lang &  base &     1 &    10 &   100 &  1000  \\
    \midrule
    \multicolumn{6}{c}{mBERT}\\
\midrule
 tel & 58.45 & 58.19 & 57.53 & 56.55 & 56.50 \\
 eng & 56.14 & 55.89 & 55.91 & 53.05 & 53.89 \\
 ara & 54.83 & 54.73 & 54.40 & 49.28 & 50.99 \\
 rus & 50.37 & 49.93 & 49.15 & 36.16 & 43.74 \\
 fin & 50.13 & 50.09 & 50.45 & 44.16 & 46.85 \\
 kor & 47.83 & 46.59 & 46.74 & 44.09 & 44.38 \\
 ben & 45.13 & 46.11 & 48.05 & 45.13 & 44.78 \\

 \midrule
    \multicolumn{6}{c}{XLM-R}\\
\midrule
 tel & 56.20 & 56.46 & 55.65 & 55.72 & 56.35 \\
 eng & 52.50 & 52.82 & 53.18 & 52.89 & 52.70 \\
 ara & 51.57 & 51.69 & 49.16 & 48.02 & 51.13 \\
 rus & 47.41 & 47.32 & 45.04 & 44.19 & 46.26 \\
 fin & 45.65 & 46.24 & 45.88 & 45.10 & 45.19 \\
 ben & 44.25 & 42.39 & 43.27 & 40.97 & 43.45 \\
 kor & 42.03 & 42.32 & 42.90 & 43.01 & 43.22 \\
\bottomrule
     \end{tabular}
    \caption{TyDiQA results (continuous training)for languages present in both transfer and target set.}
    \label{tab:comp_score_TyDiQA}
\end{table*}

%% file: src_rank.tex
\begin{table*}[]
    \centering
    \begin{tabular}{l|lrr|lrr|lrr}
\toprule
\multicolumn{10}{c}{Transfer Languages Ranking using mBERT (Token Classification)}\\
\midrule
{Rank} & \multicolumn{3}{c}{\textbf{Parsing}} & \multicolumn{3}{c}{\textbf{POS Tagging}} & \multicolumn{3}{c}{\textbf{NER}} \\
{} &       Lang &  \texttt{ts} & +(\%) &       lang &  \texttt{ts} &  +(\%) &       lang &  \texttt{ts} &  +(\%) \\
\midrule
1  &          mya &     0.33 &     40.35 &  \textbf{kin} &     0.41 &     35.09 &          zho &     0.16 &      49.6 \\
2  &          ell &     0.15 &     31.58 &  \textbf{kmr} &     0.36 &     36.84 &          tel &     0.08 &      32.8 \\
3  &  \textbf{kmr} &     0.14 &     35.96 &  \textbf{mos} &     0.27 &     34.21 &          hun &     0.08 &      40.8 \\
4  &          yor &     0.14 &     33.33 &          hye &     0.27 &     36.84 &          heb &     0.04 &      34.4 \\
5  &  \textbf{pcm} &     0.13 &     31.58 &          cym &     0.22 &     37.72 &          est &     0.03 &      36.8 \\
\bottomrule
    \end{tabular}
    \caption{Transfer Languages ranked by aggregated transfer scores (\texttt{ts}) overall target languages across token classification tasks using mBERT. Languages unseen by mBERT are in \textbf{bold} font.}
    \label{tab:src_rank_t_mbert}
\end{table*}

\begin{table*}[]
    \centering
    \begin{tabular}{l|lrr|lrr|lrr}
\toprule
\multicolumn{10}{c}{Transfer Languages Ranking using XLM-R (Token Classification)}\\
\midrule
{Rank} & \multicolumn{3}{c}{\textbf{Parsing}} & \multicolumn{3}{c}{\textbf{POS Tagging}} & \multicolumn{3}{c}{\textbf{NER}} \\
{} &       Lang &  \texttt{ts} & +(\%) &       lang &  \texttt{ts} &  +(\%) &       lang &  \texttt{ts} &  +(\%) \\
\midrule
1  &          nep &     0.02 &     23.68 &          hin &     0.18 &     39.47 &          rus &     1.15 &      40.0 \\
2  &          zho &    -0.00 &     34.21 &          ben &     0.10 &     36.84 &          ell &     1.04 &      41.6 \\
3  &          mya &    -0.01 &     21.93 &          mya &     0.04 &     34.21 &          tel &     0.76 &      41.6 \\
4  &          ben &    -0.08 &     26.32 &          nep &     0.01 &     34.21 &          heb &     0.70 &      46.4 \\
5  &          hin &    -0.09 &     29.82 &          bul &     0.01 &     40.35 &          ben &     0.64 &      36.0 \\
\bottomrule
    \end{tabular}
    \caption{Transfer Languages ranked by aggregated transfer scores (\texttt{ts}) overall target languages across token classification tasks using XLM-R. Languages unseen by mBERT are in \textbf{bold} font.}
    \label{tab:src_rank_t_x}
\end{table*}

\begin{table*}[]
    \centering
    \begin{tabular}{l|lrr|lrr|lrr}
\toprule
\multicolumn{10}{c}{Transfer Languages Ranking using mBERT (Sentence Classification \& QA)}\\
\midrule
{Rank} & \multicolumn{3}{c}{\textbf{XNLI}} & \multicolumn{3}{c}{\textbf{ANLI}} & \multicolumn{3}{c}{\textbf{TyDiQA}} \\
{} &       Lang &  \texttt{ts} & +(\%) &       lang &  \texttt{ts} &  +(\%) &       lang &  \texttt{ts} &  +(\%) \\
\midrule
1  &  \textbf{hau} &    -34.42 &        0.0 &  \textbf{bam} &    -14.97 &        0.0 &          zho &        0.67 &        77.78 \\
2  &  \textbf{bam} &    -34.85 &        0.0 &  \textbf{hau} &    -17.82 &        0.0 &          jpn &        0.08 &        44.44 \\
3  &  \textbf{gub} &    -36.40 &        0.0 &  \textbf{gub} &    -18.35 &        0.0 &          gle &       -0.08 &        44.44 \\
4  &  \textbf{ewe} &    -36.73 &        0.0 &          deu &    -19.79 &        0.0 &  \textbf{wol} &       -0.12 &        44.44 \\
5  &          hin &    -37.08 &        0.0 &          fin &    -19.93 &        0.0 &          cym &       -0.14 &        33.33 \\
\bottomrule
    \end{tabular}
    \caption{Transfer Languages ranked by aggregated transfer scores (\texttt{ts}) overall target languages across Sentence Classification \& QA tasks using mBERT. Languages unseen by mBERT are in \textbf{bold} font.}
    \label{tab:src_rank_s_m}
\end{table*}

\begin{table*}[]
    \centering
    \begin{tabular}{l|lrr|lrr|lrr}
\toprule
\multicolumn{10}{c}{Transfer Languages Ranking using mBERT (Sentence Classification \& QA)}\\
\midrule
{Rank} & \multicolumn{3}{c}{\textbf{XNLI}} & \multicolumn{3}{c}{\textbf{ANLI}} & \multicolumn{3}{c}{\textbf{TyDiQA}} \\
{} &       Lang &  \texttt{ts} & +(\%) &       lang &  \texttt{ts} &  +(\%) &       lang &  \texttt{ts} &  +(\%) \\
\midrule
1  &  \textbf{ewe} &       0.44 &       93.33 &          hin &       2.25 &       100.0 &  \textbf{pcm} &        -0.42 &         55.56 \\
2  &          bre &       0.36 &       93.33 &          ell &       1.34 &        80.0 &          ell &        -0.44 &         22.22 \\
3  &  \textbf{bam} &       0.30 &       93.33 &          nep &       1.33 &       100.0 &          fin &        -0.46 &         44.44 \\
4  &  \textbf{pcm} &       0.30 &       93.33 &          ara &       1.31 &        80.0 &          zho &        -0.47 &         44.44 \\
5  &  \textbf{ibo} &       0.28 &       80.00 &          swe &       1.31 &       100.0 &          heb &        -0.47 &         11.11 \\
\bottomrule
    \end{tabular}
    \caption{Transfer Languages ranked by aggregated transfer scores (\texttt{ts}) overall target languages across Sentence Classification \& QA tasks using XLM-R. Languages unseen by mBERT are in \textbf{bold} font.}
    \label{tab:src_rank_s_x}
\end{table*}

%% file: receivers.tex
\begin{table*}[]
    \centering
    \begin{tabular}{p{2.5cm}|p{4.5cm}|p{3.5cm}|p{2.5cm}}
    \toprule
    Tasks & Never Receives & Positive Transfer (\%) (90-99] & Universal Recipient\\
    \midrule
        Dependency Parsing & ell, isl, \textbf{grc}, \textbf{mlt}, fra, \textbf{qtd}, hrv, lav, \textbf{urb}, \textbf{fas}, ukr, spa, cym, tel, \textbf{pcm}, afr, swe, est, \textbf{nor}, \textbf{hsb}, \textbf{orv}, cat, slv, \textbf{chu}, \textbf{sme}, eus, slk, hye, \textbf{gla}, urd, hin, \textbf{fro}, lit, lat, \textbf{san}, \textbf{wol}, \textbf{lij}, \textbf{got}, srp, \textbf{lzh} &
        \textbf{tpn}, \textbf{koi}, \textbf{glv}, kor, \textbf{krl}, \textbf{mdf}, jpn, \textbf{amh}, sqi
        &
        \textbf{mpu}, \textbf{gun}, \textbf{nap}
        \\\midrule
        POS Tagging & ell, nld, isl, \textbf{grc}, lav, \textbf{fas}, ukr, spa, cym, hun, \textbf{pcm}, afr, swe, est, \textbf{nor}, \textbf{hsb}, tam, cat, \textbf{chu}, \textbf{sme}, eus, hye, urd, ben, ita, ron, lit, lat, \textbf{wol}, \textbf{got}, srp, \textbf{lzh} & 
        
        fra, eng, \textbf{qhe}, dan, \textbf{myu}, \textbf{glv}, kor, tur, kaz, \textbf{akk}, \textbf{myv}, \textbf{nap}
        \\\midrule
        
        NER & nld, ell, tgl, fra, ces, bul, zho, msa, sun, lav, gle, \textbf{fas}, kat, spa, heb, hbs, afr, est, \textbf{yid}, eng, tam, bre, vie, jpn, cat, \textbf{tha}, slv, ceb, tur, mlg, slk, swa, ben, uzb, ita, ron, tat, pol, \textbf{zea}, \textbf{lin}, \textbf{ibo} & 
        
        \textbf{ksh}, \textbf{pms}, aze, \textbf{mzn}, \textbf{oci}, tgk, \textbf{roh}, \textbf{khm}, \textbf{aym}, \textbf{csb}
        &
        bak, ast, \textbf{uig}, kaz, nds, \textbf{amh}\\\bottomrule
    \end{tabular}
    \caption{We find 25 languages out of 40 which receives positive transfer from almost any transfer languages (i.e. column 90-99\% and 100\%) are unseen by mbert. (language codes in \textbf{bold} font are the unseen ones)}
    \label{tab:receive}
\end{table*}

%% file: all_max.tex
\begin{table*}[]
    \centering
\begin{tabular}{c|llll|llll|llll}
\toprule
{Transfer} & \multicolumn{4}{c}{\textbf{Parsing}} & \multicolumn{4}{c}{\textbf{POS Tagging}} & \multicolumn{4}{c}{\textbf{NER}} \\
 Language  &   1 &            10 &            100 &            1000 &   1 &            10 &            100 &            1000 &   1 &            10 &            100 &            1000 \\
\midrule
gub & mpu &          mpu &          gun & \textbf{gub} & aqz &          nap &          nap & \textbf{nap} & amh &          amh & \textbf{amh} &          bar \\
est & nap &          tpn &          gun & \textbf{gun} & nap &          nap &          nap & \textbf{nap} & amh &          amh & \textbf{amh} &          som \\
bre & nap &          nap &          gun & \textbf{gun} & nap &          nap &          nap & \textbf{nap} & amh &          amh & \textbf{amh} &          amh \\
eng & nap &          mpu &          gun & \textbf{gun} & nap &          nap &          nap & \textbf{nap} & amh &          sin & \textbf{amh} &          nds \\
ben & nap &          nap &          nap & \textbf{gun} & cop &          nap &          nap & \textbf{nap} & amh &          sin &          amh & \textbf{amh} \\
kmr & nap &          mpu &          gun & \textbf{kmr} & urb &          nap &          nap & \textbf{nap} & amh &          amh & \textbf{amh} &          amh \\
spa & nap &          mpu &          gun & \textbf{gun} & nap &          nap & \textbf{nap} &          nap & amh &          amh & \textbf{amh} &          roh \\
bul & nap &          nap &          nap & \textbf{gun} & nap &          nap &          nap & \textbf{amh} & amh &          amh & \textbf{amh} &          som \\
pms & nap & \textbf{nap} &          nap &          mpu & nap &          nap &          nap & \textbf{nap} & amh &          amh & \textbf{amh} &          amh \\
gle & nap &          nap &          mpu & \textbf{gun} & aqz &          nap &          nap & \textbf{nap} & amh &          amh & \textbf{amh} &          som \\
nep & nap &          tpn &          gun & \textbf{gun} & aqz &          urb &          nap & \textbf{nap} & amh &          sin & \textbf{amh} &          roh \\
cym & nap &          nap &          gun & \textbf{gun} & nap &          nap &          nap & \textbf{amh} & amh & \textbf{sin} &          amh &          som \\
fin & nap &          nap &          tpn & \textbf{gun} & nap &          nap &          nap & \textbf{nap} & amh &          sin & \textbf{amh} &          som \\
hye & nap &          nap &          nap & \textbf{gun} & nap &          nap &          nap & \textbf{nap} & uig &          sin & \textbf{amh} &          som \\
mya & nap &          nap &          wbp & \textbf{gun} & aqz &          urb &          nap & \textbf{nap} & amh &          amh & \textbf{amh} &          amh \\
hin & nap &          tpn &          gun & \textbf{gun} & aqz &          aqz &          nap & \textbf{nap} & amh &          amh & \textbf{amh} &          som \\
tel & nap &          nap &          gun & \textbf{gun} & aqz &          nap &          nap & \textbf{amh} & amh &          sin & \textbf{amh} &          roh \\
tam & nap &          nap &          gun & \textbf{gun} & nap &          nap &          nap & \textbf{nap} & amh &          sin & \textbf{amh} &          som \\
kor & tpn &          mpu &          mpu & \textbf{tpn} & nap &          nap &          nap & \textbf{nap} & amh &          amh & \textbf{amh} &          roh \\
ell & nap &          tpn &          nap & \textbf{gun} & nap &          nap &          nap & \textbf{nap} & amh &          amh & \textbf{amh} &          som \\
hun & nap &          mpu &          gun & \textbf{gun} & nap &          nap &          nap & \textbf{nap} & amh &          amh & \textbf{amh} &          som \\
heb & nap &          nap &          nap & \textbf{gun} & nap &          nap &          nap & \textbf{nap} & amh &          amh & \textbf{amh} &          som \\
zho & tpn &          nap &          nap & \textbf{mpu} & nap &          nap &          nap & \textbf{nap} & amh &          amh & \textbf{amh} &          amh \\
ara & nap &          nap &          gun & \textbf{gun} & nap &          nap &          nap & \textbf{nap} & uig &          amh & \textbf{amh} &          amh \\
swe & nap &          nap &          gun & \textbf{gun} & nap &          nap &          nap & \textbf{nap} & amh &          sin & \textbf{amh} &          som \\
jpn & nap &          mpu &          mpu & \textbf{tpn} & nap &          nap &          nap & \textbf{nap} & amh &          amh &          amh & \textbf{amh} \\
fra & nap &          mpu &          gun & \textbf{gun} & nap &          nap &          nap & \textbf{nap} & amh &          amh & \textbf{amh} &          som \\
deu & tpn &          mpu &          gun & \textbf{gun} & nap &          nap &          nap & \textbf{nap} & amh &          amh & \textbf{amh} &          som \\
rus & nap &          nap &          gun & \textbf{gun} & nap &          nap & \textbf{nap} &          nap & amh &          sin & \textbf{amh} &          roh \\
bam & mpu &          wbp & \textbf{wbp} &          gun & nap & \textbf{nap} &          amh &          bam & amh &          uig &          amh & \textbf{amh} \\
ewe & nap &          gun &          tpn & \textbf{gun} & nap & \textbf{nap} &          amh &          nap & amh &          sin & \textbf{amh} &          nds \\
hau & mpu &          nap &          gun & \textbf{gun} & aqz & \textbf{nap} &          nap &          amh & amh &          sin &          amh & \textbf{amh} \\
ibo & tpn &          mpu &          gun & \textbf{gun} & aqz & \textbf{nap} &          nap &          mpu & sin &          amh &          amh & \textbf{amh} \\
kin & mpu & \textbf{nap} &          nap &          tpn & nap &          nap &          nap & \textbf{nap} & amh &          amh & \textbf{amh} &          amh \\
mos & mpu &          aqz &          gun & \textbf{gun} & nap & \textbf{nap} &          amh &          nap & sin &          sin & \textbf{amh} &          amh \\
pcm & nap &          nap &          wbp & \textbf{gun} & kfm &          nap &          nap & \textbf{nap} & amh &          amh & \textbf{amh} &          amh \\
wol & nap &          aqz &          gun & \textbf{gun} & nap &          nap &          nap & \textbf{amh} & amh &          sin & \textbf{amh} &          nds \\
yor & nap &          nap &          wbp & \textbf{gun} & nap &          nap &          nap & \textbf{nap} & amh &          amh & \textbf{amh} &          som \\
\bottomrule
\end{tabular}

    \caption{Only bar and nds are seen by mbert. All other languages receiving maximum benefits continuously are unseen by mbert (kfm, urb, gun, aqz, cop, roh, bam, tpn, som, kmr, uig, mpu, amh, sin, wbp, gub, nap). The maximum score across different steps of training are \textbf{bolded}.}
    \label{tab:max_unseen}
\end{table*}

%% file: transfer_variance.tex
\begin{figure*}[t]
    \centering
    \begin{tabular}{l}
    (a)\\
\includegraphics[width=\textwidth]{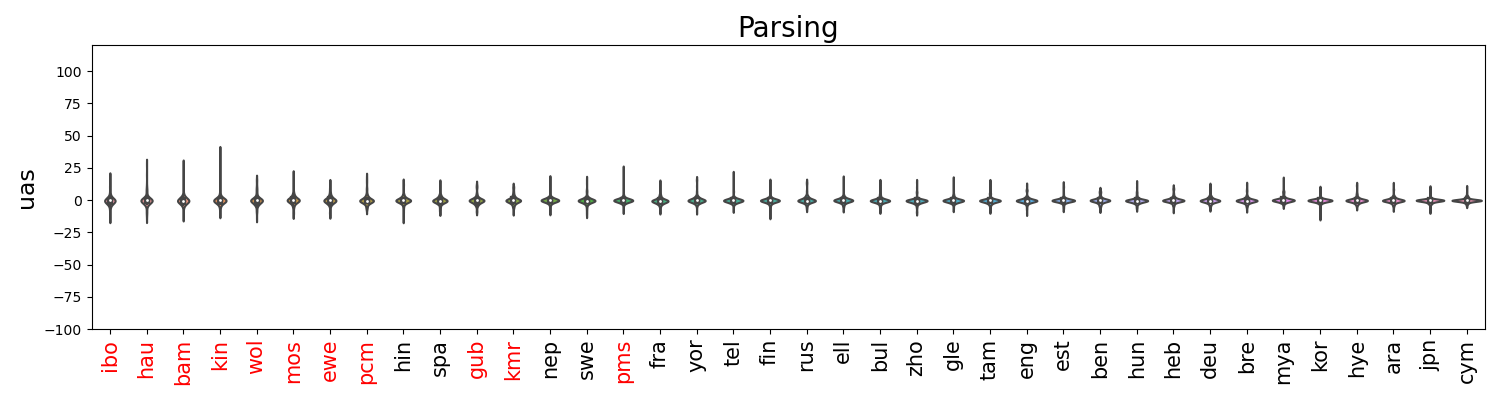} \\
(b)\\
\includegraphics[width=\textwidth]{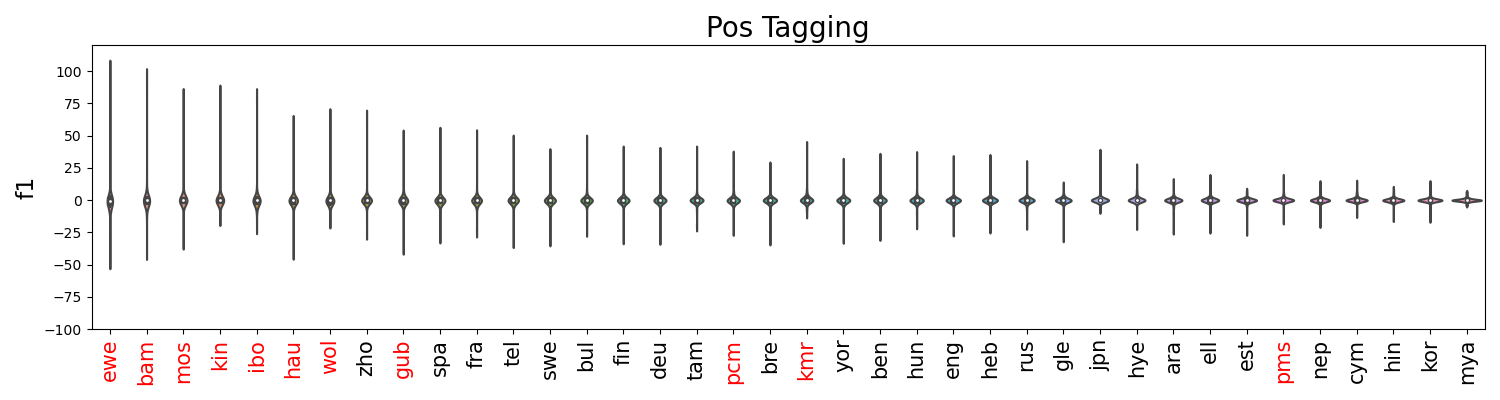} \\
(c)\\
\includegraphics[width=\textwidth]{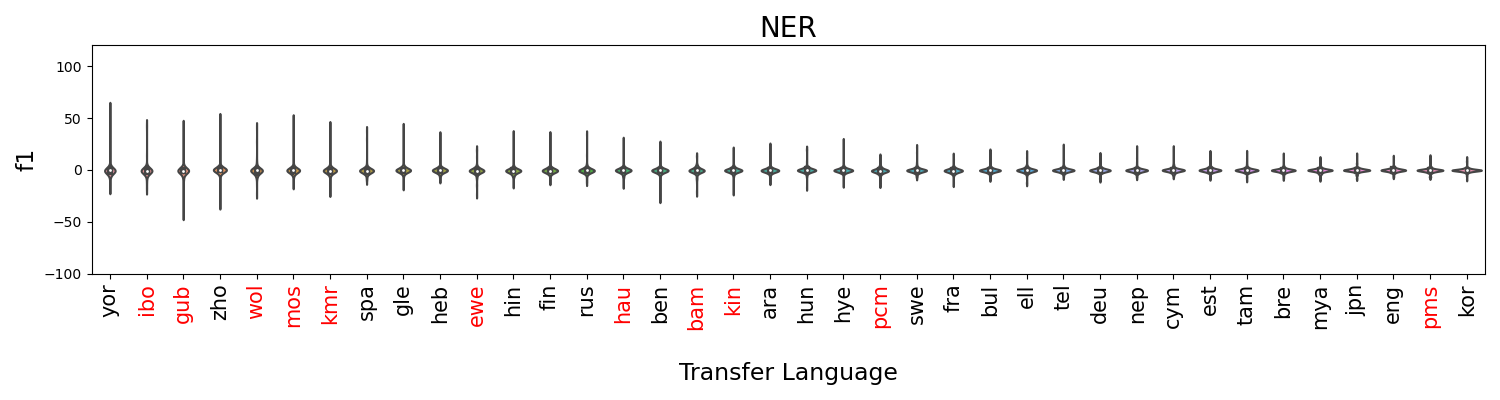} \\

    \end{tabular}
    \caption{Violin plots of transfer languages sorted by transfer score variance. mBERT unseen languages are in \textcolor{red}{red} color font.}
    \label{fig:trans_var}
\end{figure*}

\begin{table*}[t]

    \centering
    \small
    \begin{tabular}{l|lcc|lcc|lcc}
    \toprule
    {} & \multicolumn{3}{c}{\textbf{Parsing}} & \multicolumn{3}{c}{\textbf{POS Tagging}} & \multicolumn{3}{c}{\textbf{NER}} \\
    {Rank} &           Lang &    Transfer &  (Max, Min)  &          Lang &    Transfer &  (Max, Min)  &       Lang &    Transfer &  (Max, Min)   \\
    {} &            &    (Var.) &  \# &                &    (Var.) &  \# &                 &    (Var.) &  \#           \\
\midrule
1  &  \textbf{ibo} &   0.05 (23.5) &  (10, 10) &  \textbf{ewe} &  -0.79 (120.4) &  (5, 31) &         yor &  -0.32 (44.4) &   (2, 6) \\
2  &  \textbf{hau} &   0.04 (22.7) &    (2, 2) &  \textbf{bam} &  -0.12 (101.2) &   (5, 0) &  \textbf{ibo} &  -0.54 (42.0) &  (3, 18) \\
3  &  \textbf{bam} &   0.02 (21.5) &  (11, 15) &  \textbf{mos} &    0.27 (69.8) &   (4, 3) &  \textbf{gub} &  -0.98 (41.2) &  (3, 14) \\
4  &  \textbf{kin} &  -0.06 (21.4) &    (3, 1) &  \textbf{kin} &    0.41 (69.1) &   (3, 0) &         zho &   0.16 (32.7) &  (46, 3) \\
5  &  \textbf{wol} &   0.03 (17.6) &    (5, 6) &  \textbf{ibo} &   -0.14 (68.0) &  (5, 10) &  \textbf{wol} &  -0.41 (32.4) &   (5, 6) \\
6  &  \textbf{mos} &   0.09 (16.1) &   (13, 2) &  \textbf{hau} &   -0.34 (52.5) &   (1, 3) &  \textbf{mos} &    0.0 (29.5) &   (2, 1) \\
7  &  \textbf{ewe} &  -0.08 (14.5) &    (4, 5) &  \textbf{wol} &   -0.14 (51.9) &  (8, 10) &  \textbf{kmr} &  -0.61 (25.6) &   (3, 8) \\
8  &  \textbf{pcm} &   0.13 (13.4) &   (1, 11) &         zho &   -0.04 (45.6) &  (26, 7) &         spa &  -0.28 (24.5) &   (1, 1) \\
9  &         hin &   -0.1 (13.1) &    (4, 3) &  \textbf{gub} &   -0.34 (41.3) &   (3, 4) &         gle &  -0.07 (21.1) &   (1, 1) \\
10 &         spa &  -0.18 (11.5) &    (2, 3) &         spa &   -0.53 (39.6) &   (0, 2) &         heb &   0.04 (21.1) &   (3, 0) \\
11 &  \textbf{gub} &  -0.14 (10.7) &    (5, 3) &         fra &   -0.41 (37.2) &   (2, 6) &  \textbf{ewe} &  -0.76 (19.0) &   (0, 5) \\
12 &  \textbf{kmr} &   0.14 (10.2) &    (0, 1) &         tel &   -0.31 (36.6) &   (0, 1) &         hin &  -0.72 (18.6) &  (1, 12) \\
13 &         nep &   0.12 (10.1) &    (8, 1) &         swe &   -0.66 (32.2) &   (1, 3) &         fin &  -0.29 (18.3) &   (0, 1) \\
14 &         swe &  -0.21 (10.0) &    (0, 2) &         bul &   -0.08 (28.9) &   (0, 0) &         rus &  -0.34 (17.4) &   (0, 1) \\
15 &  \textbf{pms} &    0.09 (9.7) &    (3, 1) &         fin &   -0.35 (28.9) &   (0, 1) &  \textbf{hau} &  -0.07 (17.1) &   (7, 0) \\
16 &         fra &   -0.37 (9.1) &    (1, 8) &         deu &   -0.44 (26.0) &   (0, 0) &         ben &  -0.54 (17.0) &   (4, 3) \\
17 &         yor &    0.14 (8.9) &    (3, 3) &         tam &   -0.14 (22.9) &   (0, 0) &  \textbf{bam} &  -0.69 (16.1) &   (4, 1) \\
18 &         tel &    0.05 (8.7) &    (1, 2) &  \textbf{pcm} &   -0.16 (22.1) &  (3, 14) &  \textbf{kin} &  -0.56 (13.2) &   (1, 3) \\
19 &         fin &   -0.26 (8.6) &    (1, 1) &         bre &   -0.39 (21.4) &   (1, 0) &         ara &  -0.32 (13.1) &   (0, 1) \\
20 &         rus &    0.11 (8.5) &    (0, 1) &  \textbf{kmr} &    0.36 (21.2) &  (11, 6) &         hun &   0.08 (12.2) &   (2, 0) \\
21 &         ell &    0.15 (8.3) &    (1, 0) &         yor &   -0.16 (21.0) &   (1, 1) &         hye &  -0.17 (12.1) &   (0, 1) \\
22 &         bul &   -0.21 (6.9) &    (1, 1) &         ben &   -0.08 (20.6) &   (3, 2) &  \textbf{pcm} &  -0.69 (11.8) &  (6, 25) \\
23 &         zho &   -0.48 (6.9) &   (5, 10) &         hun &   -0.29 (20.4) &   (3, 1) &         swe &  -0.11 (10.1) &   (0, 0) \\
24 &         gle &    0.03 (6.8) &    (1, 1) &         eng &   -0.35 (18.8) &   (2, 1) &         fra &   -0.59 (9.5) &   (0, 5) \\
25 &         tam &   -0.24 (6.7) &    (1, 3) &         heb &   -0.26 (17.3) &   (0, 1) &         bul &   -0.18 (9.0) &   (0, 1) \\
26 &         eng &   -0.22 (6.4) &    (0, 0) &         rus &   -0.28 (15.3) &   (0, 0) &         ell &   -0.29 (8.8) &   (1, 1) \\
27 &         est &     0.0 (6.1) &    (1, 1) &         gle &   -0.34 (13.7) &   (0, 2) &         tel &    0.08 (8.6) &   (1, 0) \\
28 &         ben &   -0.06 (6.0) &    (3, 1) &         jpn &    0.18 (13.6) &  (11, 0) &         deu &    -0.2 (8.4) &   (0, 0) \\
29 &         hun &   -0.23 (5.9) &    (0, 2) &         hye &    0.27 (12.8) &   (2, 0) &         nep &   -0.13 (8.1) &   (1, 1) \\
30 &         heb &    0.08 (5.8) &    (0, 1) &         ara &    -0.3 (11.7) &   (0, 0) &         cym &    0.03 (7.8) &   (2, 1) \\
31 &         deu &   -0.13 (5.7) &    (1, 4) &         ell &   -0.13 (11.1) &   (1, 0) &         est &    0.03 (7.8) &   (0, 0) \\
32 &         bre &   -0.23 (5.7) &    (1, 1) &         est &    -0.35 (9.1) &   (1, 1) &         tam &     0.0 (7.1) &   (3, 0) \\
33 &         mya &    0.33 (5.6) &    (9, 0) &  \textbf{pms} &     0.08 (8.8) &   (2, 0) &         bre &   -0.23 (5.6) &   (1, 0) \\
34 &         kor &   -0.38 (5.6) &    (0, 1) &         nep &     0.12 (8.4) &   (6, 0) &         mya &   -0.17 (5.3) &   (5, 2) \\
35 &         hye &   -0.03 (5.4) &    (0, 2) &         cym &     0.22 (6.7) &   (0, 0) &         jpn &   -0.08 (5.3) &   (4, 1) \\
36 &         ara &   -0.12 (5.1) &    (0, 0) &         hin &    -0.13 (6.4) &   (0, 3) &         eng &    0.02 (5.1) &   (5, 1) \\
37 &         jpn &   -0.19 (3.9) &   (14, 2) &         kor &    -0.07 (5.8) &   (1, 0) &  \textbf{pms} &   -0.29 (5.0) &   (1, 0) \\
38 &         cym &   -0.03 (2.9) &    (1, 3) &         mya &     0.17 (2.7) &   (3, 1) &         kor &   -0.09 (4.4) &   (7, 1) \\
\bottomrule
    \end{tabular}
    \caption{Transfer languages are sorted by transfer score variance (mBERT unseen languages are in \textbf{bold} font). \# Max Transfer and \# Min Transfer denote the count of target languages which receive maximum and minimum transfer from this particular transfer language.}
    \label{tab:trans_var}
\end{table*}

\begin{table*}[]
    \centering
    \begin{tabular}{p{1.5cm}|p{1.8cm}|p{1.5cm}|p{4cm}|p{4cm}}
    \toprule
        Type & Transfer Language & Variance & max(+)$\rightarrow$ & min(-)$\rightarrow$\\
    \midrule
        (+ \textit{and} -) & ibo & high &aii, ajp, apu, arr, ces, gle, gub, koi, krl, yor
        & grc, hsb, hye, kfm, otk, san, sme, sqi, srp, urb\\\hline
        (+ \textit{and} -) & bam & high& bho, bam, bre, bxr, kfm, kmr, kpv, mpu, rus, soj, wbp
        &ajp, ara, chu, gla, got, krl, lzh, nld, orv, qpm, qtd, swl,
 tha, tgl, zho\\\hline
        (+) & mos & high& aqz, bel, bul, eng, ind, ita, kaz, lit, myv, pol, tam, tgl,
 ukr
  & arr, wbp\\\hline
        (-) & pcm & high& tha
  & aii, bho, ell, eng, eus, hrv, isl, lat, lit, nor, qhe\\\hline
  neutral & eng&low & -
  &-\\\hline
  neutral & ara&low & -
  &-\\
        
    \bottomrule
    \end{tabular}
    \caption{Characteristics of example transfer languages with different intensity of variance derived from dependency parsing task results. max(+)$\rightarrow$ represents set of target language which receive maximum score for the specific transfer language wheres, min(-)$\rightarrow$ represents the complete oposite.}
    \label{tab:select_var}
\end{table*}

%% file: src_lang.tex
\begin{table*}[]
    \centering
\begin{tabular}{llllll}
\toprule
                  Language & iso-639 &        Family &               Genus &      Script & mBERT-seen? \\
\midrule
                Hebrew &   heb &  Afro-Asiatic &             Semitic &        Hebr &   \\
                Arabic &   ara &  Afro-Asiatic &             Semitic &        Arab &   \\
                 Hausa &   hau &  Afro-Asiatic &         West Chadic &        Latn & \tikzxmark \\
                 \midrule
                Telugu &   tel &     Dravidian &           Dravidian &        Telu &   \\
                 Tamil &   tam &     Dravidian &           Dravidian &        Taml &   \\
                 \midrule
              Armenian &   hye & Indo-European &            Armenian &        Armn &   \\
                Breton &   bre & Indo-European &              Celtic &        Latn &   \\
\bottomrule
\end{tabular}
    \caption{Transfer Languages we use in our study for mBERT fine-tuning}
    \label{tab:src_lang}
\end{table*}

%% file: ud_train_data.tex
\onecolumn
\small
\begin{longtable}{lllllll}
\toprule
\multicolumn{7}{c}{\textbf{UDP and POS Tagging Task Adapter Training Dataset}} \\\hline\\
                     Language & iso-639 &            UD Identifier &   \# Examples &         Family &               Genus &    Script \\
\midrule
                   Coptic &   cop & cop\_scriptorium &   403 &   Afro-Asiatic &     Egyptian-Coptic &    Coptic \\
                   Arabic &   ara &        ar\_nyuad &  1963 &   Afro-Asiatic &             Semitic &      Arab \\
                   Hebrew &   heb &          he\_htb &   491 &   Afro-Asiatic &             Semitic &      Hebr \\
                  Maltese &   mlt &         mt\_mudt &   518 &   Afro-Asiatic &             Semitic &      Latn \\
                   Kazakh &   kaz &          kk\_ktb &  1047 &         Altaic &              Turkic &      Cyrl \\
\end{longtable}   
\twocolumn
\normalsize

%% file: ner_train_data.tex
\onecolumn
\small

\begin{longtable}{lllll}
\toprule
\multicolumn{5}{c}{\textbf{NER Task Adapter Training Dataset}} \\\hline\\
                   Language & iso-639 &            Family &                      Genus &         Script \\
\midrule
                 Somali &   som &      Afro-Asiatic &      Lowland East Cushitic &           Latn \\
                 Arabic &   ara &      Afro-Asiatic &                    Semitic &           Arab \\
                Amharic &   amh &      Afro-Asiatic &                    Semitic &           Ethi \\
                 Hebrew &   heb &      Afro-Asiatic &                    Semitic &           Hebr \\
                Maltese &   mlt &      Afro-Asiatic &                    Semitic &           Latn \\
\end{longtable}

\begin{longtable}{lllll}
\toprule
\multicolumn{5}{c}{\textbf{NLI Task Adapter Training Dataset}} \\\hline\\
                   Language & iso-639 &            Family &                      Genus &         Script \\
\midrule
                Arabic &   ara &   Afro-Asiatic &  Semitic &    Arab \\
                Turkish &   tur &         Altaic &   Turkic &    Latn \\
             Vietnamese &   vie & Austro-Asiatic &   Vietic &    Latn \\
                 German &   deu &  Indo-European & Germanic &    Latn \\
                English &   eng &  Indo-European & Germanic &    Latn \\
   Modern Greek (1453-) &   ell &  Indo-European &    Greek &    Grek \\
                   Urdu &   urd &  Indo-European &    Indic &    Arab \\
                  Hindi &   hin &  Indo-European &    Indic &    Deva \\
                 French &   fre &  Indo-European &  Romance &    Latn \\
                Spanish &   spa &  Indo-European &  Romance &    Latn \\
              Bulgarian &   bul &  Indo-European &   Slavic &    Cyrl \\
                Russian &   rus &  Indo-European &   Slavic &    Cyrl \\
Swahili (macrolanguage) &   swa &    Niger-Congo &    Bantu &    Latn \\
                Chinese &   zho &   Sino-Tibetan &  Chinese & Chinese \\
                   Thai &   tha &      Tai-Kadai &  Kam-Tai &    Thai \\

\bottomrule
\caption{Task Adapter training dataset details (taken from XNLI~\cite{conneau-etal-2018-xnli}) for Natural Language Inference. }
    \label{tab:nli_dataset}
\end{longtable}

\begin{longtable}{lllll}
\toprule
\multicolumn{5}{c}{\textbf{Extractive Question Answering Task Adapter Training Dataset}} \\\hline\\
                   Language & iso-639 &            Family &                      Genus &         Script \\
\midrule
Arabic &   ara &  Afro-Asiatic &   Semitic &   Arab \\
             Indonesian &   idn &   Astronesian &     Malay &   Latn \\
                 Telugu &   tel &     Dravidian & Dravidian &   Telu \\
                English &   eng & Indo-European &  Germanic &   Latn \\
                Bengali &   ben & Indo-European &     Indic &   Beng \\
                Russian &   rus & Indo-European &    Slavic &   Cyrl \\
                 Korean &   kor &        Korean &    Korean &   Kore \\
Swahili (macrolanguage) &   swa &   Niger-Congo &     Bantu &   Latn \\
                Finnish &   fin &        Uralic &    Finnic &   Latn \\

\bottomrule
\caption{Task Adapter training dataset details (taken from TyDiQA~\cite{clark-etal-2020-tydi}) for Extractive Question Answering. }
    \label{tab:qa_dataset}
\end{longtable}

\twocolumn
\normalsize

%% file: evald.tex
\onecolumn
\small

\begin{longtable}{llllllll}
\toprule
\multicolumn{8}{c}{\textbf{Evaluation Languages}} \\\hline\\
{} &                          Language &    iso-639 &   NER  &    UDP and POS  & XNLI & ANLI & TyDiQA\\
\midrule
1   &                   Achinese &  ace &   &  \tikzxmark&\tikzxmark&\tikzxmark&\tikzxmark \\
2   &                  Afrikaans &  afr &   &  &\tikzxmark&\tikzxmark&\tikzxmark \\
3   &       Assyrian Neo-Aramaic &  aii &  \tikzxmark &  &\tikzxmark&\tikzxmark&\tikzxmark \\
4   &     South Levantine Arabic &  ajp &  \tikzxmark &  &\tikzxmark&\tikzxmark&\tikzxmark \\
5   &                   Akkadian &  akk &  \tikzxmark &  &\tikzxmark&\tikzxmark&\tikzxmark \\
\end{longtable}   
\twocolumn
\normalsize

%% file: transfer_graphs.tex
\begin{figure*}[t]
    \centering
    \begin{tabular}{c}
gub \\
\includegraphics[width=.8\textwidth]{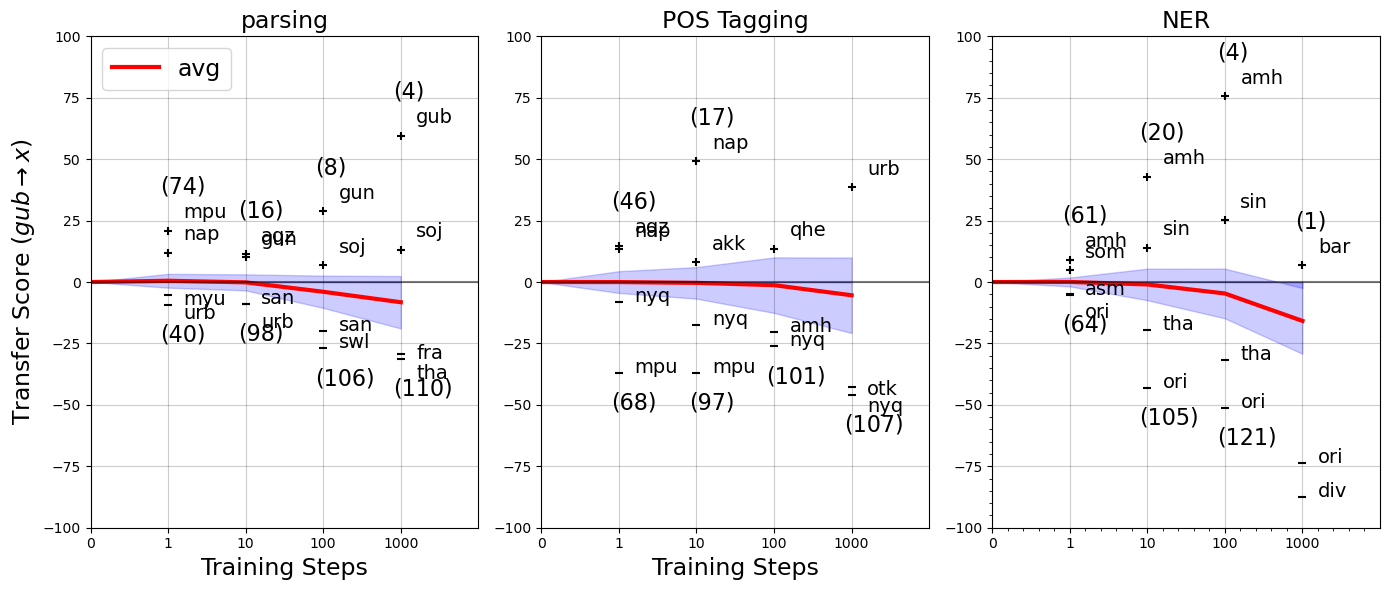} \\
est \\
\includegraphics[width=.8\textwidth]{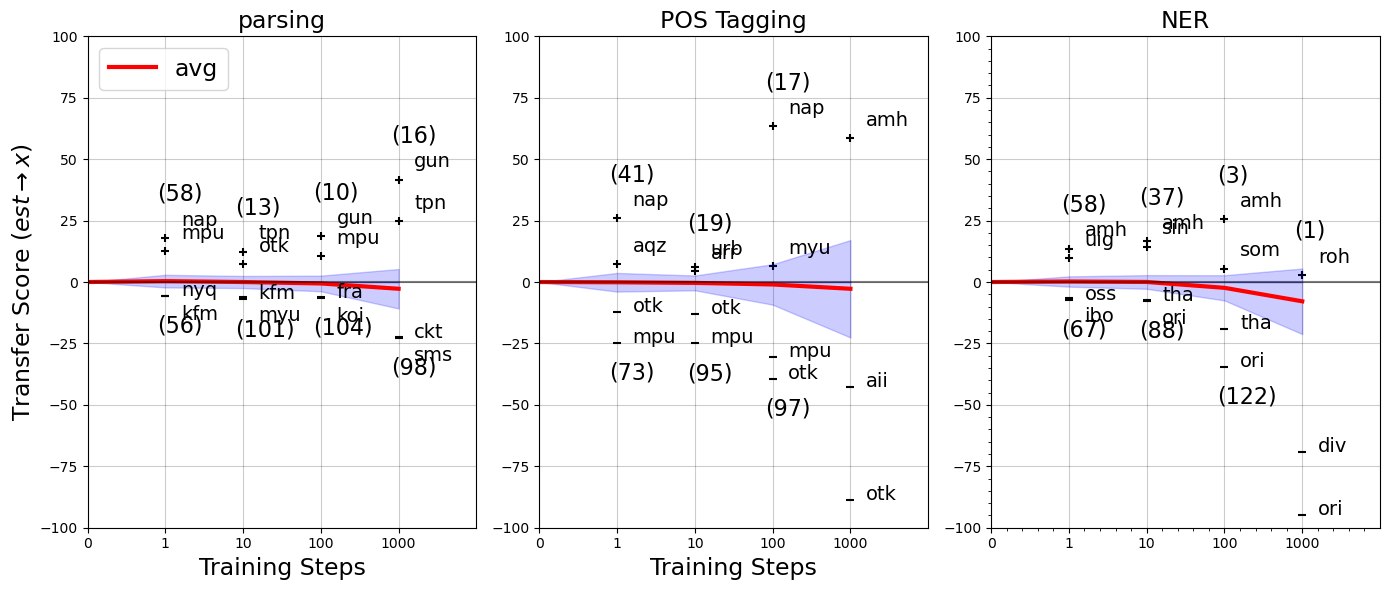} \\
bre \\
\includegraphics[width=.8\textwidth]{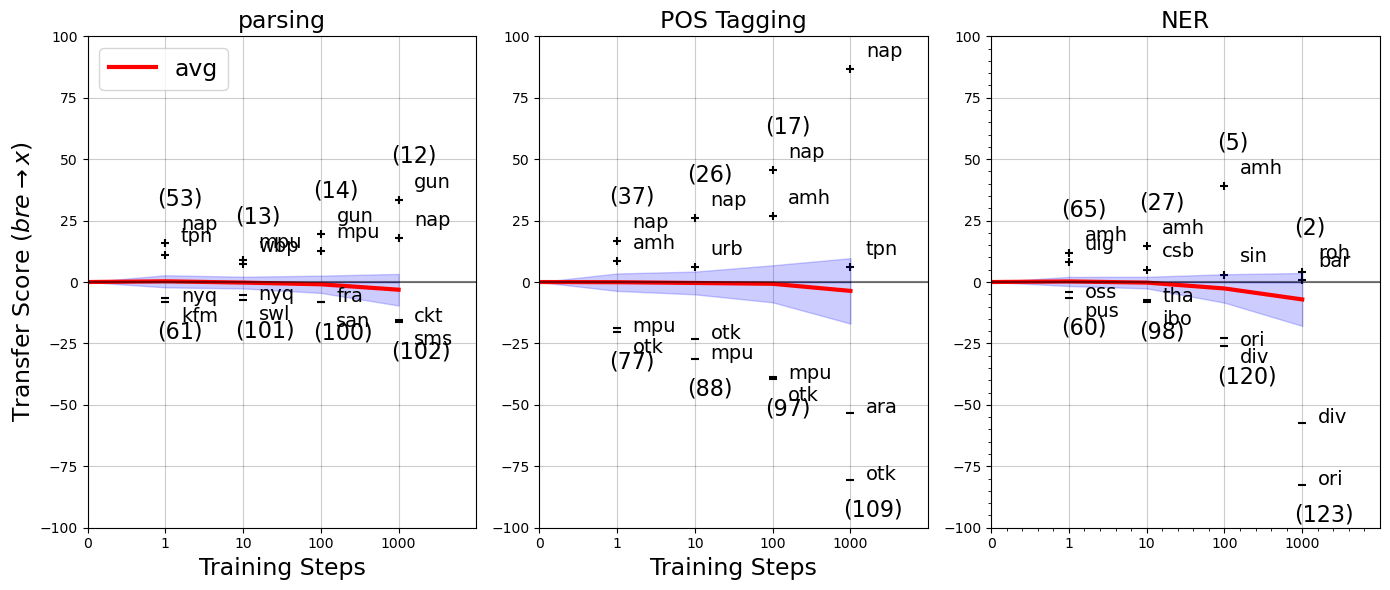} \\
eng \\
\includegraphics[width=.8\textwidth]{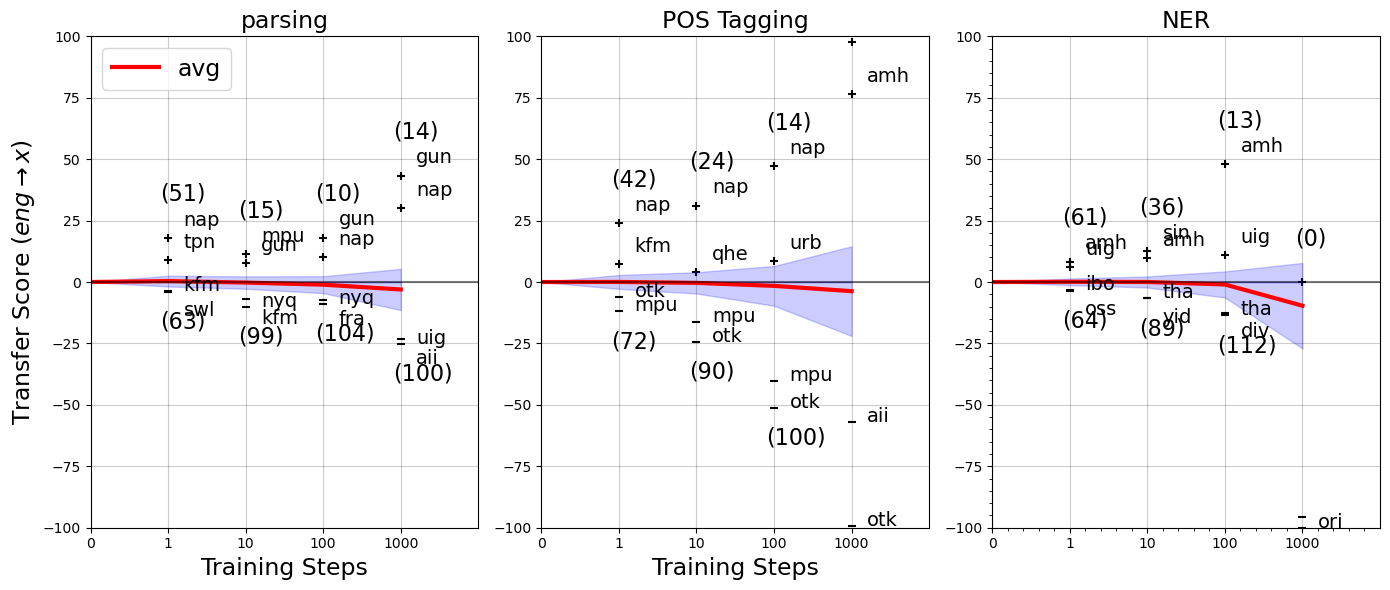} \\

    \end{tabular}
    \caption{Aggregated Transfer Progression through training steps}
    \label{fig:first}
\end{figure*}

\begin{figure*}[t]
    \centering
    \begin{tabular}{c}
ben \\
\includegraphics[width=.8\textwidth]{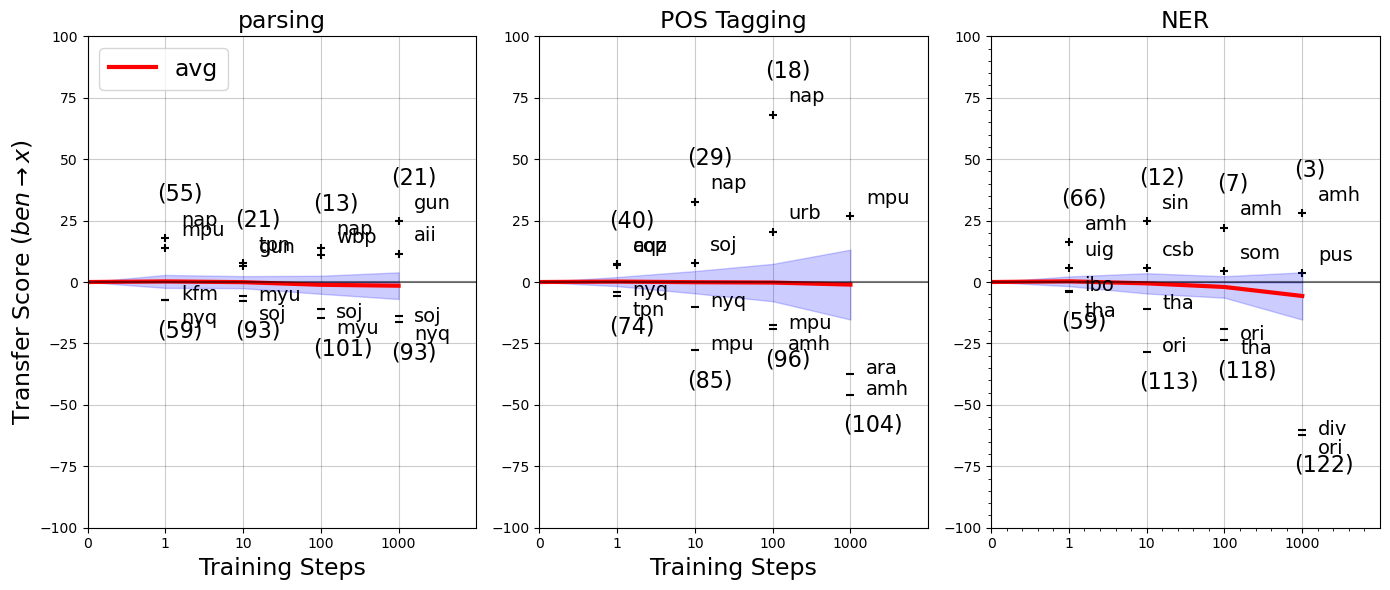} \\
kmr \\
\includegraphics[width=.8\textwidth]{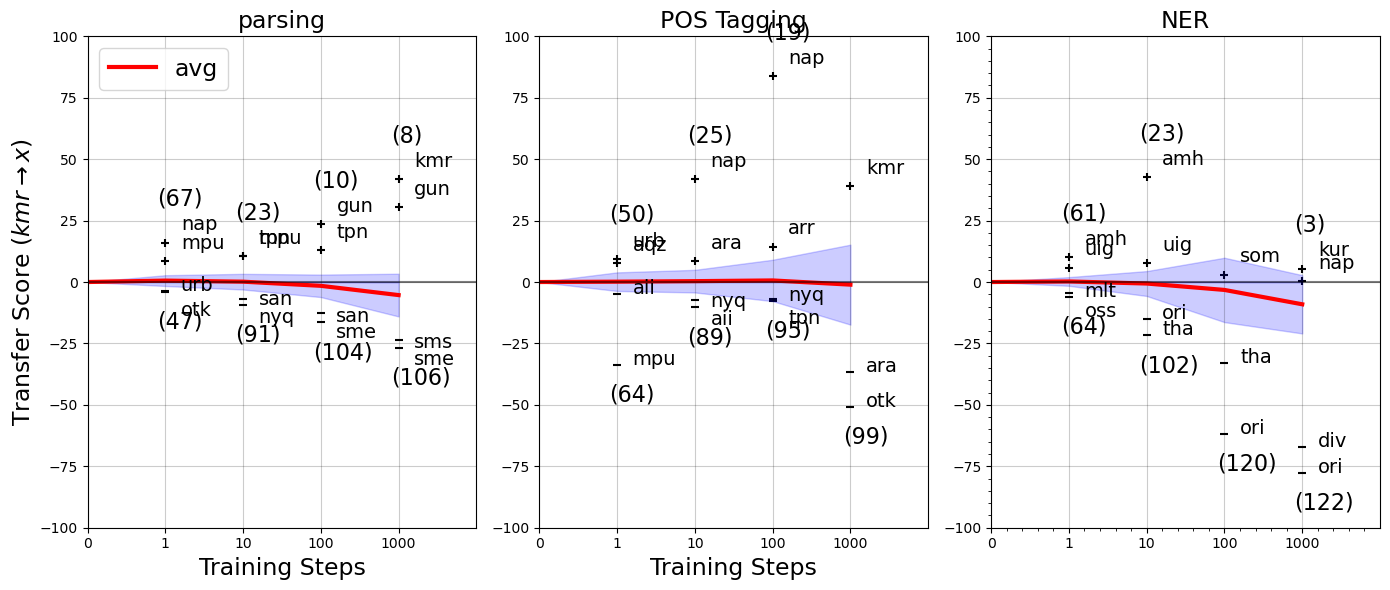} \\
spa \\
\includegraphics[width=.8\textwidth]{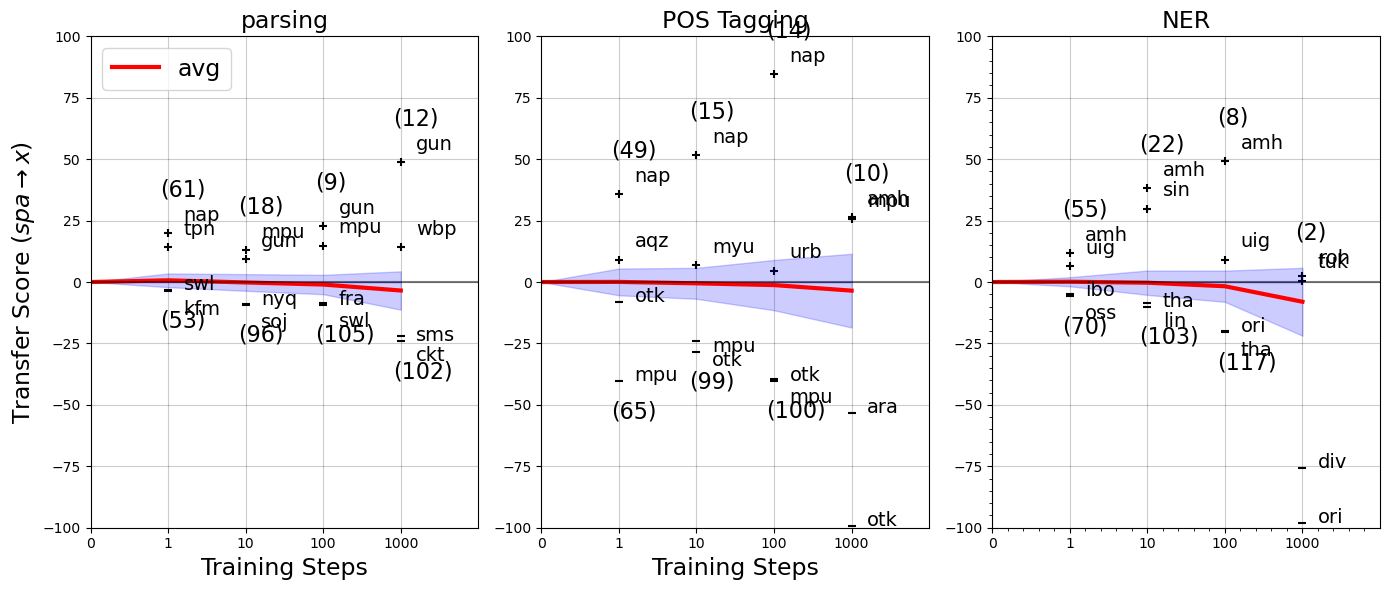} \\
bul \\
\includegraphics[width=.8\textwidth]{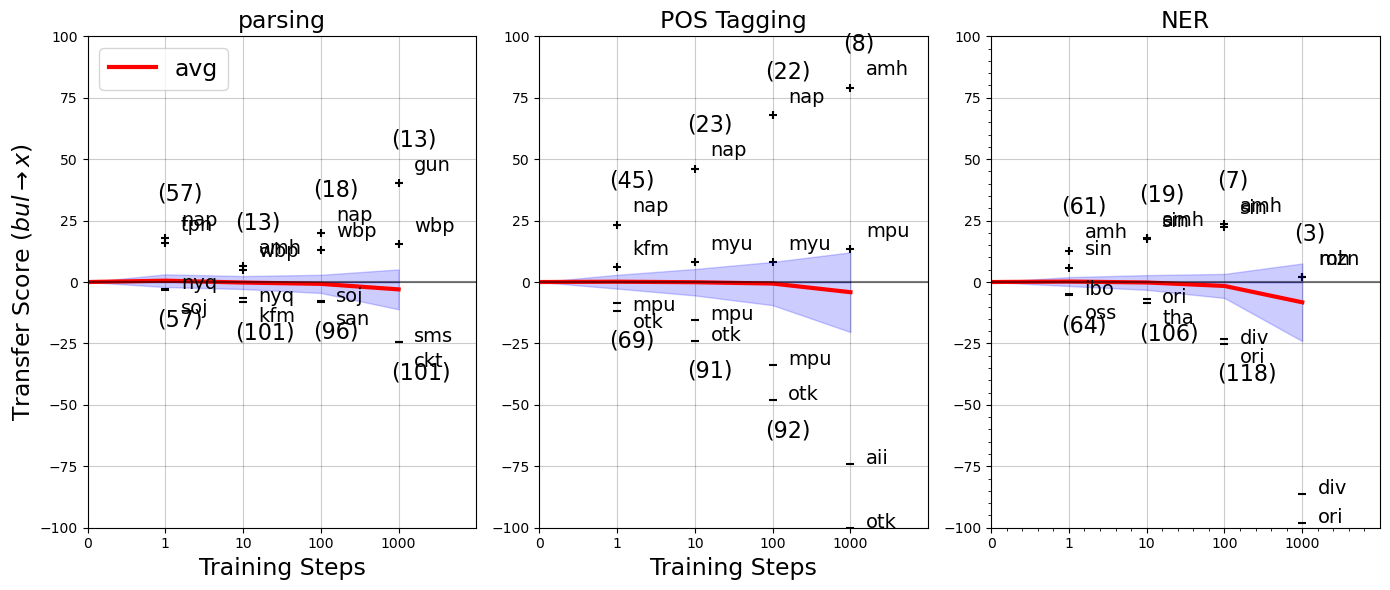} \\

    \end{tabular}
    \caption{Aggregated Transfer Progression through training steps}
    \label{fig:plot_gen}
\end{figure*}

\begin{figure*}[t]
    \centering
    \begin{tabular}{c}
pms \\
\includegraphics[width=.8\textwidth]{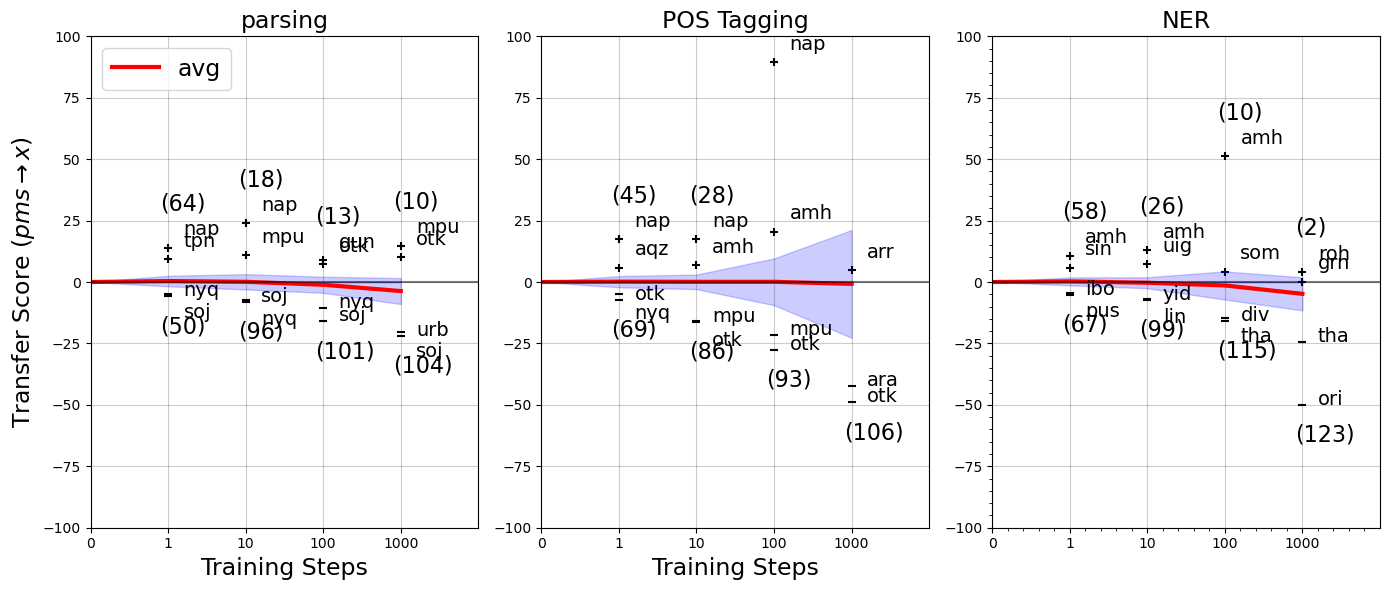} \\
gle \\
\includegraphics[width=.8\textwidth]{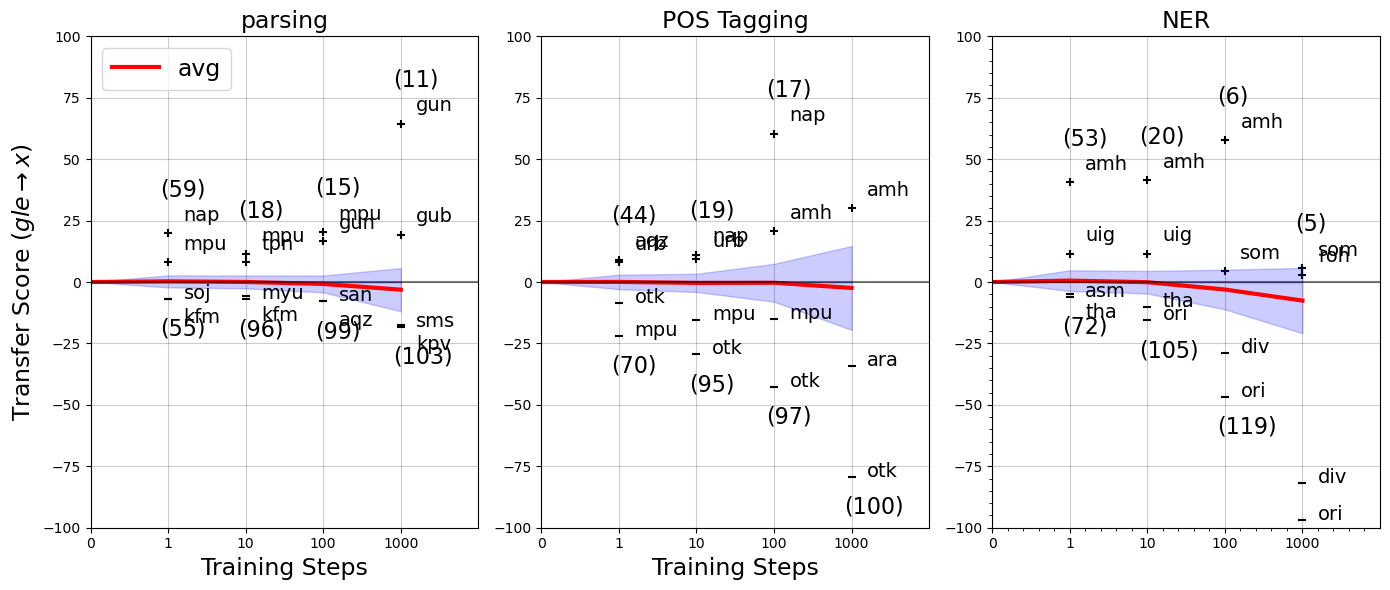} \\
nep \\
\includegraphics[width=.8\textwidth]{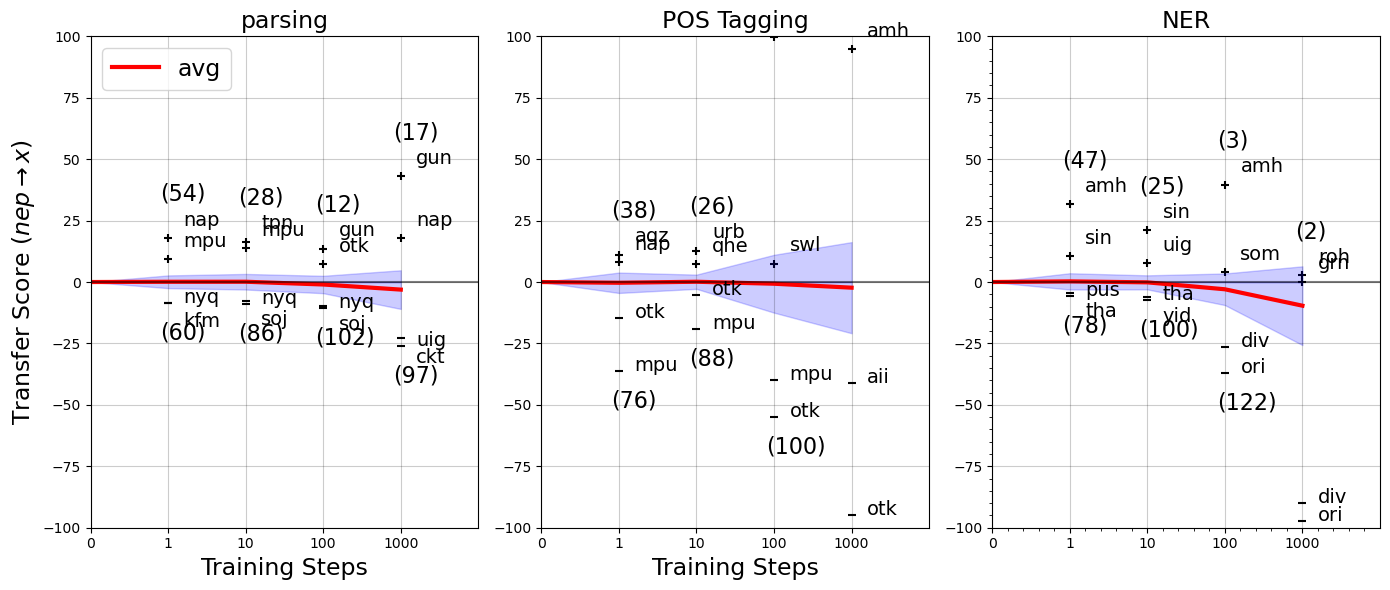} \\
cym \\
\includegraphics[width=.8\textwidth]{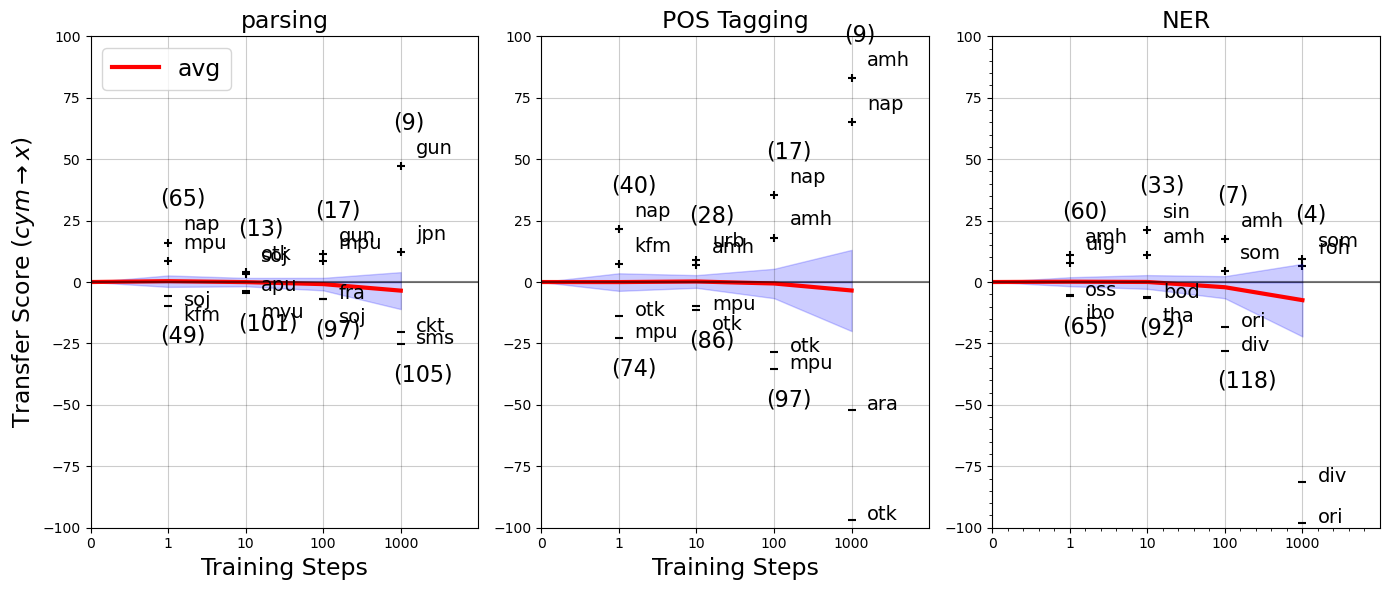} \\

    \end{tabular}
    \caption{Aggregated Transfer Progression through training steps}
    \label{fig:plot_gen}
\end{figure*}

\begin{figure*}[t]
    \centering
    \begin{tabular}{c}
fin \\
\includegraphics[width=.8\textwidth]{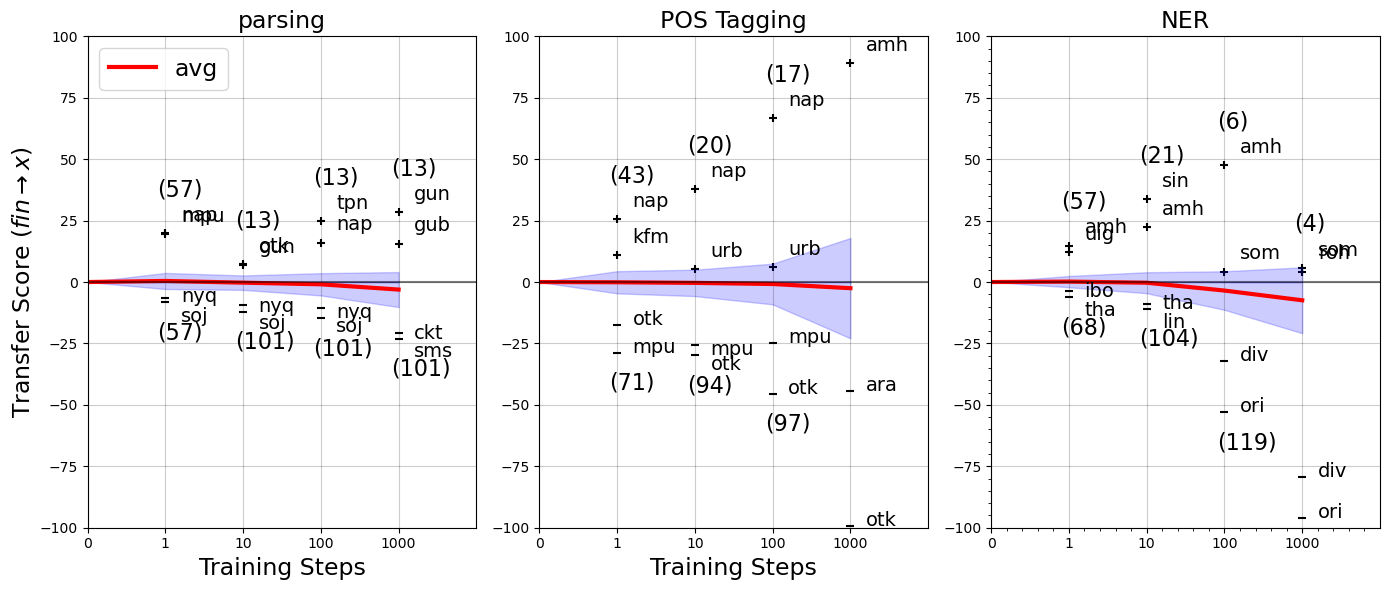} \\
hye \\
\includegraphics[width=.8\textwidth]{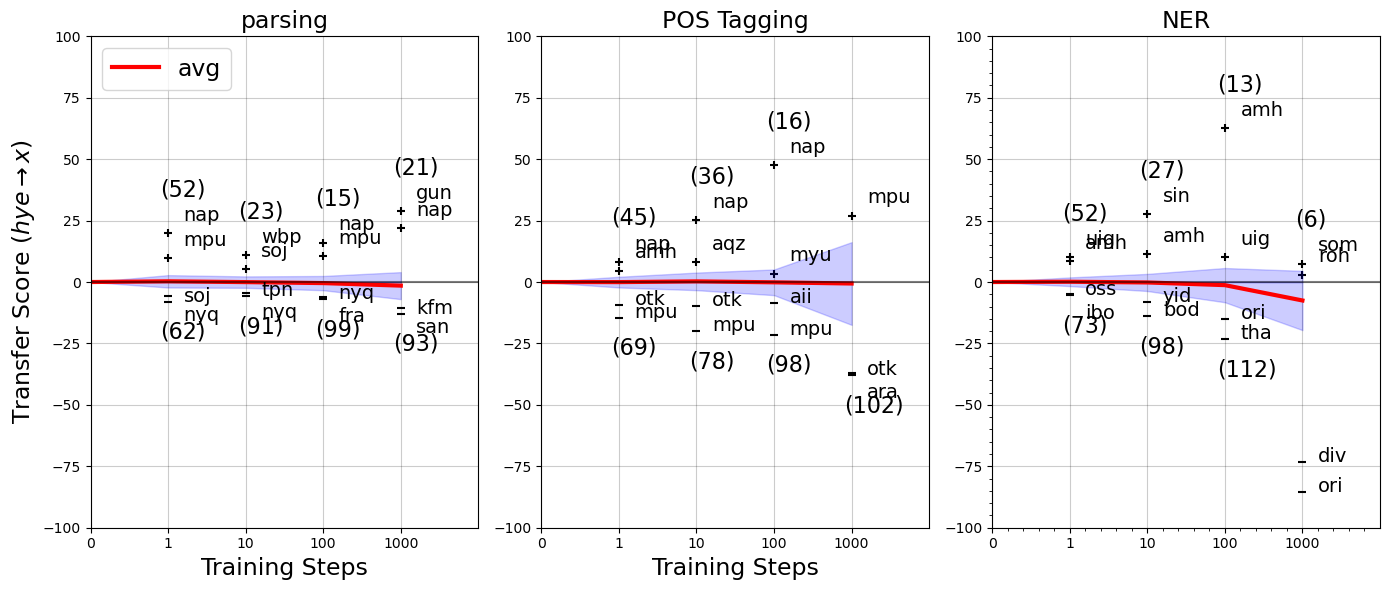} \\
mya \\
\includegraphics[width=.8\textwidth]{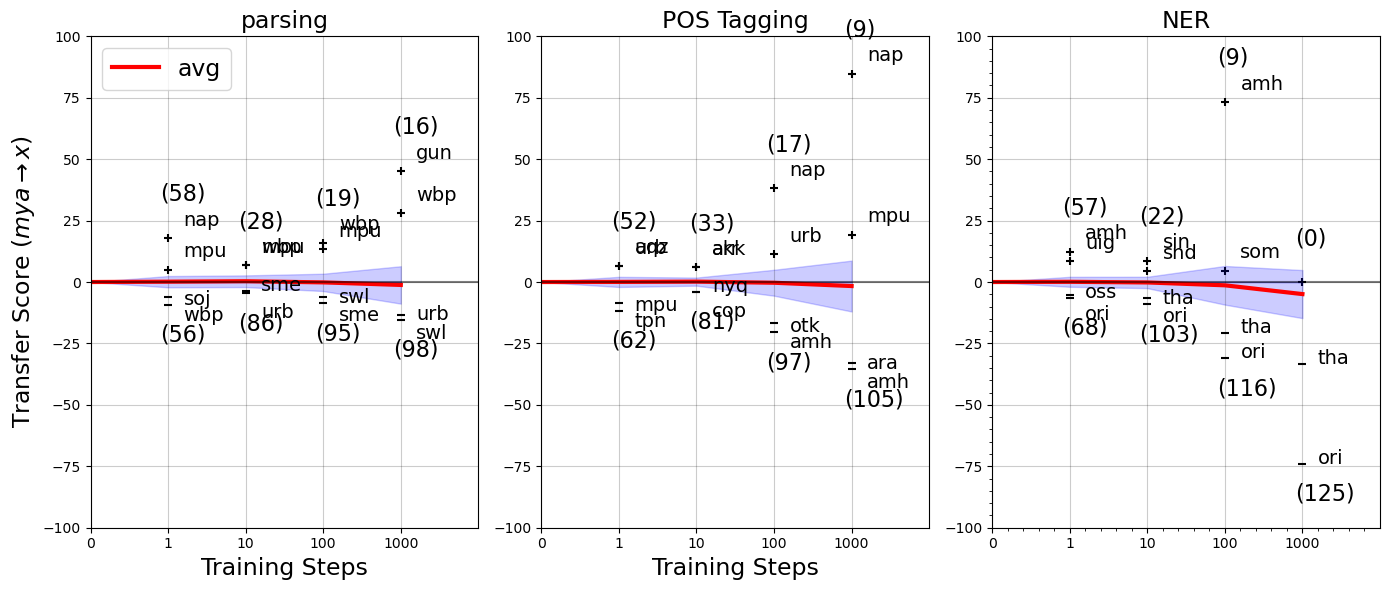} \\
hin \\
\includegraphics[width=.8\textwidth]{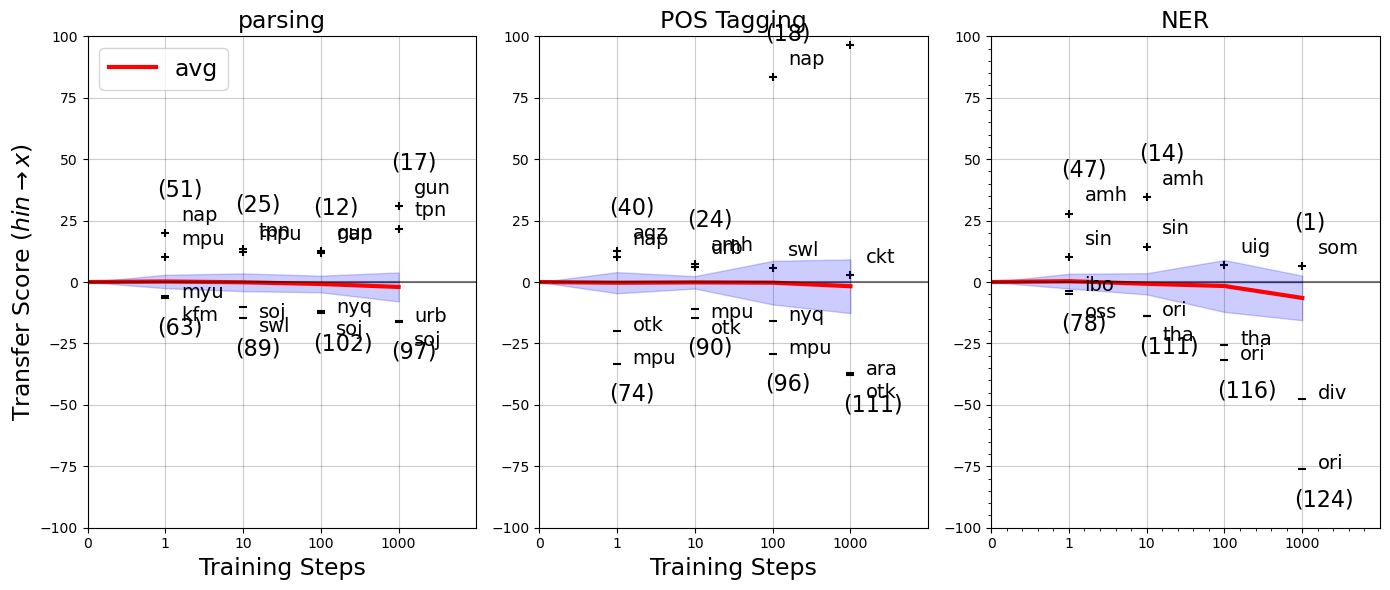} \\

    \end{tabular}
    \caption{Aggregated Transfer Progression through training steps}
    \label{fig:plot_gen}
\end{figure*}

\begin{figure*}[t]
    \centering
    \begin{tabular}{c}
tel \\
\includegraphics[width=.8\textwidth]{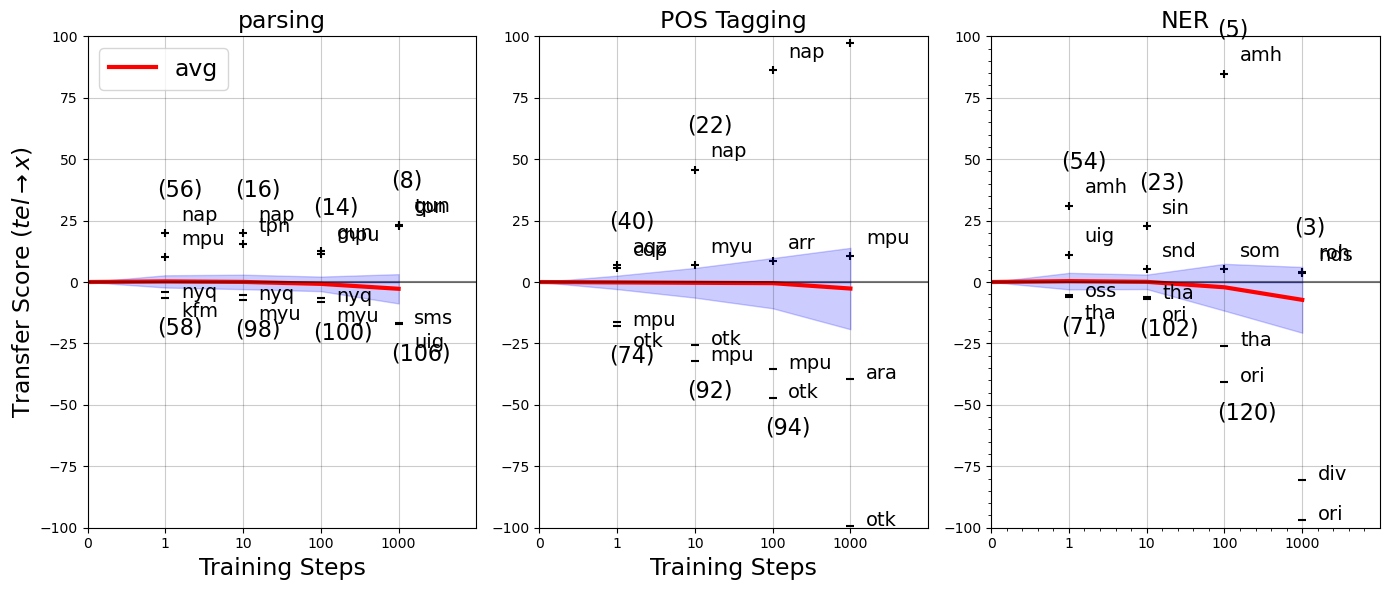} \\
tam \\
\includegraphics[width=.8\textwidth]{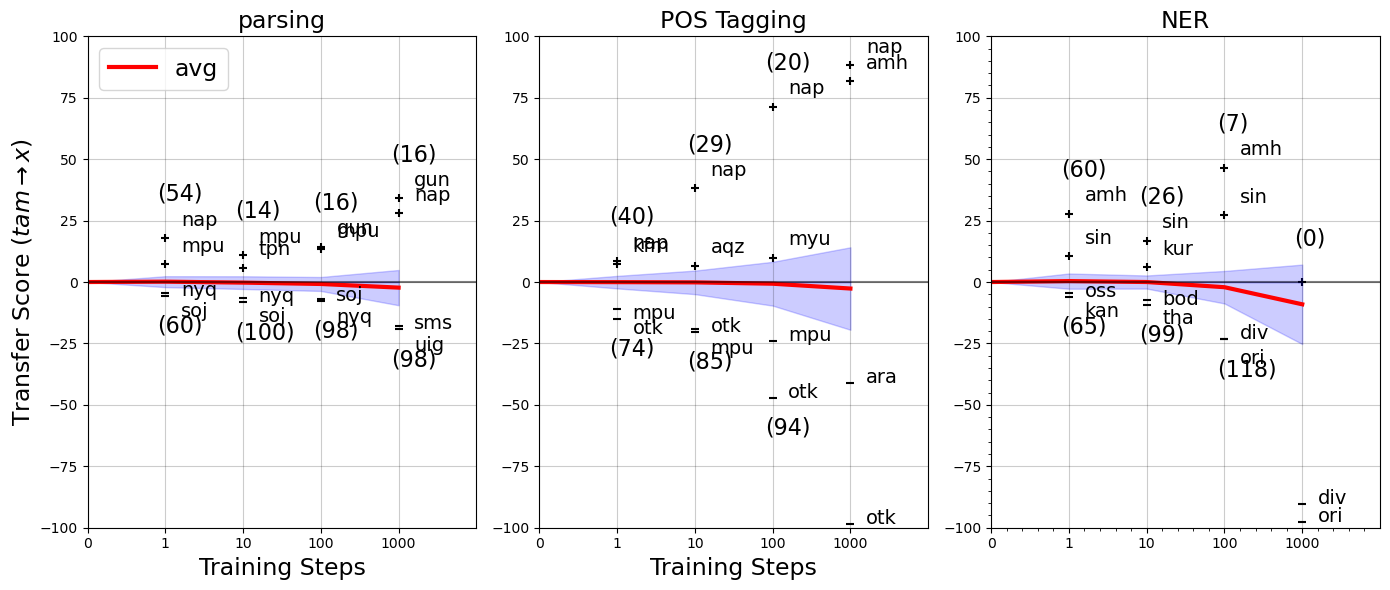} \\
kor \\
\includegraphics[width=.8\textwidth]{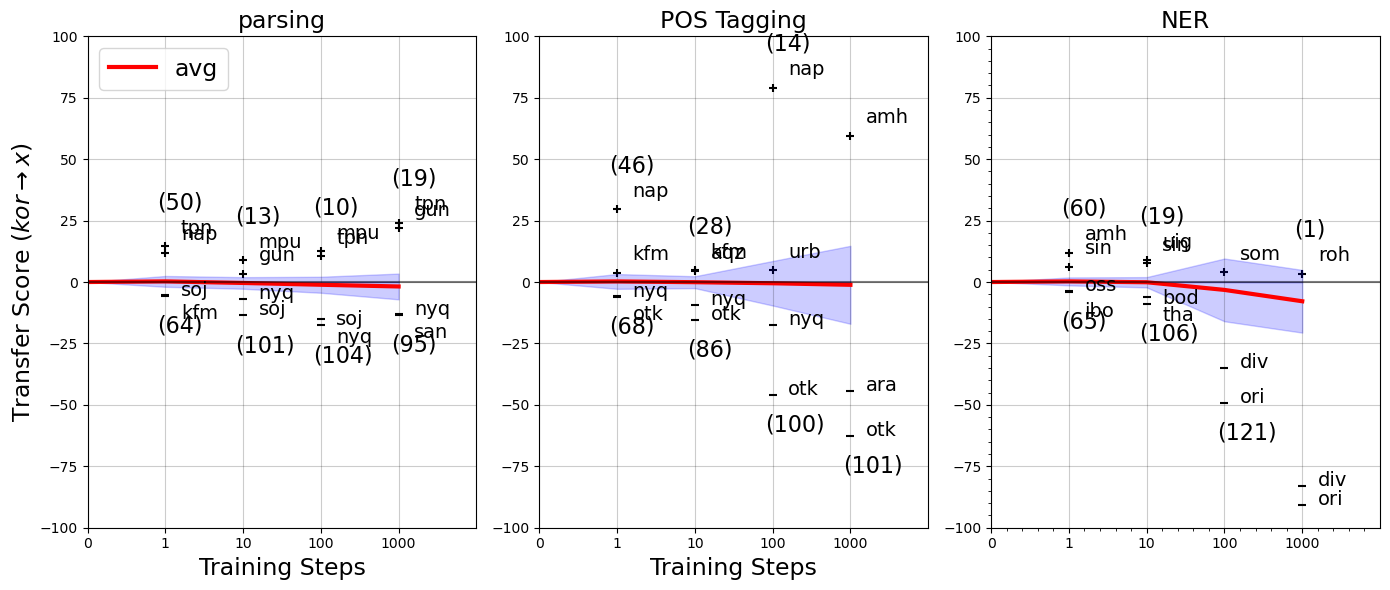} \\
ell \\
\includegraphics[width=.8\textwidth]{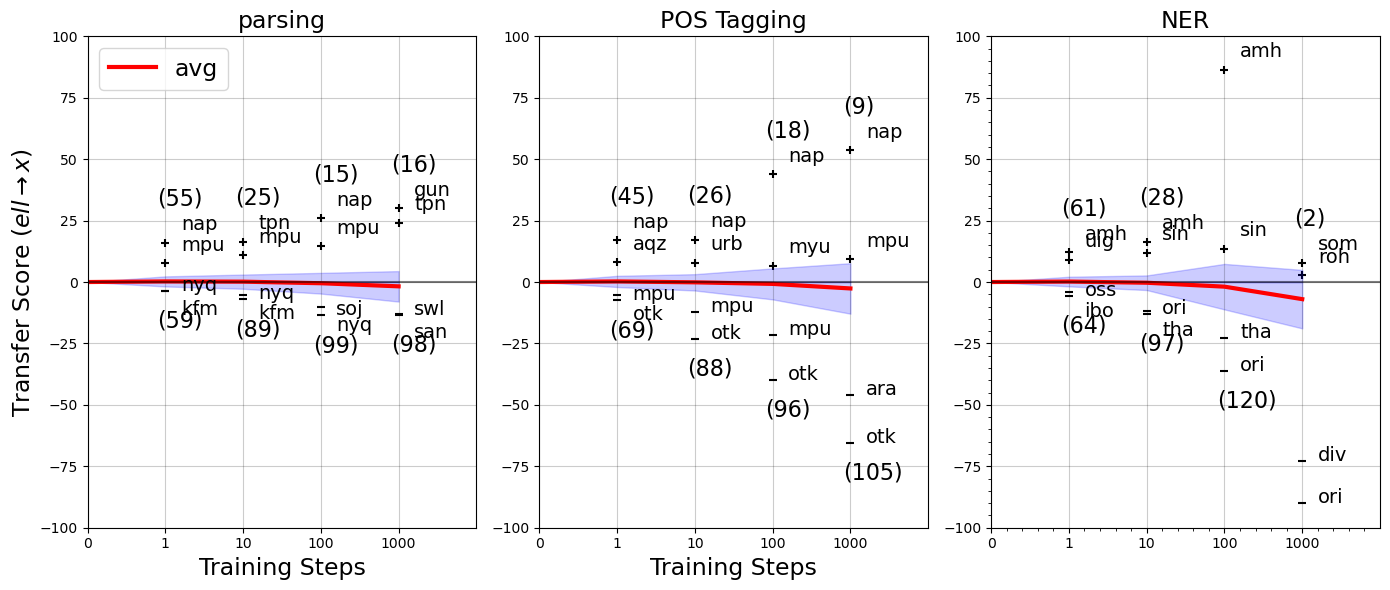} \\

    \end{tabular}
    \caption{Aggregated Transfer Progression through training steps}
    \label{fig:plot_gen}
\end{figure*}

\begin{figure*}[t]
    \centering
    \begin{tabular}{c}
hun \\
\includegraphics[width=.8\textwidth]{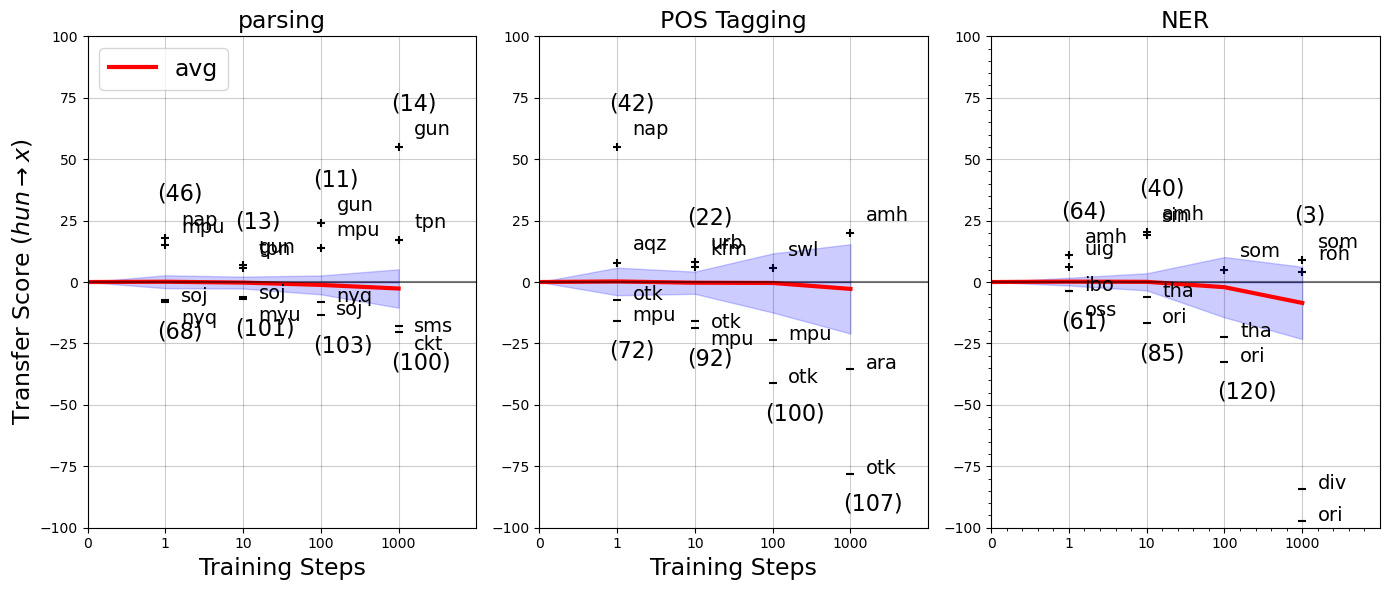} \\
heb \\
\includegraphics[width=.8\textwidth]{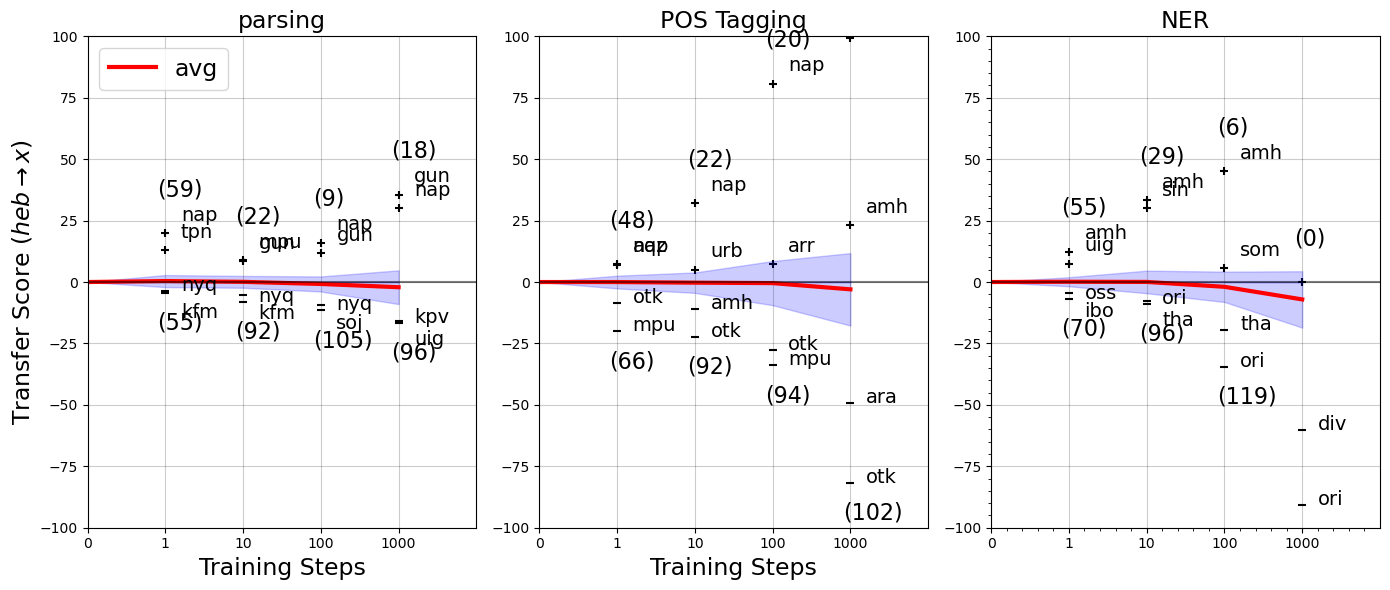} \\
zho \\
\includegraphics[width=.8\textwidth]{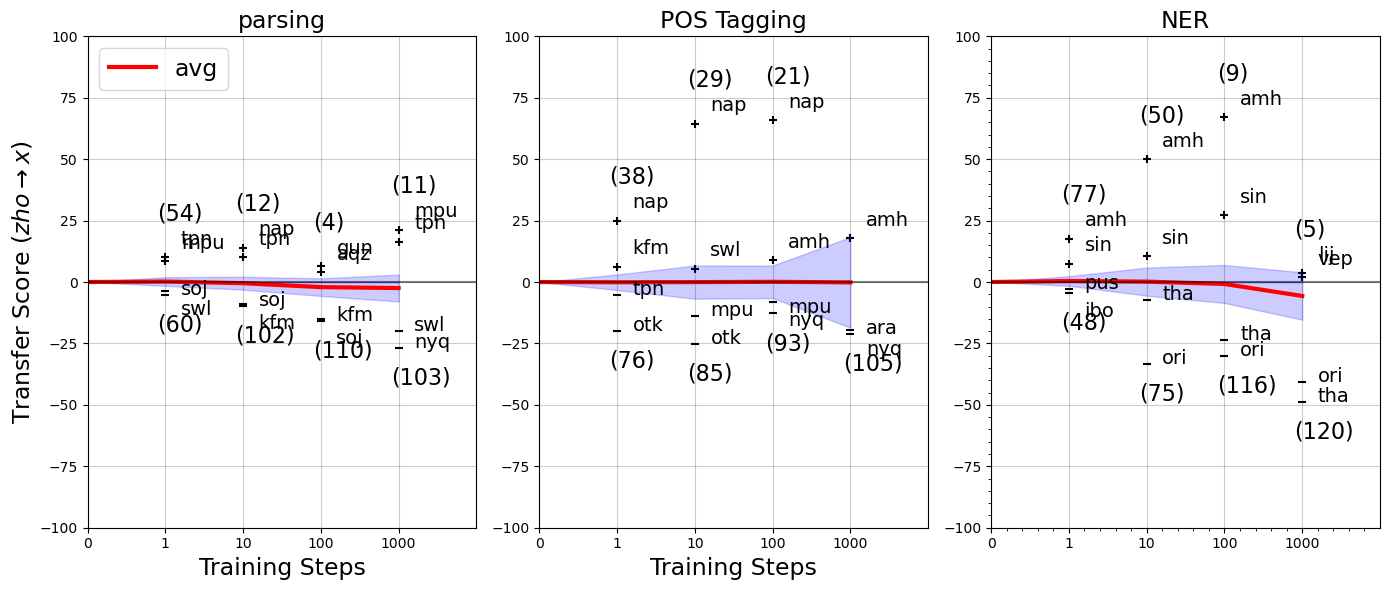} \\
ara \\
\includegraphics[width=.8\textwidth]{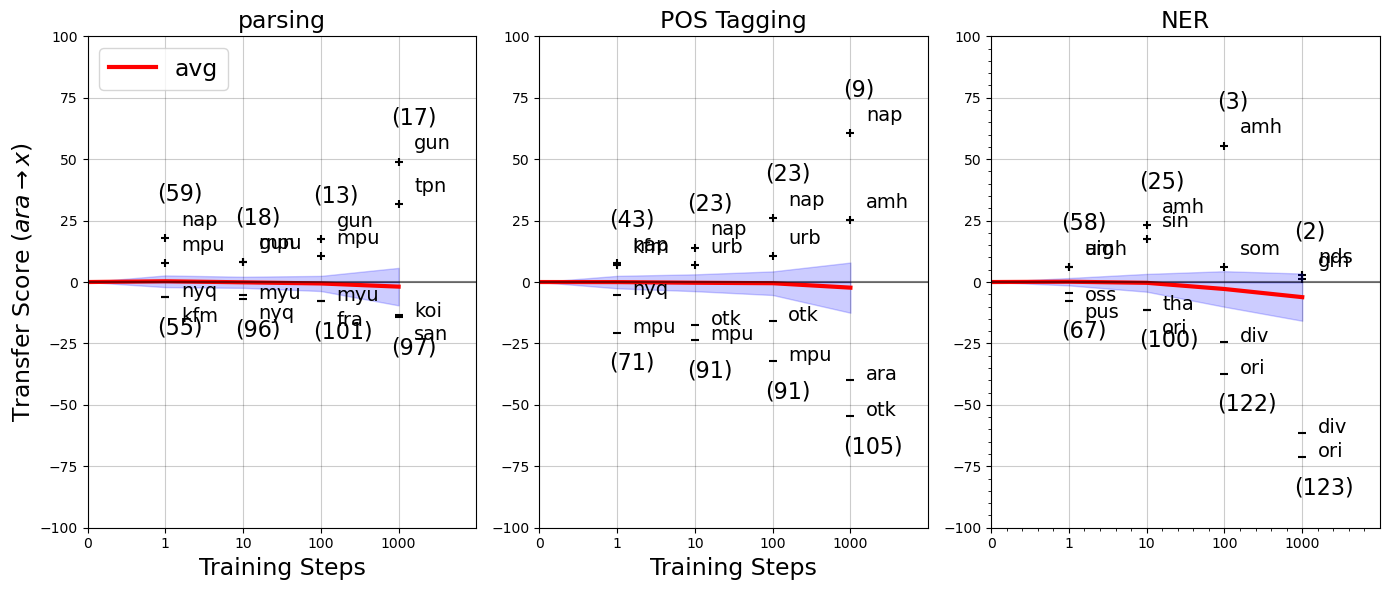} \\

    \end{tabular}
    \caption{Aggregated Transfer Progression through training steps}
    \label{fig:plot_gen}
\end{figure*}

\begin{figure*}[t]
    \centering
    \begin{tabular}{c}
swe \\
\includegraphics[width=.8\textwidth]{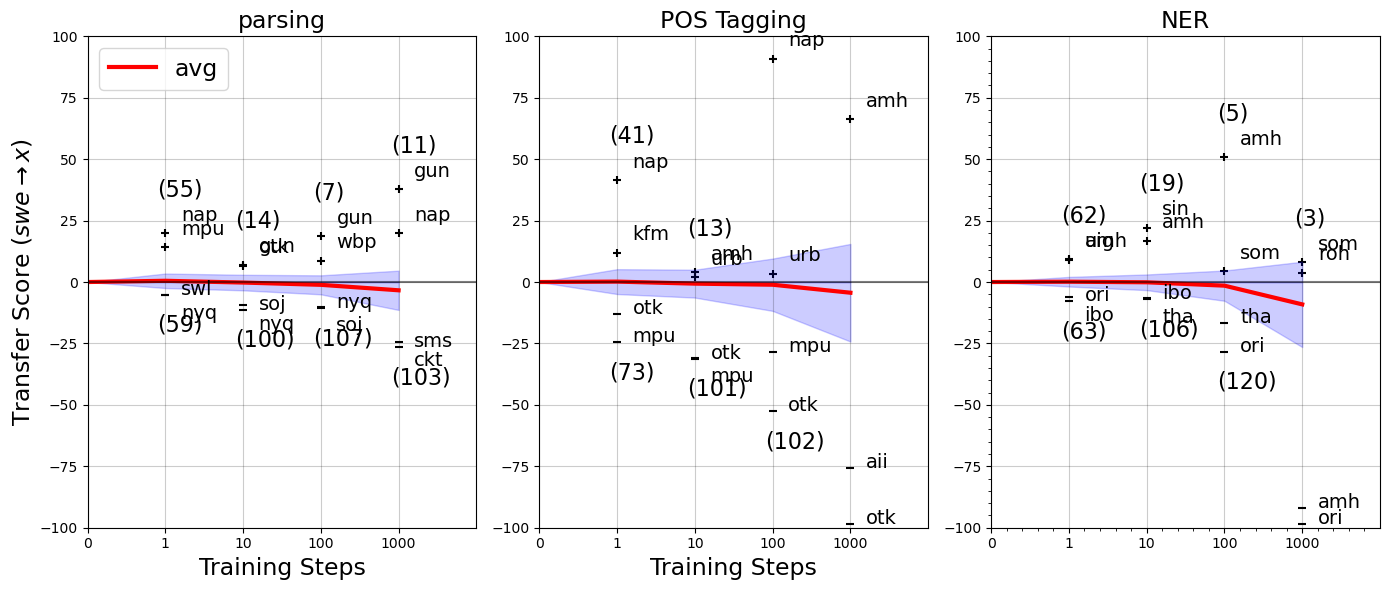} \\
jpn \\
\includegraphics[width=.8\textwidth]{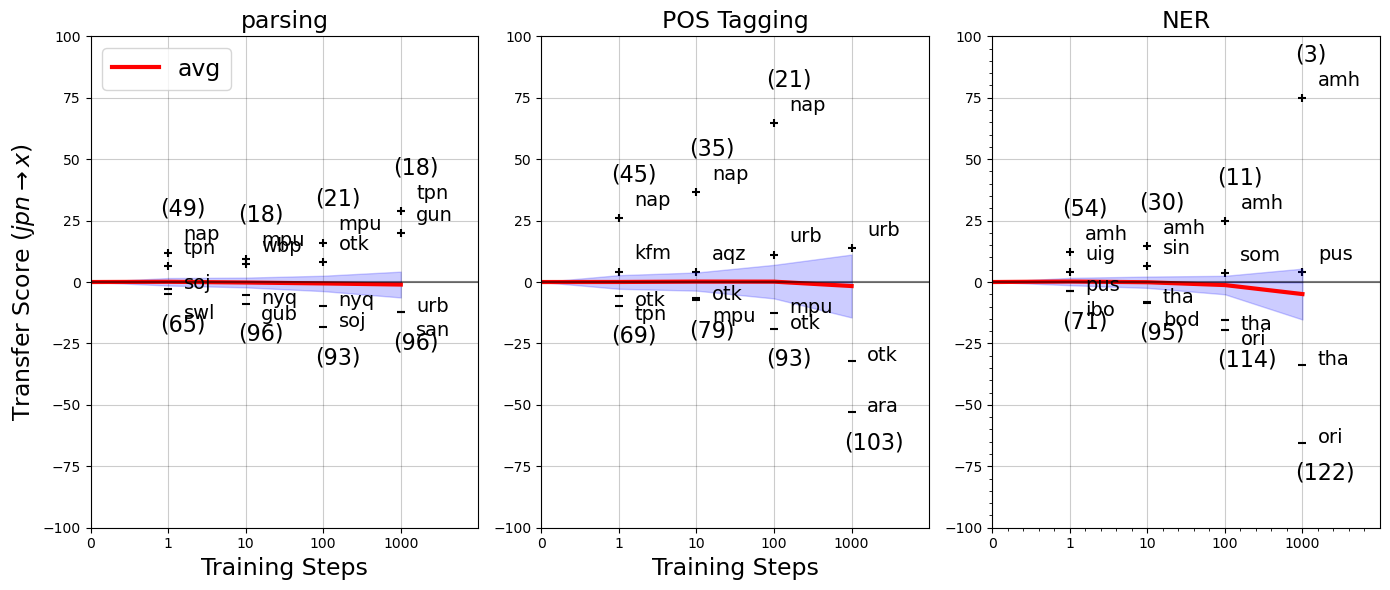} \\
fra \\
\includegraphics[width=.8\textwidth]{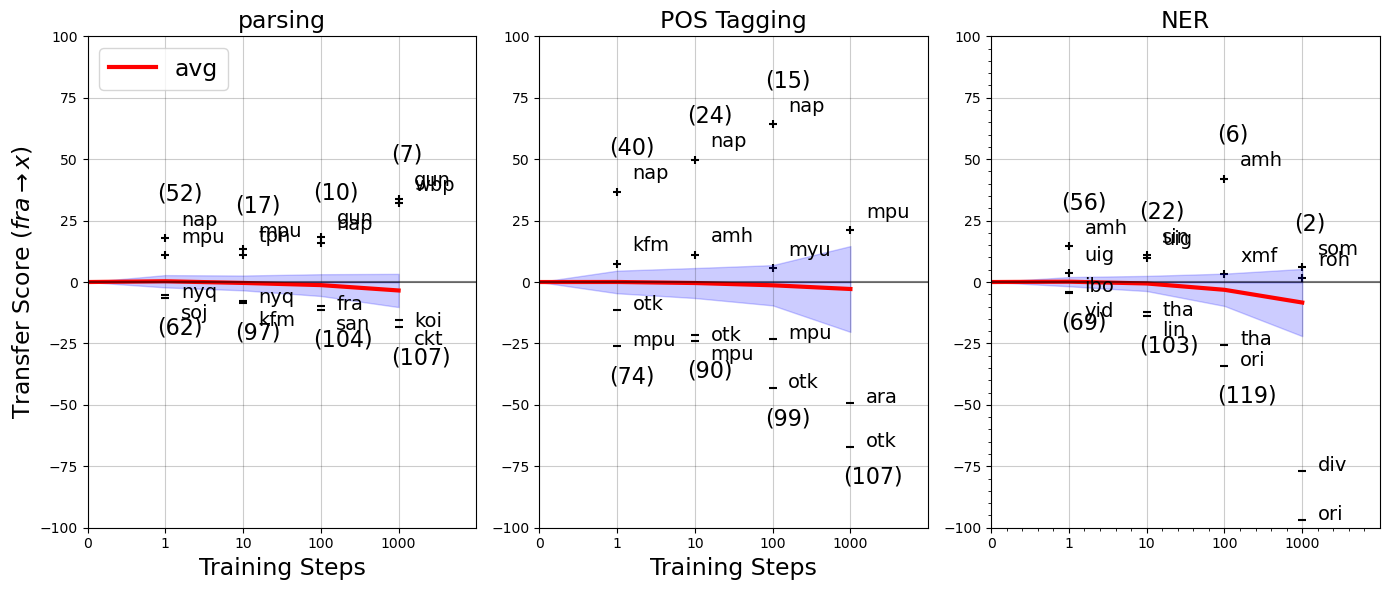} \\
deu \\
\includegraphics[width=.8\textwidth]{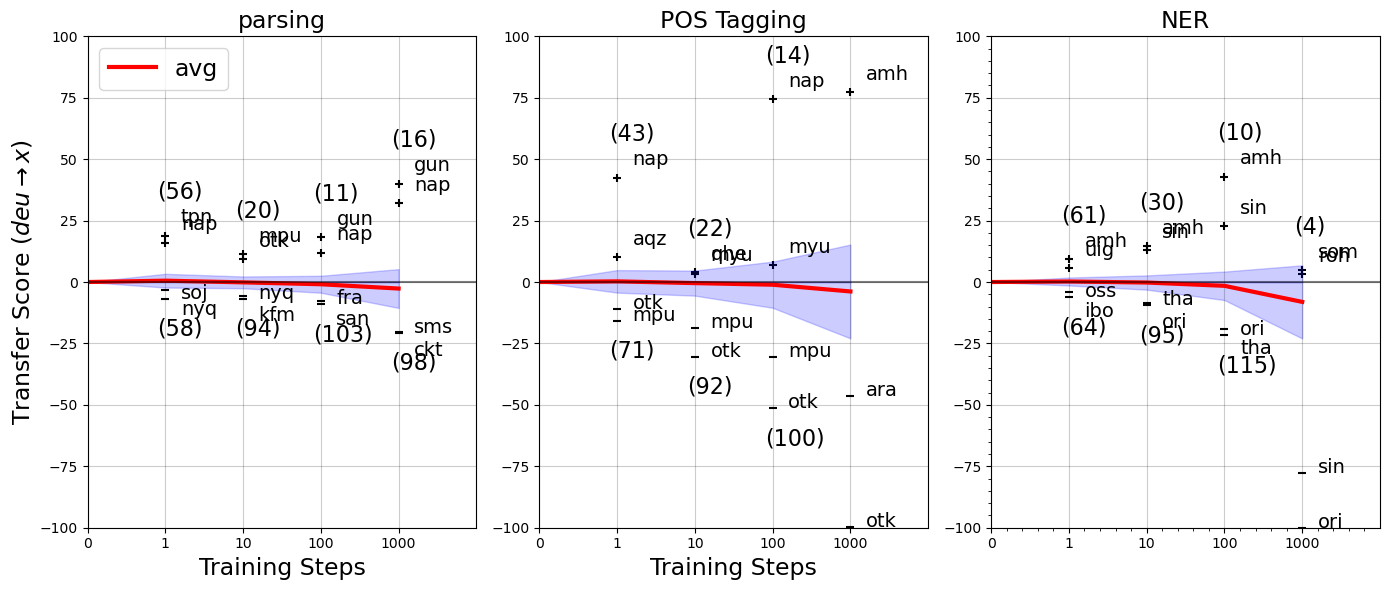} \\

    \end{tabular}
    \caption{Aggregated Transfer Progression through training steps}
    \label{fig:plot_gen}
\end{figure*}

\begin{figure*}[t]
    \centering
    \begin{tabular}{c}
rus \\
\includegraphics[width=.8\textwidth]{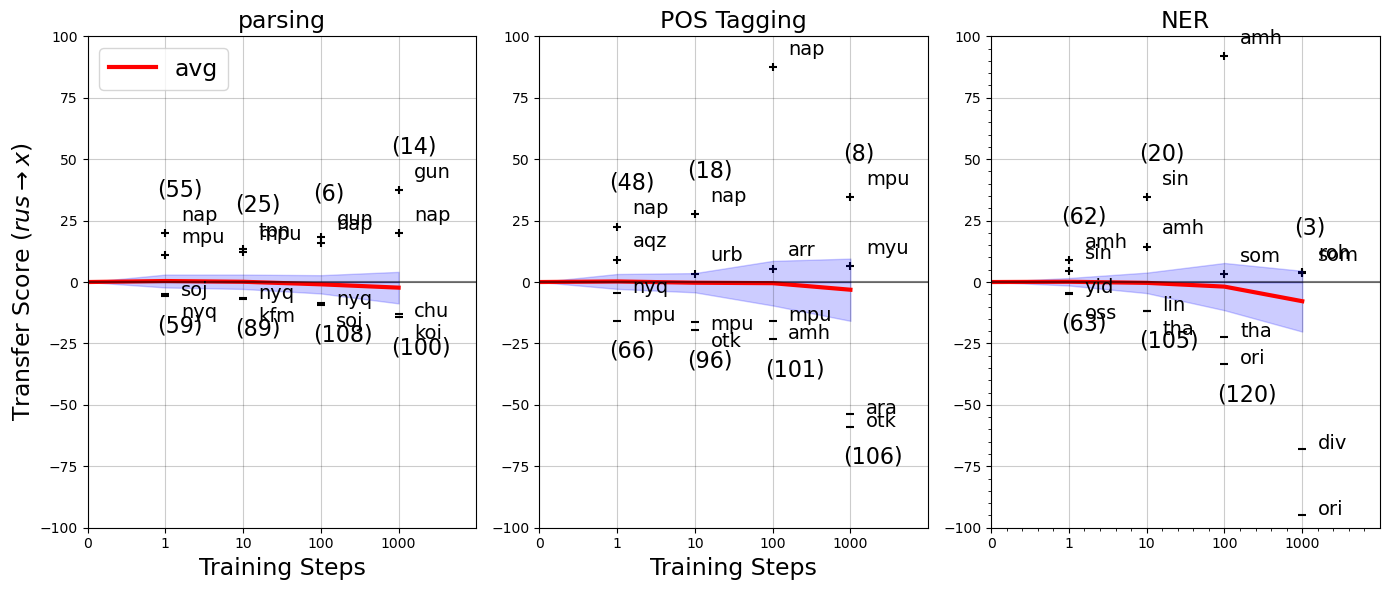} \\
bam \\
\includegraphics[width=.8\textwidth]{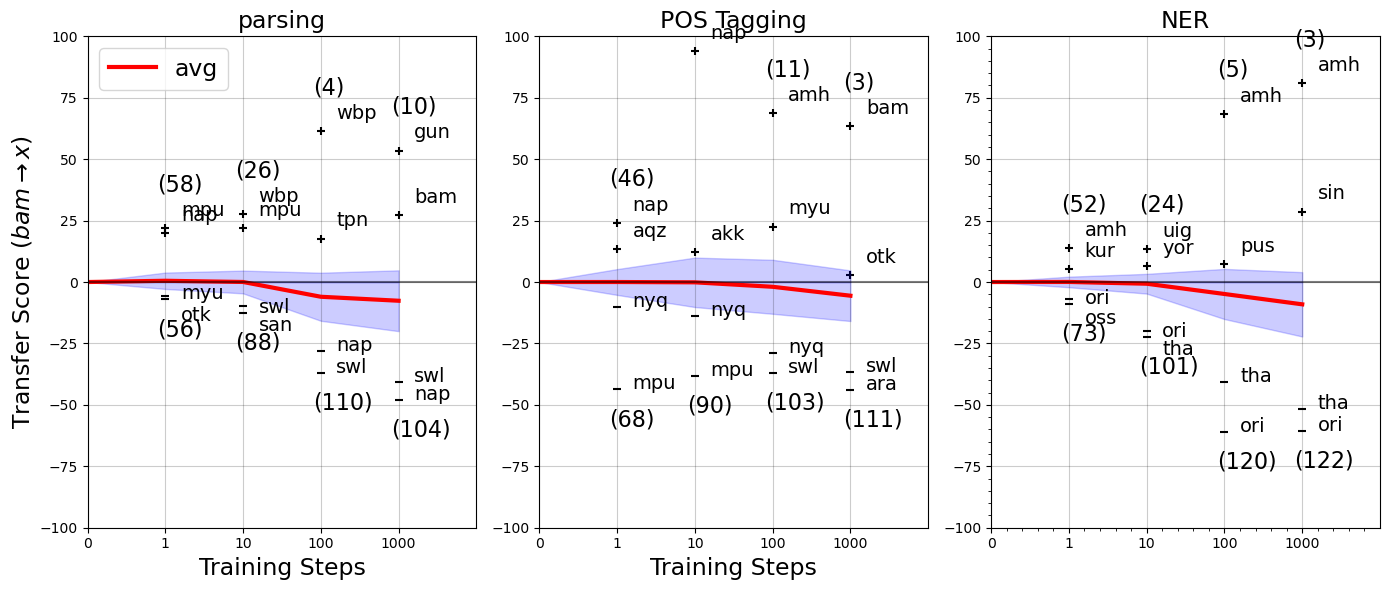} \\
ewe \\
\includegraphics[width=.8\textwidth]{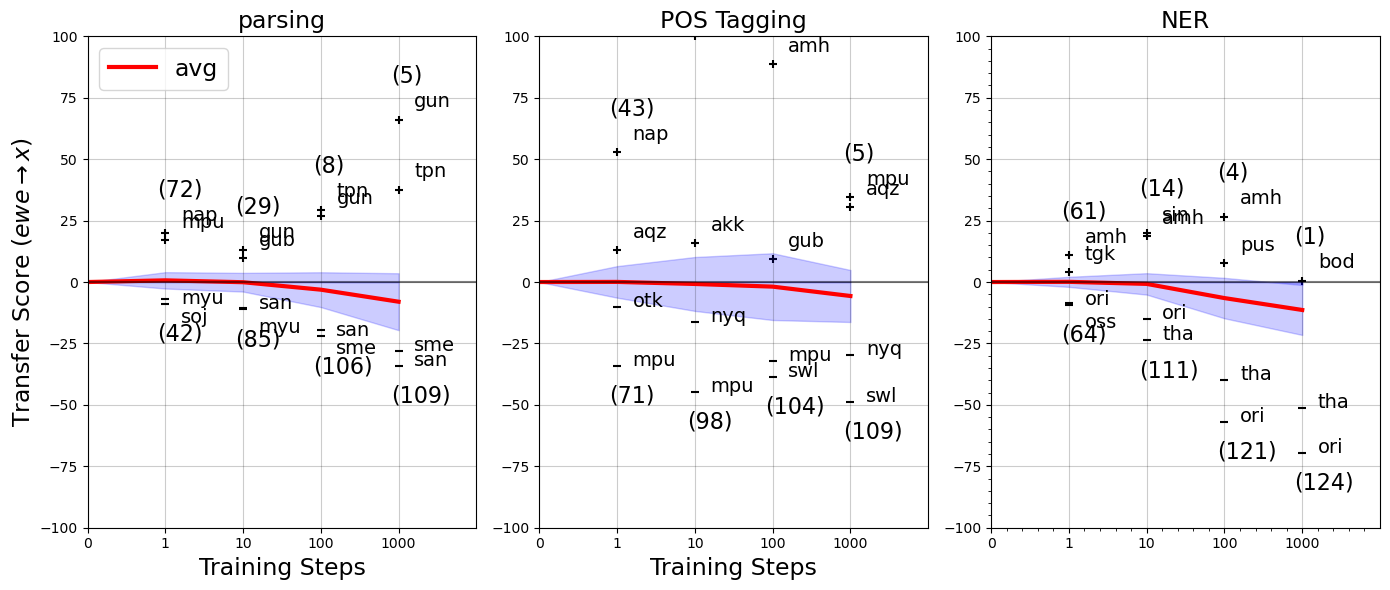} \\
hau \\
\includegraphics[width=.8\textwidth]{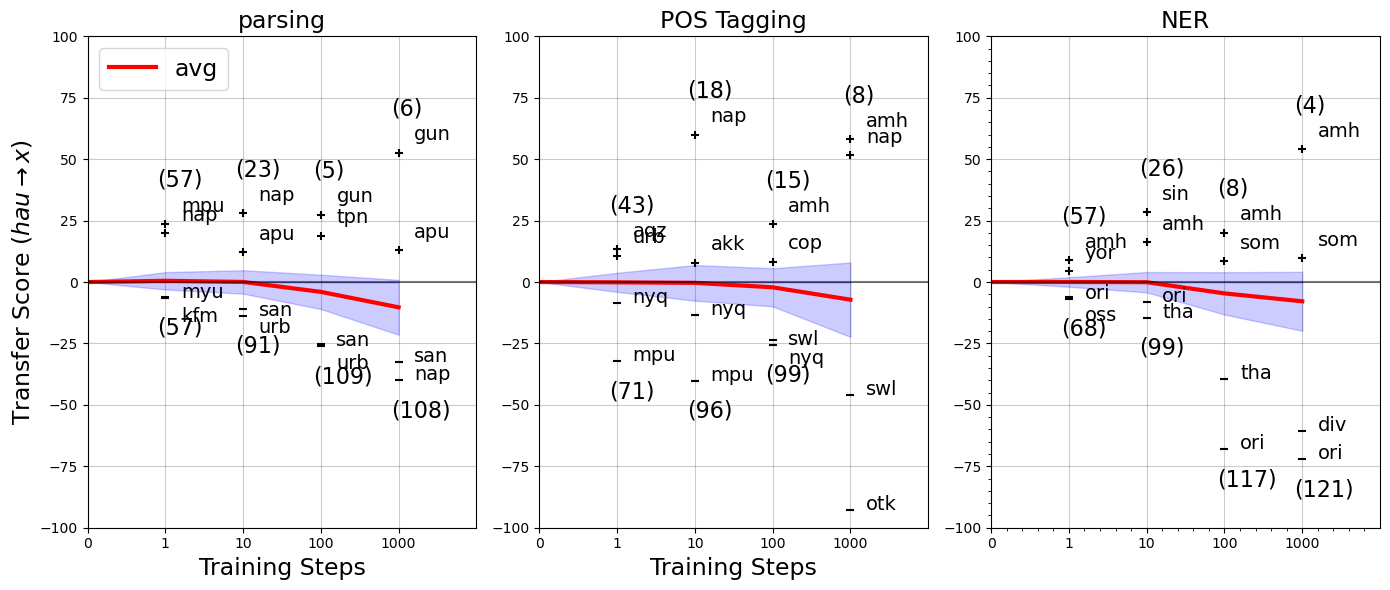} \\

    \end{tabular}
    \caption{Aggregated Transfer Progression through training steps}
    \label{fig:plot_gen}
\end{figure*}

\begin{figure*}[t]
    \centering
    \begin{tabular}{c}
ibo \\
\includegraphics[width=.8\textwidth]{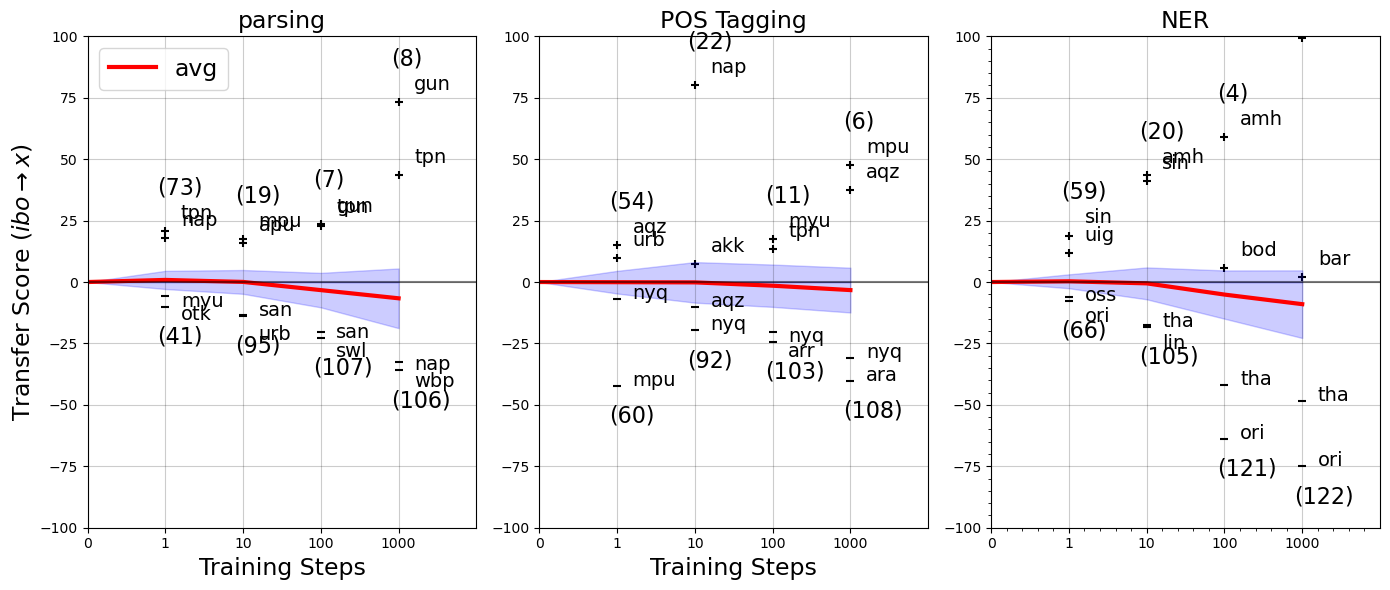} \\

kin \\
\includegraphics[width=.8\textwidth]{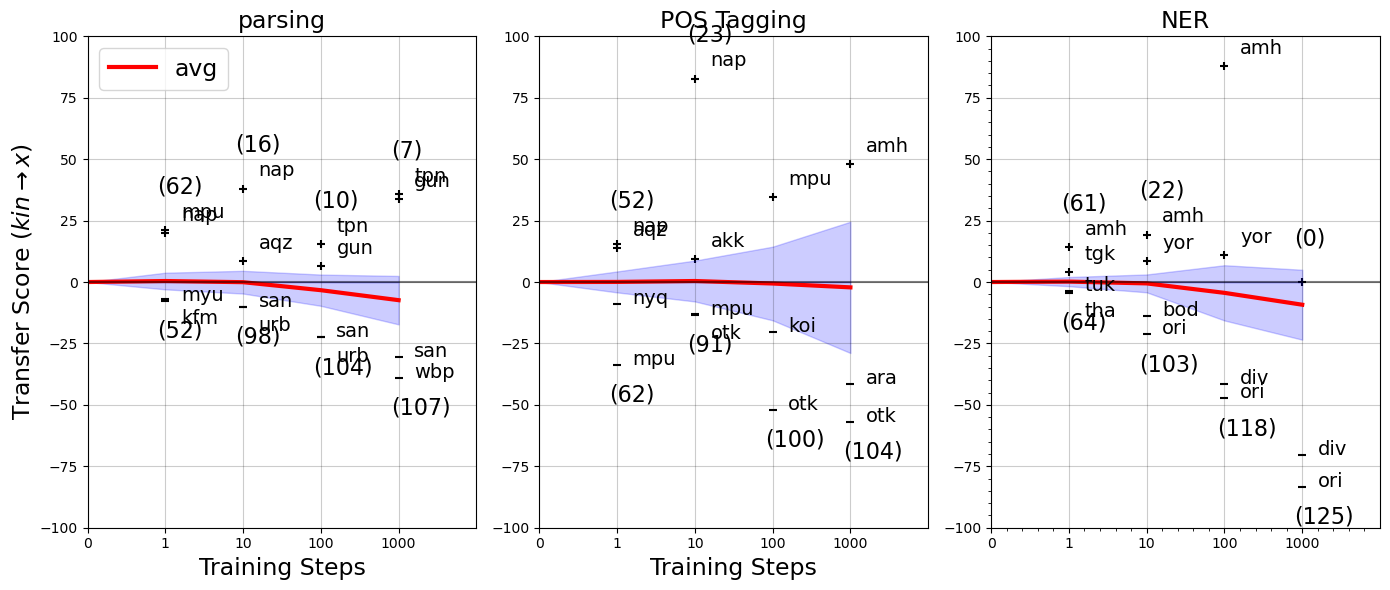} \\
mos \\
\includegraphics[width=.8\textwidth]{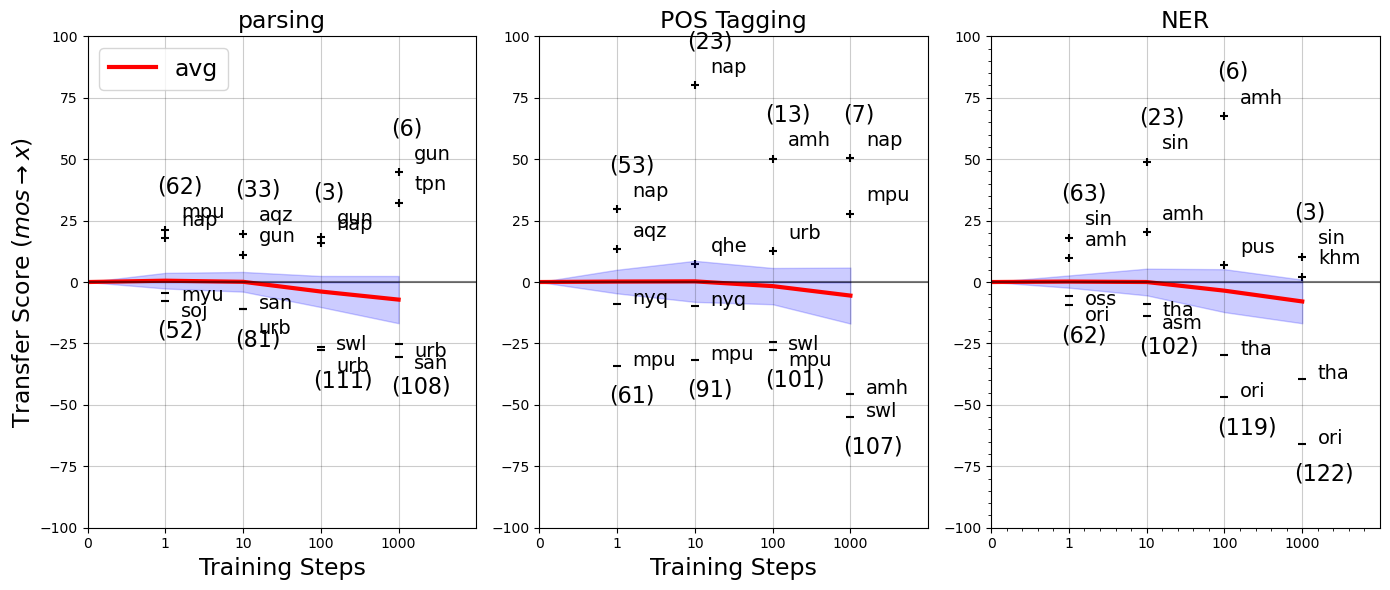} \\
pcm \\
\includegraphics[width=.8\textwidth]{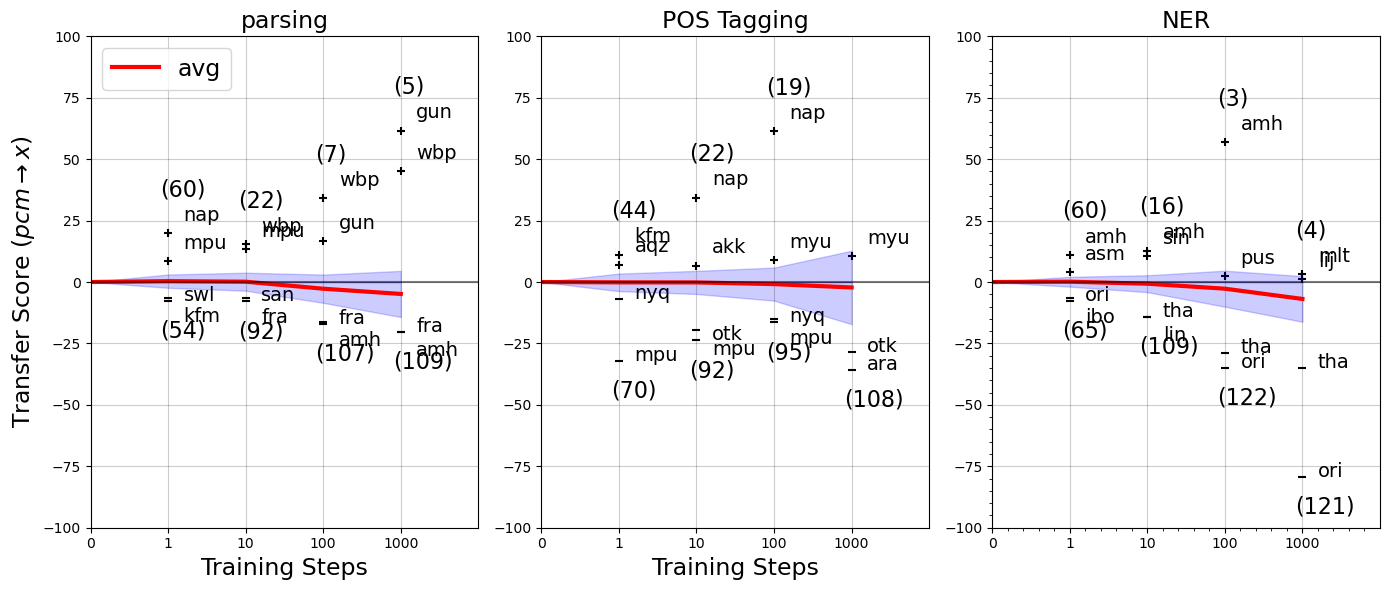} \\

    \end{tabular}
    \caption{Aggregated Transfer Progression through training steps}
    \label{fig:plot_gen}
\end{figure*}

\begin{figure*}[t]
    \centering
    \begin{tabular}{c}
wol \\
\includegraphics[width=.8\textwidth]{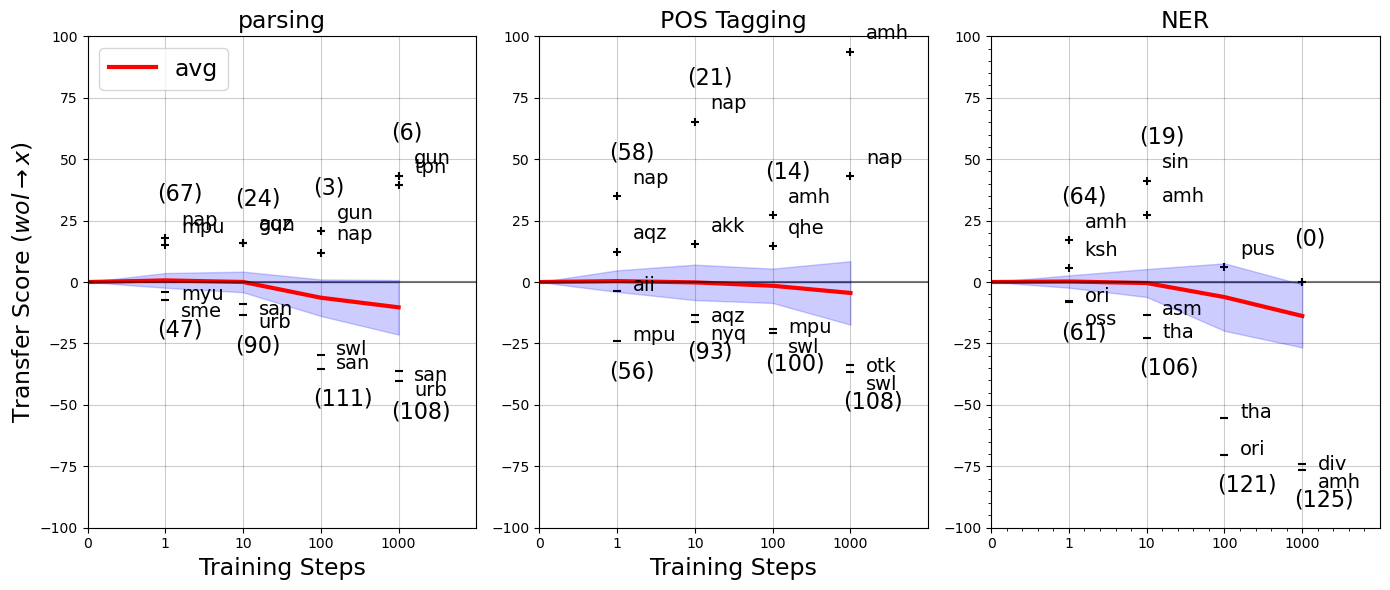} \\
yor \\
\includegraphics[width=.8\textwidth]{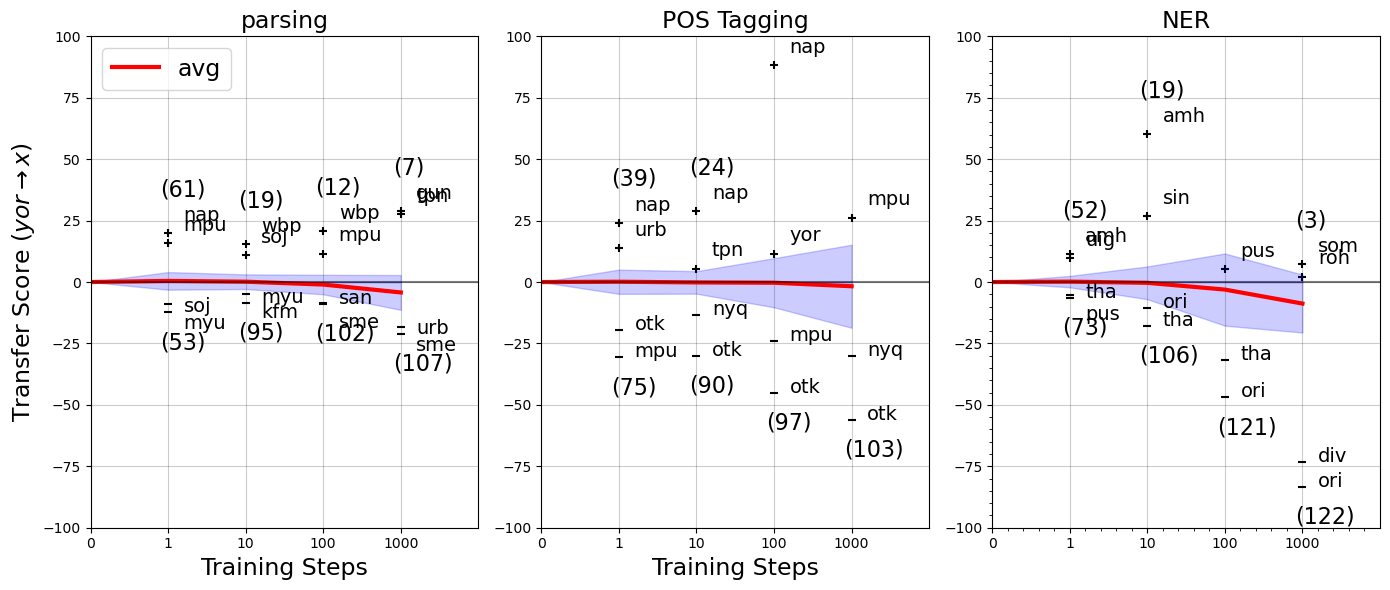} \\

    \end{tabular}
    \caption{Aggregated Transfer Progression through training steps}
    \label{fig:last}
\end{figure*}

